\documentclass{article}

\PassOptionsToPackage{numbers, compress}{natbib}

\usepackage[preprint]{neurips_2024}

\usepackage[utf8]{inputenc} 
\usepackage[T1]{fontenc}    
\usepackage{hyperref}       
\usepackage{url}            
\usepackage{booktabs}       
\usepackage{amsfonts}       
\usepackage{nicefrac}       
\usepackage{microtype}      
\usepackage{xcolor}         

\usepackage{graphicx}
\usepackage[font=small,labelfont=bf]{caption}

\title{Large Language Models for Subjective Language Understanding: A Survey}

\author{%
  Changhao Song \\
College of Intelligence and Computing\\
  Tianjin University\\
  \texttt{songchanghao@tju.edu.cn} \\
  \And
  Yazhou Zhang\thanks{Corresponding authors.} \\
College of Intelligence and Computing\\
  Tianjin University\\
  \texttt{yzhou\_zhang@tju.edu.cn} \\
  \And
  Hui Gao \\
College of Intelligence and Computing\\
  Tianjin University\\
  \texttt{hui\_gao@tju.edu.cn} \\
  \And
  Ben Yao \\
School of Nursing\\
  The Hong Kong Polytechnic University\\
  \texttt{benyao@polyu.edu.hk} \\
  \And
  Peng Zhang$^{*}$ \\
College of Intelligence and Computing\\
  Tianjin University\\
  \texttt{pzhang@tju.edu.cn} \\
}

\begin{document}

\maketitle
\begin{center}
\captionsetup{type=figure}
\includegraphics[width=0.9\linewidth]{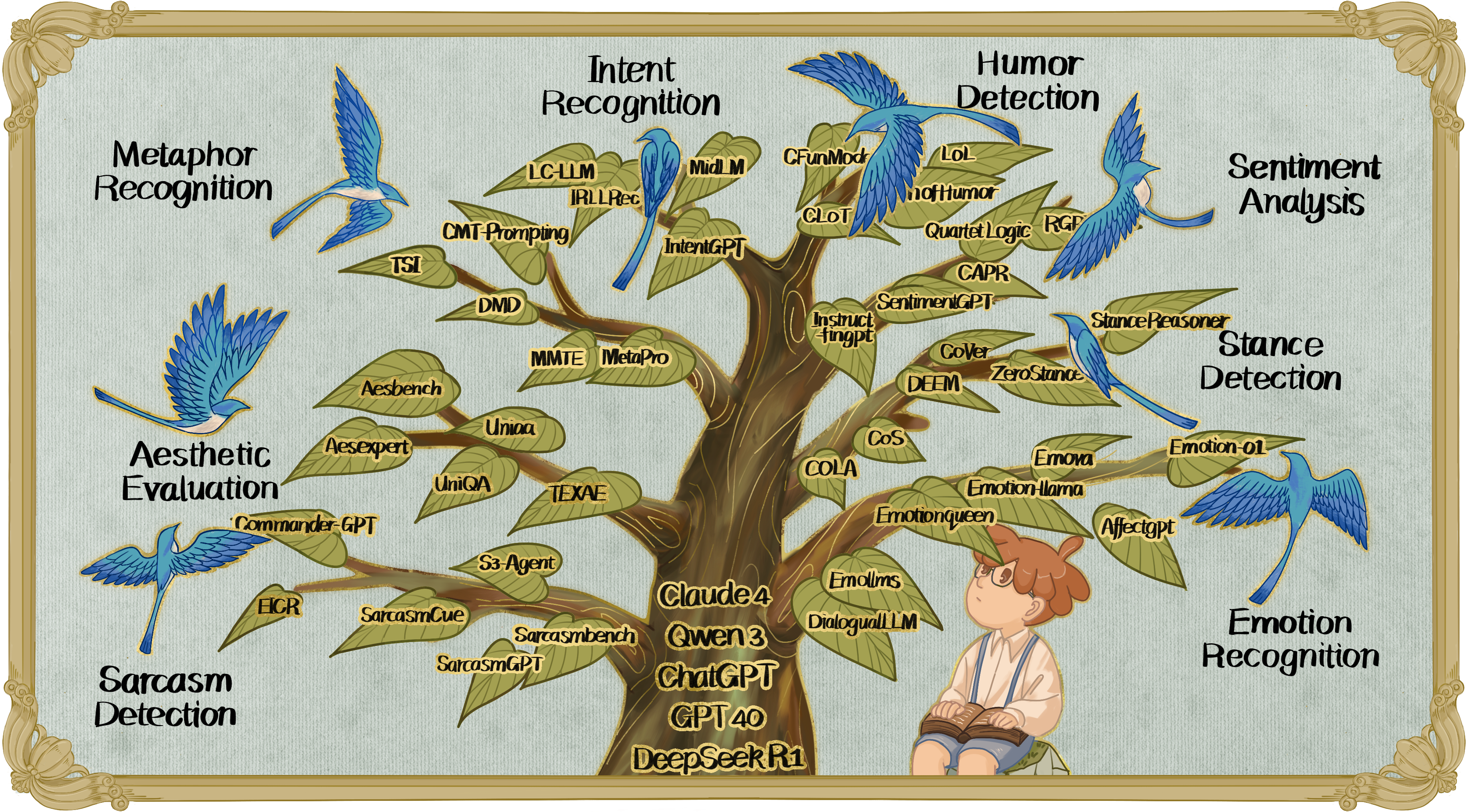}
\label{fig:graphical-abstract}
\end{center}

\begin{abstract}
  Subjective language understanding refers to a broad set of natural language processing tasks where the goal is to interpret or generate content that conveys personal feelings, opinions, or figurative meanings rather than objective facts. With the advent of large language models (LLMs) such as ChatGPT, LLaMA, and others, there has been a paradigm shift in how we approach these inherently nuanced tasks. In this survey, we provide a comprehensive review of recent advances in applying LLMs to subjective language tasks, including sentiment analysis, emotion recognition, sarcasm detection, humor understanding, stance detection, metaphor interpretation, intent detection, and aesthetics assessment. We begin by clarifying the definition of subjective language from linguistic and cognitive perspectives, and we outline the unique challenges posed by subjective language (e.g. ambiguity, figurativeness, context dependence). We then survey the evolution of LLM architectures and techniques that particularly benefit subjectivity tasks, highlighting why LLMs are well-suited to model subtle human-like judgments. For each of the eight tasks, we summarize task definitions, key datasets, state-of-the-art LLM-based methods, and remaining challenges. We provide comparative insights, discussing commonalities and differences among tasks and how multi-task LLM approaches might yield unified models of subjectivity. Finally, we identify open issues such as data limitations, model bias, and ethical considerations, and suggest future research directions. We hope this survey will serve as a valuable resource for researchers and practitioners interested in the intersection of affective computing, figurative language processing, and large-scale language models.

\end{abstract}

\section{Introduction}

\subsection{Background and Motivation}

Human communication is rich with \textbf{subjective language} – expressions of sentiment, emotion, opinion, humor, sarcasm, metaphor, and other non-literal or evaluative meanings. Understanding such language is crucial for AI systems that interact with humans or analyze human-generated text. Traditional NLP approaches to these tasks often relied on task-specific models and carefully crafted features or annotated data. For example, sentiment analysis has long been tackled with machine learning classifiers or pre-trained language models fine-tuned on labeled sentiment corpora, and sarcasm detection research evolved from manual pattern-based methods to neural models over the past decade. However, these narrow models typically handled each subjective phenomenon in isolation and lacked generalization: a model trained for sentiment would not handle humor, and vice versa.

The rise of large language models (LLMs) has brought new opportunities to address subjective language understanding in a more unified and general way. Modern LLMs such as OpenAI’s GPT-4, Google’s PaLM, Meta’s LLaMA, etc., have demonstrated remarkable capabilities in natural language understanding and generation across a wide range of tasks, via techniques like zero-shot/few-shot prompting and instruction tuning. Intuitively, many subjective language tasks might benefit from these capabilities: for instance, an LLM might recognize subtle sarcastic cues by virtue of having seen many examples in its training data, or it might generate more empathetic responses by leveraging learned patterns of emotional expression. Early successes with LLMs (e.g., GPT-3) on tasks like zero-shot sentiment analysis suggested that they internalize a great deal of subjective semantic knowledge. As a result, the community has shifted toward exploring how prompting or fine-tuning LLMs can solve affective and subjective NLP tasks that were previously considered very challenging.

Despite this optimism, subjective language understanding remains unsolved. These tasks often require nuanced contextual and commonsense reasoning, understanding of tone and pragmatics, and even a theory of mind. There are growing research efforts to evaluate how well LLMs truly “understand” emotions, humor, or figurative meanings, and results have been mixed. For example, while LLMs have made progress in sentiment analysis and emotion recognition, they still struggle with sarcasm and humor. In one study, GPT-4 was found to perform roughly at a human level on sentiment, emotion intensity, and political stance classification, but sarcasm detection remained a stumbling block. Such findings motivate a closer look at each type of subjective task to identify what unique challenges it poses and the progress current LLMs have made toward meeting those challenges.

\subsection{What Is Subjective Language}

Subjective language can be defined as any utterance or text whose meaning or interpretation depends on personal perspectives, feelings, or opinions, rather than objective facts. From a linguistic perspective, subjectivity in language is often indicated by the presence of opinionated words, emotion-laden expressions, first-person viewpoints, or figurative devices. For example, the sentence “The movie was an absolute masterpiece!” is subjective because it expresses the speaker’s positive evaluation (it’s not a verifiable fact, but an opinion). Similarly, “I feel upset about what happened” is subjective, revealing an emotional state. Linguists such as Banfield and Wiebe have studied how subjective expressions can be marked in text (e.g., through certain adjectives, intensifiers, or discourse structures), and how they differ from objective statements of fact. A classic NLP problem formulation is subjectivity classification: determining if a given sentence is subjective or objective. This can be seen as a coarse form of subjective language understanding.

From a cognitive perspective, subjective language relates to the speaker’s or writer’s internal state – their emotions, attitudes, beliefs, or intentions. Cognitive and psychological studies of language indicate that understanding subjectivity often requires theory of mind (inferring the speaker’s intent or feelings) and empathy. For instance, sarcasm and irony are quintessential subjective uses of language: the literal content diverges from the intended meaning, and the listener must infer the speaker’s true attitude (often the opposite of the literal words). Similarly, humor involves cognitive processes like surprise, incongruity, and shared knowledge between the interlocutors. Affective science provides insights into how humans express and perceive emotions through language (e.g. certain metaphors like “heartbroken” to indicate sadness). Thus, subjective language understanding is inherently interdisciplinary, bridging NLP with cognitive psychology.

In this survey, subjective language is an umbrella term encompassing affective language (sentiments, emotions, and attitudes) and figurative or non-literal language (sarcasm, humor, metaphor, etc.), as well as other subjectivity phenomena like personal intent or aesthetic preference. All these facets share the quality that purely literal or surface-level analysis often fails – one must grasp the underlying subjective meaning. We will clarify the scope of tasks covered in the next subsection.

\subsection{The Scope of The Paper}

The Scope of this Paper includes eight interrelated tasks that we categorize as core to Subjective Language Understanding: (1) Sentiment Analysis, (2) Emotion Recognition, (3) Sarcasm Detection, (4) Humor Detection, (5) Stance Detection, (6) Metaphor Recognition, (7) Intent Detection, and (8) Aesthetics Identification. These tasks span a range of applications and research communities, from traditional sentiment analysis in product reviews, to detecting humorous or sarcastic content on social media, to identifying a user’s intent in dialogue systems, to evaluating the aesthetic quality of creative content. By no means is this list exhaustive of all subjective phenomena in language (for example, bias detection, hate speech/offensive tone detection, and moral sentiment analysis are also subjective tasks, but we focus on the eight listed areas as they are most prominently addressed with LLM-era techniques in our surveyed literature). We aim to provide a unified treatment, highlighting common challenges and techniques, while also diving into task-specific details.

Figure 1 provides a conceptual taxonomy of these tasks, which we describe here. Affective tasks include sentiment analysis and emotion recognition: these deal with identifying feelings or attitudes expressed in text. Sentiment analysis typically focuses on polarity (positive/negative/neutral sentiment towards a target), whereas emotion recognition assigns more fine-grained emotion categories (happy, sad, angry, etc.) or even emotion intensity levels. Figurative language tasks cover sarcasm, humor, and metaphors. They require understanding non-literal meanings and often involve cultural or contextual knowledge – sarcasm and humor can overlap (sarcasm is often a bitter form of humor), and metaphors are imaginative expressions mapping one concept onto another. Stance detection is about inferring the position (pro/con/neutral) of the author with respect to a specific topic or claim – it’s subjective in that it reveals opinion, though often about external issues (politics, etc.). Intent detection (in user conversations or commands) is somewhat different but still subjective: it involves understanding the underlying goal or intention behind an utterance (for example, whether a question is actually a request, or what the user wants to achieve). Finally, aesthetics identification is an emerging area where the task is to evaluate the aesthetic or subjective quality of content – often images (image aesthetics rating) but also text style. It intersects with sentiment and with multi-modal understanding.

Despite differences, all these tasks are unified by requiring the model to go beyond literal meaning and often to incorporate world knowledge and cultural context. Our survey specifically investigates how LLMs have been applied to each task, and what advantages or limitations they bring.

\subsection{Distinction from Existing LLM Surveys}

It is important to clarify how our survey differs from prior surveys, especially those focusing on LLMs or affective computing. One closely related work is \cite{zhang2024affective}. Their survey centers on how LLMs can be used for affective computing tasks, mainly sentiment and emotion analysis and affective text generation. In contrast, our survey covers a broader notion of subjective language, not limited to emotions but also including stance, figurative language (sarcasm, humor, metaphor), user intent, and aesthetic judgement. Thus, we address a wider range of tasks under the umbrella of subjectivity. Another difference is in emphasis: we delve into how LLMs perform understanding (analysis) of subjective language. We also include tasks like metaphor and humor which might not be treated in-depth in an affective computing survey.

Moreover, existing general LLM surveys mention these tasks only briefly, if at all. To our knowledge, this is the first comprehensive survey specifically targeting subjective language understanding in the LLM era. We synthesize results from over 200 recent papers and highlight trends such as prompt engineering for subjectivity, multi-task learning of subjective phenomena, and integrating domain knowledge into LLMs. We also draw on benchmarks and studies evaluating LLMs on these tasks.

\subsection{Contribution and Structure of The Paper}

The contributions of this survey are as follows:
\begin{itemize}
\item We define and motivate subjective language understanding as a field, clarifying its scope and importance in NLP. We connect linguistic definitions of subjectivity with the challenges faced by AI, providing a conceptual foundation for readers (Section 2).

\item We provide an overview of LLMs (Section 3) with a focus on their relevance to subjective tasks. This includes a brief history of language model evolution leading to current state-of-the-art models, and a discussion of why the properties of LLMs (such as in-context learning and knowledge integration) make them promising for subjective language understanding.

\item For each of the eight tasks (Sections 4 - 11), we present a task definition, key datasets, LLM-based methods, and challenges. We thoroughly review literature in each area: for instance, how LLMs have been fine-tuned or prompted for sentiment analysis, how they’ve been evaluated on humor and sarcasm, what novel techniques have been proposed, etc. We highlight representative papers and methods, and we analyze their strengths and weaknesses in context. Wherever applicable, we cite quantitative results from papers or benchmarks to give a sense of the state-of-the-art performance.

\item We perform a comparative analysis in Section 12, discussing commonalities and differences among the tasks. We examine, for example, how sarcasm detection and humor detection overlap in needing cultural knowledge, or how sentiment and emotion recognition differ in granularity but share methodical approaches. We also discuss the potential of unified models or multi-task training to handle multiple subjective tasks together, referencing any multi-task studies we found. We compare single-task fine-tuning versus multi-task (or instruction-based) approaches in the context of subjectivity: which yields better performance or efficiency, based on recent experiments.

\item We outline challenges and open issues (Section 13) that emerged from the literature review. These include technical challenges (e.g., handling context and pragmatics, avoiding LLM hallucinations in subjective inference, data scarcity for less common tasks), as well as ethical considerations (e.g., the risk of bias when an AI system judges what is “beautiful” or interprets user emotion, and privacy issues in emotion/intent detection). We also discuss how subjective language understanding by AI can impact society (for instance, the use of stance detection in monitoring social media could raise fairness concerns).

\item We conclude (Section 14) by summarizing key findings – for example, which tasks LLMs have significantly advanced and which remain very challenging – and by calling for a unified research framework and evaluation for subjective language understanding. We emphasize that as LLMs become central to NLP, it’s crucial to develop standardized benchmarks that cover the spectrum of subjective tasks, and to encourage research that bridges these areas rather than treating each in isolation. Ultimately, truly human-like language understanding by AI will require competence in all these subjective dimensions. We hope our survey accelerates progress toward that goal.
\end{itemize}

The structure of the paper follows the outline above. Readers interested in specific tasks can refer directly to Sections 4–11 for detailed surveys of each area. We now proceed to formally define subjective language understanding and present a taxonomy of tasks (Section 2), before discussing LLM foundations (Section 3) and then diving into each task.

\section{Defining Subjective Language Understanding}
\subsection{Definitions of Subjectivity: Linguistic and Cognitive Perspectives}

Subjectivity has been a topic of interest in both linguistics and cognitive science, each providing a complementary perspective. From the linguistic perspective, subjectivity in language is about the expression of personal stance. Linguist Janet Besnier noted that subjectivity is “the linguistic encoding of the speaker’s perspective” – this can manifest as opinions, evaluations, or other attitude markers in text. Classic work by Wiebe et al. (2004) in computational linguistics distinguished subjective sentences (those containing opinions, sentiments, or feelings) from objective sentences (factual descriptions). Linguistically, clues to subjectivity include: opinion adjectives (e.g. “beautiful,” “terrible”), modal verbs and hedges (which indicate uncertainty or perspective, e.g. “I think,” “probably”), first-person references (“I believe...”), and intensifiers (“very happy,” “extremely costly”). Even punctuation or tone words (exclamation marks, emotive interjections like “ugh”) signal subjectivity. These linguistic markers have been used historically to build subjectivity lexicons and classifiers. For example, a sentence like “In my opinion, this is a huge mistake!” is clearly subjective due to the phrase “in my opinion” and the evaluative term “huge mistake.” On the other hand, “The water boils at 100°C.” is objective. However, there are many gray areas and subtle cases; a sentence can convey a subjective attitude without explicit markers, especially if context is required.

From the cognitive perspective, subjectivity ties into how humans process language and infer others’ mental states. Cognitive scientists consider Theory of Mind (ToM) – the ability to attribute thoughts, intentions, or emotions to others – as crucial for understanding subjective aspects of communication. When someone says “Sure, I just love getting stuck in traffic for hours,” an listener with theory of mind will recognize the likely sarcastic intent (the speaker’s true attitude is the opposite of the literal words). Thus, cognitively, subjective language often demands inference beyond the literal text, involving knowledge of speaker intentions, cultural context, and sometimes shared experiences. Emotion understanding is another cognitive aspect: humans have an innate ability to read emotional cues in language (certain words or even the rhythm of text can imply an emotional state). Cognitive and social psychology also discuss how people use language to perform actions. For instance, being polite or rude, being humorous or serious. These aspects highlight that subjective language understanding is not just a textual analysis problem, but an exercise in modeling human-like interpretations.

In summary, linguistically we can describe subjectivity through observable markers in language, while cognitively we explain subjectivity by the mental processes a listener/reader uses to interpret those utterances. An effective AI system must bridge both: detect the markers and patterns, and apply reasoning to interpret them correctly.

\subsection{Key Characteristics of Subjective Language}

What makes subjective language particularly challenging for computational models? We outline a few key characteristics:

 \paragraph{Ambiguity and Subtlety} Subjective expressions are often ambiguous. The same phrase can have different meanings depending on context or tone. For example, “Yeah, right.” could be sincere agreement or a sarcastic dismissal, depending on context and perhaps the speaker’s intonation. Subjective language relies heavily on context (both linguistic context and real-world context). Small cues can flip the interpretation. This subtlety is why tasks like sarcasm detection are hard – there is no single keyword that always signals sarcasm.

 \paragraph{Figurative and Non-literal meaning} Much of subjective language is non-literal. Metaphors, idioms, and jokes involve meaning that cannot be obtained by straightforward dictionary lookup. For instance, “kick the bucket” meaning “to die” or “spill the tea” meaning “to gossip” are idiomatic and subjective. Similarly, metaphors like “a rollercoaster of emotions” convey subjective experience via analogy. LLMs have shown some ability to interpret idioms and metaphors, which is a positive sign. But generating or identifying non-literal language remains challenging. The non-literal nature often overlaps with humor and sarcasm.

 \paragraph{Presence of Implicit Context or Knowledge} Understanding subjective content often requires commonsense or cultural knowledge. A joke might rely on a cultural reference; a sentimental statement might assume knowledge of what's considered positive or negative in a domain. Stance detection requires knowing the topic discussed. For example, “We need another Einstein in our time.” – to understand the stance, one needs to know Einstein = symbol of genius, implying we lack genius now. LLMs have a lot of world knowledge, giving them an edge in subjective tasks compared to earlier models. However, knowledge can be a double-edged sword if not properly constrained – e.g., an LLM might hallucinate facts to justify an emotional inference.

 \paragraph{Highly Subjective Evaluation} Subjective tasks often lack a single “ground truth” among humans. This is reflected in inter-annotator disagreement for datasets. Emotion or sentiment labels can vary between annotators. Humor is famously subjective: one person’s joke might fall flat for another. Aesthetic judgments differ widely. This characteristic means models might reflect one plausible interpretation even if it doesn’t match a gold label. It complicates both training and evaluation. Recent works have used distributional evaluation or multiple human ratings to mitigate this, and LLMs might output a probability or score that correlates with degree of human agreement.

 \paragraph{Influence of Personal and Societal Biases} Subjective language ties to personal perspective, which can reflect biases. Models learning from subjective data risk absorbing biases (e.g., associating certain sentiments with demographic groups, or having skewed humor that might be offensive). We highlight this because it’s both a characteristic and an ethical challenge: understanding subjective language requires recognizing whose perspective is reflected (for instance, the stance in a tweet may depend on the tweeter’s political alignment). LLMs need mechanisms to handle this – either by being neutral or not amplifying harmful biases. This is discussed more in Section 13 on ethical implications.

These characteristics show why subjective language understanding is a tough problem for AI and why success here is a good proxy for genuine natural language “understanding,” as it goes beyond surface text. Next, we classify the tasks under subjective language understanding in a unified taxonomy.

\subsection{A Unified Taxonomy of Subjective Tasks}

We identified 8 key tasks as the focus. Here, we briefly define each and position it in our taxonomy:

 \paragraph{Sentiment Analysis} Determine the sentiment polarity expressed in a given text, often categorized as positive, negative, or neutral. Sometimes “sentiment analysis” also includes aspect-based sentiment analysis (polarity toward specific aspects of an entity) and intensity (strength of sentiment). It traditionally answers questions like, “Is this product review positive or negative?” or “How does this tweet’s author feel about topic X?”. In our taxonomy, sentiment analysis is a fundamental subjective task concerning evaluative attitude. It is usually considered “simpler” than full emotion recognition because it deals with broad valence (good/bad) rather than specific emotions. However, it is ubiquitous in industry (e.g., opinion mining) and is a cornerstone of affective NLP.

  \paragraph{Emotion Recognition} Identify the emotion(s) expressed in a text. This could be a classification into categories (joy, sadness, anger, fear, etc.) or a regression in an emotional dimension (such as valence, arousal). Emotion recognition can be seen as richer labeling than sentiment: “I am furious about the delay” has negative sentiment, but more specifically the emotion is anger. Emotion recognition might involve multiple labels if more than one emotion is present. We also include related tasks like emotion cause detection under this umbrella, although the main focus is classification of emotion from text. This lies in the affective branch of our taxonomy, alongside sentiment.

  \paragraph{Sarcasm Detection}  Determine if a given text is sarcastic or not. Sarcasm is usually a form of verbal irony where the intended meaning is opposite to the literal wording, often to mock or convey contempt. For instance, “Oh, great, another Monday morning meeting. I’m so excited.” is likely sarcastic. This task is binary (sarcastic vs not), though some research considers degrees of sarcasm or types of sarcastic expression. Sarcasm detection is a prototypical figurative language task in our taxonomy, requiring high-level pragmatic inference. It’s notoriously hard because it depends on subtle cues and sometimes knowledge of the speaker’s personality or context. We include irony detection here as well, as computationally the two overlap a lot.

  \paragraph{Humor Detection} Identify if a text is intended to be humorous or not (and possibly, how funny it is). Humor detection overlaps with sarcasm in that both involve non-literal cues and surprise, but humor is broader – not all humor is sarcastic; it could be puns, absurdity, etc. This task might be binary (humorous vs not) or involve scoring jokes by funniness. It’s subjective because humor reception varies across audiences. It sits with sarcasm under figurative language understanding. An example: “I told my computer I needed a break, and now it won’t stop sending me KitKat ads.” A model should detect this is a joke (wordplay on “break”). Humor understanding might also involve explaining the joke, but our focus is mainly on detection/classification and understanding.

  \paragraph{Stance Detection} Given a text and a specific target or claim, determine whether the author’s stance is in favor, against, or neutral toward the target. For example, in a debate forum post about climate change, does the author support or oppose the existence of human-induced climate change? Stance is similar to sentiment, but specifically anchored to a target proposition and not necessarily about personal feelings – one can have a stance on an issue without an emotional tone. Still, it’s subjective as it reflects opinion. Stance detection can be closed-target or open-target. This task is important in analyzing social media, fake news, and online conversations. In our taxonomy it is somewhat between affective and opinion tasks – we categorize it under subjective opinion analysis.

  \paragraph{Metaphor Recognition}  Determine which words or phrases in a text are used metaphorically (as opposed to literally), or more generally identify and interpret metaphors. For example, in “After the argument, a wave of anger washed over him,” the phrase “wave of anger” is metaphorical. Metaphor recognition can be a token-level sequence labeling task (label each word as literal or metaphoric) or a classification of a phrase/sentence as containing metaphor. It’s a figurative language task. Interpretation of metaphor is a related challenge – e.g., GPT-4 has shown an ability to interpret novel metaphors by providing explanations. In our scope, we primarily consider recognition. Metaphors are subjectively used to convey concepts in a more vivid way, often tied to creativity and cognition.

  \paragraph{Intent Detection} Intent detection involves classifying a user’s utterance according to its underlying intent. It’s a key component in task-oriented dialogue systems. Although this seems more “semantic” than “subjective,” we include it because recognizing user intent is related to interpreting implicit meaning in their request – essentially a pragmatic understanding task. For instance, the user query “I’m hungry” has the intent FindRestaurant implicitly. Or in open-ended conversation, “It’s cold here” could be an indirect intent for the thermostat to be turned up. Intent detection also includes detecting intent strength or ambiguity. It’s subjective as the model must infer the human’s goal from context, and different users might phrase intentions in diverse, personal ways.

  \paragraph{Aesthetics Identification}  This is a relatively novel task in NLP – assessing the aesthetic quality or style of content. Traditionally, this has been more common in computer vision (image aesthetics rating), but with multi-modal models and stylistic text generation, it’s coming to NLP. Here we consider tasks like: given an image and possibly a description, rate its aesthetic appeal; or given a piece of text, judge its writing style aesthetics (is it eloquent, is it engaging). The Textual Aesthetics work (Jiang et al., 2024) introduced a dataset and method to fine-tune LLMs to produce more aesthetically pleasing text outputs. And on the image side, AesBench (Huang et al., 2024) is a benchmark that asks LLM-based vision-language models to perform various aesthetic understanding tasks on images. This task is subjective by nature – “beauty is in the eye of the beholder.” It intersects with sentiment (pleasing vs not pleasing) but goes beyond, into artistic elements and human preference. We place it in its own category, touching both affective and cognitive.

  This taxonomy shows the diverse landscape we cover. Relationships exist between tasks: sarcasm and humor are linked, sentiment and stance both deal with evaluation but target differently, emotion and intent sometimes intersect. One ambition of the field is to handle overlapping phenomena jointly.

\subsection{Why LLMs Matter for Subjectivity}
We expect LLMs to be particularly suitable for subjective language understanding for several reasons, supported by recent research:

\paragraph{In-Context Learning \& Few-Shot Ability} LLMs can perform tasks with little to no task-specific training, by virtue of prompting. This is very useful for tasks where we might not have large labeled datasets. For example, prompting GPT-4 with “Is the following statement sarcastic? …” and a bit of instruction can yield reasonable answers, whereas smaller models would fail without explicit training. This makes it feasible to tackle low-resource subjective tasks. The ability to incorporate a few examples in the prompt (few-shot learning) can further improve performance, essentially allowing the model to adapt on the fly to the style of the task or domain.

\paragraph{Knowledge and Common Sense} LLMs embed a vast amount of world knowledge and common sense acquired during pre-training. Many subjective interpretations require such knowledge. For instance, understanding the humor in “My phone’s battery is the Usain Bolt of dying” requires knowing Usain Bolt is extremely fast (so the phone dies fast – a humor through metaphor). A well-trained LLM likely knows about Bolt and can connect “fast at dying” as a humorous exaggeration. The knowledge aspect also helps in stance detection (knowing background of topics), and in metaphors (knowing typical mappings and even rare ones). This is something earlier task-specific models lacked; they’d see words but not truly “know” facts or cultural references.

\paragraph{Advanced Language Generation for Explanations} For subjective tasks, generating explanations or justifications is valuable (for interpretability and possibly improving accuracy). LLMs can produce natural language explanations via chain-of-thought prompting or by design. For example, an LLM might detect sarcasm better if prompted to explain the joke and then decide—essentially using its generative prowess to reason. Some approaches use chain-of-thought (CoT) prompting, where the model thinks step by step about why a sentence might be sarcastic. This capability of explaining makes LLMs flexible, turning implicit tasks into explicit reasoning processes (e.g., “The sentence says X but likely means Y, because ...”).

\paragraph{Multi-task and Transfer Learning at Scale} LLMs are typically trained on diverse internet text. This means, for instance, they’ve seen both factual text (Wikipedia articles) and subjective text (tweets, novels, reviews) to some extent. This exposure might allow transfer learning internally; the model could transfer what it “learned” about sentiment while reading movie reviews to help in understanding sentiment in a new context. The scale of training might also let it capture patterns that smaller models miss. Research has indeed indicated that certain abilities (like understanding idioms or performing basic reasoning) emerge only as model scale increases. 

\paragraph{Unified Handling of Language and Multi-modality} Some of the latest “LLM” systems are multi-modal. For tasks like aesthetics, which involve images, or emotion recognition from multimodal cues, these architectures extend the LLM paradigm. The extension of these to subjective queries (e.g., “Is this person in the image happy or sad?” or “Rate the aesthetics of this photograph”) is a current research frontier. LLMs provide a coherent way to integrate modalities – by converting everything to a “language” (descriptions, dialogues) and then processing with a powerful language reasoning core. This could be more effective than earlier multi-modal systems that treated vision and text separately. Our survey will touch on some multimodal aspects (especially in aesthetics and in a few humor/sarcasm datasets that have context or images).

In summary, LLMs matter for subjectivity because they bring general intelligence-like capabilities to NLP: flexibility, knowledge, and adaptability. However, as we will see, they are not a panacea. There are also reasons LLMs might struggle or require augmentation: e.g., they might lack true understanding of emotion (they predict patterns but don’t “feel”), they might have biases, and they might produce convincing but incorrect interpretations. Throughout the survey, we will evaluate how well the promise of LLMs translates into actual task performance, citing concrete results.

Having set the stage, we will first provide a brief overview of LLMs — their evolution and current state of the art (Section 3) — before exploring each subjective language task in Sections 4–11.





\section{Large Language Models (LLMs) for Subjective Language Understanding}

\subsection{Evolution of Language Models and Emergence of LLMs}

In recent years, natural language processing has evolved dramatically in language modeling. Early models like word2vec\cite{mikolov2013efficient} and RNNs\cite{elman1990finding} (2012–2017) captured local patterns but were contextually limited. The Transformer architecture \cite{vaswani2017attention} enabled deeper, larger models. OpenAI's GPT series exemplifies this shift. GPT-1\cite{radford2018improving} (2018) with ~117M parameters highlighted pre-training on unlabeled text and task fine-tuning. GPT-2\cite{radford2019language} (2019) expanded to 1.5B parameters, showcasing coherent text generation. The breakthrough came with GPT-3\cite{brown2020language} (2020), at 175B parameters, demonstrating strong zero-shot and few-shot learning across tasks like sentiment analysis and translation without explicit training—emerging as a general-purpose NLP tool. This paradigm was quickly adopted, leading to other LLMs like Google's T5\cite{raffel2020exploring}, BERT-like encoders, and larger models such as PaLM\cite{chowdhery2023palm} (540B) and Megatron-Turing NLG\cite{smith2022using} (530B).

In 2022–2023, instruction tuning and interactive LLMs, known as chatbot models, emerged. OpenAI's InstructGPT\cite{ouyang2022training} and ChatGPT (based on GPT-3.5) were fine-tuned with human feedback to better follow instructions and conversational cues, aligning with human preferences for greater practical effectiveness. GPT-4\cite{achiam2023gpt} (2023) enhanced performance, especially in complex reasoning tasks, though its architecture and size remain unconfirmed, with estimates suggesting >170B parameters and novel training methods. Concurrently, Meta AI released LLaMA\cite{touvron2023llama} and LLaMA-2\cite{touvron2023llama2} (7B–70B parameters), open-source models that democratized LLM research. Other entrants included Claude by Anthropic and Baidu’s ERNIE. Current state-of-the-art LLMs often involve ensembles or instruction-tuned versions of these base models, with open-source projects like Alpaca\cite{taori2023alpaca} and Vicuna\cite{chiang2023vicuna} fine-tuning LLaMA for ChatGPT-like functionality.

In the context of subjective language, models specifically to handle such tasks have emerged: e.g., a model named SentimentGPT\cite{kheiri2023sentimentgpt} was proposed by Kheiri \& Karimi (2023) which analyzes how GPT-based models depart from classical ML in sentiment analysis. Specialized variants or prompting techniques (like emotion-aware LLMs) have been developed. Some research has tried to incorporate psychological theories into LLMs by fine-tuning or prompting\cite{yongsatianchot2023s}. Furthermore, the line between “language model” and “multimodal model” is blurring – GPT-4 and others can accept images as input in addition to text, allowing them to describe an image’s emotional content or aesthetics. This versatility positions LLMs as central hubs for processing subjective information across modalities.

To summarize this evolution: we went from task-specific small models, to moderate pre-trained models fine-tuned per task, to gargantuan models that can perform all tasks with minimal task-specific tuning. This is a paradigm shift: instead of building a separate classifier for sarcasm, we can now prompt one general model to do sarcasm detection, perhaps even alongside other tasks. It opens the door for multi-task subjective language models, which we discuss later in the survey (Section 12).

\subsection{Current State-of-the-Art Models}

The current landscape of Large Language Models (LLMs) for subjective language understanding is characterized by a diverse array of models and techniques, broadly categorizable into prompt-based, Supervised Fine-Tuning (SFT)-based, and reasoning-based approaches. Prominent models frequently cited in recent literature include OpenAI's GPT series (GPT-3.5, GPT-4, GPT-4o\cite{hurst2024gpt}), Google's Gemini\cite{team2023gemini} (Gemini 1.5 Flash\cite{team2024gemini}, Gemini 1.5 Pro, Gemini 2.0 Flash), Meta's Llama series (Llama-2, Llama-3\cite{dubey2024llama}, Llama-3.1), Mistral AI's models (Mistral 7B\cite{jiang2023mistral7b}, Mixtral 8x7B\cite{jiang2024mixtral}), and others like Qwen-2\cite{yang2024qwen2technicalreport}, and DeepSeek-R1\cite{guo2025deepseek} models.

\paragraph{Prompt-based LLMs} leverage the inherent capabilities of pre-trained models by providing carefully crafted input prompts to guide their responses for specific tasks. This includes zero-shot prompting, where the model performs a task without any prior examples, and few-shot prompting, where a small number of examples are included in the prompt to demonstrate the desired output. For subjective tasks, the design of the prompt is critical, as it can influence the model's perspective and reasoning. For example, techniques like Chain-of-Thought (CoT)\cite{wei2022chain} prompting, which encourage the model to generate intermediate reasoning steps, have been applied to improve performance on tasks requiring deeper understanding. However, research indicates that CoT prompting, especially for larger LLMs, might suffer from "posterior collapse," where the model relies more on pre-existing reasoning priors than on the evidence presented in the prompt, particularly in complex subjective domains like emotion and morality.

\paragraph{SFT-based LLMs} involve taking a pre-trained LLM and further training it (fine-tuning) on a specific dataset relevant to a particular subjective task. This process adapts the general knowledge of the LLM to the nuances of the target domain. For instance, models like RoBERTa\cite{liu2019roberta} and BERT\cite{devlin2019bert} have been fine-tuned for tasks such as subjectivity detection in news articles or sentiment analysis. While fine-tuning can lead to high performance on the specific dataset, it may not always generalize well to out-of-distribution data or other subjective tasks without further adaptation. Parameter-efficient fine-tuning (PEFT) methods like LoRA\cite{hu2022lora} (Low-Rank Adaptation) and QLoRA\cite{dettmers2023qlora} are also gaining traction, allowing for adaptation with reduced computational cost.

\paragraph{Reasoning-based LLMs} focus on enhancing the model's ability to perform logical inference and understand complex relationships, which is crucial for many subjective tasks. This includes methods that explicitly guide the model's reasoning process. For example, the "Reasoning through Perspective Transition" (RPT) method enables LLMs to dynamically select among direct, role, and third-person perspectives to solve subjective problems more effectively by ranking perspectives and choosing the most suitable one for a given scenario. Another approach, "Reasoning in Conversation" (RiC), simulates dialogues to mine useful contextual information for subjective tasks like metaphor recognition and dark humor detection, rather than relying solely on chain-of-thought rationales. These methods aim to overcome the limitations of standard prompting by encouraging more structured and adaptable reasoning. The development of models like GPT-4, which are reported to have improved reasoning capabilities, also falls under this umbrella.

The choice of model and approach often depends on the specific task, the availability of labeled data, computational resources, and the desired level of interpretability and generalization. For instance, while proprietary models like GPT-4 often lead in performance, open-source models like Llama and Mistral provide flexibility for customization and fine-tuning . The ongoing research explores hybrid approaches, knowledge distillation from larger to smaller models, and methods to improve the robustness and reliability of LLMs in subjective understanding.

\subsection{Multi-Model LLMs for Subjective Language Understanding}

The concept of Multi-Model LLMs is increasingly significant in addressing the complexities of subjective language understanding. This approach recognizes that a single LLM may be insufficient to capture all aspects of subjectivity, particularly with multimodal inputs or tasks requiring diverse expertise. For example, in sarcasm detection, visual cues accompanying text can be vital for clarifying meaning. The Commander-GPT\cite{zhang2025commander} framework is proposed for multimodal sarcasm detection, where a central LLM (e.g., GPT-4) coordinates specialized models ("generals") skilled in areas like image content analysis or textual analysis. This ensemble approach aims to leverage the strengths of different models for more robust sarcasm recognition than a single model can achieve. Similarly, in Speech Emotion Recognition (SER), systems are developed that combine audio encoders (e.g., Whisper-large-v3\cite{radford2023robust}) with LLMs (e.g., Gemma-2-2B-it\cite{team2024gemma}) to process both speech signals and transcriptions, forming a unified multimodal architecture. These systems align features from different modalities and manage token overload from high-dimensional audio embeddings.

Another dimension of Multi-Model encompasses ensemble methods or collaborative agent frameworks, wherein multiple LLM instances or diverse models collaborate. For subjectivity detection in news, an ensemble of multiple LLMs is employed, combining predictions via majority voting to enhance robustness and mitigate biases. In stance detection, the COLA\cite{lan2024stance} framework utilizes LLMs in a three-stage collaborative process, assigning distinct roles to address challenges like multi-aspect knowledge and advanced reasoning, enabling nuanced analysis beyond a single LLM’s capacity. Additionally, hybrid approaches integrating smaller, fine-tuned models (e.g., BERT) with larger LLMs (e.g., GPT-4, Llama-3) are emerging. For intent detection, uncertain predictions from a fine-tuned BERT may be routed to an LLM, with BERT information dynamically generating prompts for the LLM to reduce label space, balancing computational efficiency with advanced LLM capabilities. These multi-model strategies signify a shift towards developing sophisticated systems for subjective language understanding, surpassing reliance on a singular LLM.

To conclude this section, LLMs have rapidly become the toolkit of choice for subjective language understanding. We now have a variety of ways to use them (direct prompting, fine-tuning, etc.) and a variety of models to choose from. The remaining sections will detail each task. We will see that for tasks with limited data (like humor, sarcasm), creative prompting and large models (GPT-4) often lead the pack, whereas for tasks with lots of data (like sentiment), fine-tuned smaller models can still compete or outperform in some cases – though the gap is closing as LLMs get instruction-tuned on sentiment during their general training. Next, we dive into Sentiment Analysis as the first task, which historically is one of the most studied and will illustrate many general points.








\section{Sentiment Analysis}

\subsection{Task Definition of Sentiment Analysis}

Sentiment analysis, also known as opinion mining, is a core task in NLP that focuses on identifying, extracting, quantifying, and studying affective states and subjective information from text. The primary goal is to determine the attitude or emotional tone of a writer or speaker with respect to a particular topic, product, person, or entity. This attitude can be categorized in various ways, most commonly as \textit{positive, negative, or neutral}. More fine-grained approaches may also identify the intensity of the sentiment (e.g., very positive, slightly negative) or detect specific emotions. Sentiment analysis can be performed at different levels of granularity: \textbf{document-level} (classifying the overall sentiment of an entire document), \textbf{sentence-level} (determining sentiment for individual sentences), or \textbf{aspect-level} (identifying sentiment towards specific aspects or features of an entity mentioned in the text, e.g., "The camera is good, but the battery life is poor"). The task is crucial for a wide range of applications, including brand monitoring, market research, customer feedback analysis, product recommendation, and social media analytics. The challenge lies in accurately interpreting nuanced language, sarcasm, irony, and context-dependent expressions that can alter the perceived sentiment.

\subsection{Dataset of Sentiment Analysis}
Sentiment analysis has been built on a set of well-established supervised benchmarks spanning reviews and social media. Core movie and product review resources include \textbf{MR} (movie reviews, also known as the Polarity dataset), the Stanford Sentiment Treebank (\textbf{SST}; phrase-level labels with SST-2 binary and SST-5 five-class variants), \textbf{IMDb} (50k balanced reviews), \textbf{Amazon} product reviews (millions of star ratings), and \textbf{Yelp} Reviews (hundreds of thousands, widely used in LLM studies). Multi-domain review collections (e.g., books, electronics) are used to test domain shift. Large Twitter datasets became standard as well: \textbf{Sentiment140} (1.6M tweets, balanced binary) and the \textbf{Twitter US Airline Sentiment} dataset (14,160 tweets with positive/neutral/negative) are common baselines. The \textbf{SemEval-2017 Task 4} Twitter benchmarks remain central for topic-oriented sentiment. Earlier works also use the University of Michigan dataset and a series of domain-specific Twitter corpora, including Tariyal et al.’s 1,150 product-review tweets, Hemakala and Santhoshkumar’s 14,640 Indian airline tweets, and Rahat et al.’s ~10k tweets.

Beyond overall polarity, several datasets drive fine-grained and pragmatic analysis. \textbf{SemEval-2014 Task 4} established Aspect-Based Sentiment Analysis (ABSA), requiring sentiment toward specific aspects in text. Nuanced social media phenomena are captured by \textbf{SemEval-2018 Irony} in Tweets and sarcasm/emoji-focused resources. Extending this line, \cite{bhargava2025impact} released ~5,929 tweets about nuclear power with explicit sarcasm annotations to study how irony and emoji shift sentiment labels. Financial sentiment has long been supported by the Financial PhraseBank and by stock-related Twitter corpora; more recently, \cite{deng2023llms} introduced a Reddit-based market sentiment dataset labeled bullish, bearish, or neutral to address data scarcity in finance. Multilingual product-review datasets (e.g., Arabic, Chinese) broaden coverage across languages, and niche multimodal or “aesthetic” sentiment resources illustrate crossovers with vision-language and aspect-centric judgments.

In the LLM era (2022–2025), classic benchmarks continue to anchor evaluation while new practices and specialized datasets expand the landscape. Studies routinely test zero-/few-shot LLMs on SST-2/SST-5, MR/Polarity, IMDb, Amazon/Yelp, and Twitter/SemEval sets; for example, \cite{zhang2025pushing} report LLM results on SST-2 and MR. Prompt-based evaluation suites\cite{bhargava2025impact} have also appeared, complementing large-scale benchmarks with targeted probes. Meanwhile, domain-specific datasets—especially in finance (e.g., Reddit market sentiment) and in sarcasm-aware settings—highlight persistent challenges that remain even when models perform strongly on standard corpora. Taken together, today’s sentiment datasets span binary, ternary, and five-class labeling; phrase- and aspect-level annotation; multiple domains and languages; and scales from thousands to over a million instances, providing comprehensive coverage for training and assessing LLM-based sentiment analysis.

\subsection{LLM methods of Sentiment Analysis}

Large language models (LLMs) have reset expectations for sentiment analysis, yet recent evaluations urge caution. Broad benchmarking shows LLMs do well on simpler settings but lag on complex tasks requiring structured outputs or deeper inference, motivating a more realistic evaluation agenda \cite{zhang2023sentiment}. Evidence suggests LLMs exhibit basic sentiment sensitivity but struggle with strong polarity extremes, sarcasm, and irony \cite{liu2024large}. Mechanistically, sentiment appears to lie along a largely linear direction in activation space; causal ablations at key tokens (e.g., commas) degrade zero-shot accuracy, offering both interpretability and a warning about brittleness \cite{tigges2023linear}. Fairness audits reveal persistent social biases despite fine-tuning, especially around age, underscoring the need for bias testing in deployment \cite{radaideh2025fairness}. Surveys synthesize these trade-offs across domains, noting computational cost, domain sensitivity, and ethics as recurring themes, and framing finance as a distinctive setting where “what is financial sentiment” itself demands care \cite{gautam2025survey}\cite{mahendran2025comparative}\cite{kirtac2025large}.

\paragraph{Few-shot Prompting} Few-shot prompting is label-efficient but variably reliable. Techniques such as SuperICL and bootstrapping strengthen generative LLMs for financial news, producing stable, explainable signals that improve portfolio construction \cite{mun2025leveraging}. Complementarily, AI-generated exemplars can aid context extraction, though their benefits depend on prompt design and task complexity \cite{ahmed2024leveraging}. Evidence across domains shows the data regime is pivotal: in data-scarce software engineering corpora, larger LLMs achieve zero-shot SOTA, whereas with sufficient labels, fine-tuned small LMs retake the lead \cite{zhang2025revisiting}. Zero-shot multilingual ABSA remains challenging; leaner prompts often outperform elaborate self-consistency or self-debate strategies, especially in English \cite{Wu_2025}. In Chinese healthcare ABSA, compact sLLMs are competitive and efficient, follow instructions well, and support privacy-preserving deployment \cite{YangLGPWYI25}. Detailed prompts help zero/few-shot ABSA but become less critical after fine-tuning \cite{abs-2310-18025}. Overall, few-shot methods offer speed and low labeling cost but can be unstable across domains and remain vulnerable to irony and subtle cues \cite{zhang2023sentiment}\cite{liu2024large}.

\paragraph{Chain-of-Thought Reasoning} Reasoning-oriented prompting can improve reliability and transparency but introduces latency and new hallucination pathways. In finance, Domain Knowledge Chain-of-Thought (DK-CoT) integrates domain expertise with CoT, boosting robustness and weighted F1 for news sentiment \cite{ChenLZZ25}. For weak supervision, Reddit pipelines pair CoT with multiple reasoning paths to stabilize weak labels and train efficient downstream models \cite{deng2023llms}. For policy analytics, multi-task reasoning frameworks jointly infer travel modes, sentiments, and rationales from tweets, enabling insights without manual labels \cite{abs-2411-02666}. Nevertheless, in multilingual ABSA, complex self-improvement or self-debate prompts do not consistently outperform simple zero-shot baselines, indicating diminishing returns from heavier reasoning prompts across languages \cite{Wu_2025}.

\paragraph{Fine-Tuning and Ensemble Approaches} Fine-tuning, ensembling, and continual learning deliver durable gains when compute and data allow. On the high end, fine-tuned GPT-3.5 sets SOTA on SemEval-2014 ABSA, albeit at higher cost \cite{abs-2310-18025}. On the efficiency frontier, compact models excel on the speed–accuracy trade-off: EmoBERTTiny surpasses 7B-chat LLM baselines with millisecond inference, making it well suited for real-time use \cite{StigallKANP24}. For ensembling, RGPT adaptively reweights hard instances and aggregates historical predictions to boost specialized LLM classifiers, outperforming SOTA LLMs and even average human performance on multiple benchmarks \cite{zhang2025pushing}. For non-stationary settings, continual learning with domain-decoupled adapters preserves prior knowledge while acquiring new domains and performs domain positioning at inference without explicit IDs, achieving SOTA across 19 ABSA datasets \cite{abs-2405-05496}. In finance, a domain-specific LLaMA-2 paired with summarization of long filings improves return prediction and robustness, and even lifts traditional models \cite{CHIU2025102632}. Cross-model comparisons suggest GPT-4 generalizes best, FinBERT excels on structured financial text, and T5 lags in recall/generalization \cite{seshakagari2025dynamic}, echoing reviews that highlight FinBERT’s reliability and the cost–benefit calculus of task-specific fine-tuning \cite{mahendran2025comparative}\cite{zhang2025revisiting}.

\paragraph{LLMs for Data Augmentation/Annotation} LLMs are effective labelers and data generators. In low-resource ABSA, few-shot prompting to synthesize annotations raises F1, especially for aspect-level sentiment with modest seed data \cite{HellwigFW25}. At scale, weak-labeling pipelines provide tractable supervision: CoT-stabilized Reddit market sentiment \cite{deng2023llms}, GPT-4–labeled Baijiu stock forums followed by LLaMA fine-tuning \cite{ZhuangWCH25}, and multi-LLM majority voting for large social studies \cite{WANG2025115802}. In health-related social media, domain-aware prompting and targeted fine-tuning outperform lexicon baselines yet still fall short of high accuracy, motivating hybrid workflows and practical prompting guidance \cite{He2024.03.19.24304544}. Overall, the advantages are rapid coverage and flexible domain adaptation; the risks are label noise, ethical and copyright constraints, and amplification of existing biases \cite{deng2023llms}\cite{radaideh2025fairness}.

\paragraph{Handling Nuances and Context} Handling nuance and context often calls for architectural or retrieval enhancements. Retrieval-augmented LLMs ground instructions in external evidence, mitigating pretraining–task mismatch and short-text context gaps, and thereby improving financial sentiment accuracy \cite{zhang2023enhancing}. Summarization layers condense long filings into analysis-ready text, boosting sentiment fidelity and return predictiveness \cite{CHIU2025102632}. Multi-source fusion of Twitter/news sentiment with dynamic asset models improves both short- and long-horizon stock forecasts \cite{ShaoYCWG25}. For trading, LLM-based, sentiment-driven strategies with improved prompting deliver profitable, stable performance with explainable rationales \cite{mun2025leveraging}, and financial DK-CoT further stresses cost-effective reliability and class-weighted metrics given asymmetric downside risks \cite{ChenLZZ25}. Nonetheless, sarcasm, brevity, and domain shifts remain difficult \cite{liu2024large}, multilingual ABSA is brittle \cite{Wu_2025}, and the linear sentiment mechanism is both a lever for control and a single-point vulnerability—motivating more realistic benchmarks and protocols \cite{tigges2023linear}\cite{zhang2023sentiment}.

A practical recipe emerges. When labels are scarce, start with few-shot prompting or weak-label pipelines; add CoT/DK-CoT for interpretability and stability in finance or policy analyses. Under tight latency or privacy budgets, prefer compact or small LLMs fine-tuned for the domain. For shifting domains, adopt continual learning with decoupled adapters and retrieval augmentation. In finance, combine summarization with domain knowledge to translate sentiment into actionable returns. At every stage, integrate fairness auditing and robust evaluation practices to counter bias and over-optimism, guided by domain surveys and reviews that articulate the cost, robustness, and ethical trade-offs intrinsic to LLM-based sentiment analysis.

\subsection{Key Challenges of Sentiment Analysis}

Despite the significant progress enabled by LLMs in sentiment analysis, several key challenges persist. One prominent challenge is the accurate identification and classification of neutral sentiment. Neutral statements often lack explicit emotional cues or may contain a mix of positive and negative aspects that cancel each other out, making them difficult for models to categorize correctly. For example, a statement like "The product arrived on time" is factual and neutral, but models might incorrectly assign a positive sentiment if they overemphasize words like "on time" without considering the overall neutrality of the expression. The ResearchGate article on BERT applications specifically points out that the detection of neutral reviews is a problem impacting model accuracy. This difficulty is compounded when neutral expressions are subtle or when the model is trained on imbalanced datasets where neutral examples are underrepresented. Improving the model's ability to distinguish between genuinely neutral content and weakly positive/negative content remains an active area of research. This often involves curating more balanced datasets, developing more sophisticated feature representations, or employing techniques that specifically target the nuances of neutral language.

Another significant challenge is the subjectivity and inherent ambiguity in human language, which directly impacts sentiment analysis. Sentiment is not always explicitly stated and can be conveyed through sarcasm, irony, figurative language, or cultural context, all of which are difficult for models to interpret accurately. For instance, a statement like "Great, another Monday!" might be interpreted as positive by a naive model focusing on the word "great," while a human would easily recognize the negative sentiment conveyed through sarcasm. The inherent ambiguity means that even human annotators may disagree on the sentiment label for a particular piece of text, leading to noisy training data and affecting model performance. The subjective nature of sentiment also means that what one person perceives as positive, another might see as negative or neutral, depending on their personal experiences, beliefs, and cultural background. This variability makes it challenging to create universally applicable sentiment analysis models. Addressing this requires not only more sophisticated models but also a deeper understanding of pragmatics and context, potentially through the integration of commonsense knowledge and world knowledge into LLMs.

Finally, the presence of false or deceptive reviews in datasets poses a considerable challenge to the accuracy and reliability of sentiment analysis models. On many online platforms, particularly e-commerce and review sites, businesses or individuals may post fake positive reviews to boost their own reputation or fake negative reviews to damage a competitor's. These deceptive reviews are often crafted to mimic genuine expressions of sentiment, making them difficult for automated systems to detect. When models are trained on datasets contaminated with such false reviews, they can learn incorrect associations and produce unreliable sentiment predictions. The ResearchGate article suggests that future research could focus on constructing false review categorization models to mitigate this issue. This involves developing techniques to identify and filter deceptive content before it is used for training sentiment analysis models, or to build models inherently more robust to such noise. This is critical for applications where sentiment analysis supports decision-making, as predictions based on manipulated data can lead to erroneous conclusions and unfair outcomes.

In summary, LLMs bring excellent generalization and an ability to incorporate context and world knowledge into sentiment analysis. With the right prompting, they can even outperform some task-specific models, especially in zero or few-shot settings. Fine-tuning and advanced prompting further close the gap for hard cases, making LLMs the new state of the art for sentiment analysis in many evaluations. Yet, ensuring they correctly handle tricky linguistic phenomena remains an active research challenge. The lessons learned in sentiment analysis – about prompting, augmentation, and hybrid deployment – carry over to other subjective tasks, as we explore next.

\section{Emotion Recognition}

\subsection{Task Definition of Emotion Recognition}

Emotion recognition in NLP is the task of identifying and classifying the emotional state expressed in textual data. This involves going beyond positive or negative sentiment to discern specific emotions such as joy, sadness, anger, fear, surprise, or disgust, and often nuanced or complex emotional states. The task is crucial for enabling machines to understand and respond to human emotions more effectively, enhancing human-computer interaction. The core challenge lies in the subjectivity and ambiguity of human emotions, which are influenced by context, cultural background, and individual differences. Emotion recognition systems aim to analyze textual cues, such as word choice, sentence structure, and punctuation, to infer the emotional tone. This process often involves feature engineering techniques, such as extracting n-grams or using lexicons, or more advanced deep learning models that can capture contextual information and semantic nuances. The ultimate goal is to develop models that can accurately interpret the emotional content of text, enabling applications in areas like customer feedback analysis, mental health monitoring, and empathetic conversational AI.

The definition of emotion recognition tasks can vary depending on the specific application and the granularity of emotions being considered. Some approaches focus on a small set of basic emotions, while others aim for a more fine-grained classification, including complex or ambiguous emotions. For instance, in conversational AI, emotion recognition is often applied at the utterance level within a dialogue, requiring an understanding of how emotions evolve and interact between speakers. This involves not only recognizing the emotion in a single utterance but also considering the conversational history and speaker dependencies. Furthermore, the task can be formulated as single-label classification (assigning one primary emotion to a text segment) or multi-label classification (assigning multiple relevant emotions if the text expresses a blend of feelings). The complexity of human emotions means that a single piece of text might evoke different emotional interpretations from different annotators, highlighting the challenge of achieving high inter-annotator agreement and the need for models that can handle this ambiguity. Therefore, a comprehensive task definition for emotion recognition must consider the scope of emotions, the unit of analysis (e.g., word, sentence, document, utterance), and the potential for multiple or blended emotional states.

\subsection{Dataset of Emotion Recognition}

Progress in text-based emotion recognition has been driven by diverse, well-annotated datasets that vary in source, label scheme, and the degree of context they provide. Foundational resources established basic taxonomies and tasks: \textbf{ISEAR} collects self-reported experiences across seven emotions; \textbf{Emotion-Stimulus} links narrative sentences to emotions and their causes; and early social media corpora such as \textbf{TEC}, \textbf{CrowdFlower/Appen}, and domain-specific sets like \textbf{Electoral Tweets} support broad coverage of tweet-level emotion. Dialog-oriented corpora such as \textbf{DailyDialog}, \textbf{EmotionX}, and spoken or multimodal benchmarks like \textbf{IEMOCAP} and \textbf{MELD} helped crystallize Emotion Recognition in Conversations, where context and speaker identity shape interpretation. Dimensional perspectives are captured by \textbf{EmoBank} with VAD scores, complementing categorical labels. Within this landscape, the \textbf{SemEval 2018} “Affect in Tweets” shared task standardized multilingual benchmarks for both emotion classification and intensity (regression and ordinal), helping to consolidate practical evaluation protocols. Beyond news and social media, literary corpora with sentence- or character-level annotations have supported studies of narrative affect. These datasets also exposed core challenges—multi-label co-occurrence and low inter-annotator agreement—leading to calls for multi-perspective labeling and quality controls.

The modern large-scale era is anchored by GoEmotions\cite{demszky2020goemotions}, a 58k-comment Reddit corpus with 27 categories plus Neutral that enabled fine-grained classification and robust transfer. It has become a default benchmark for LLMs and smaller fine-tuned models alike. In dialogue and empathy-focused settings, EmpatheticDialogues\cite{rashkin2018towards} (32 emotions) supports empathetic response modeling, while MELD and IEMOCAP remain central for contextual and multimodal ERC. Specialized tasks expanded the scope: RECCON\cite{poria2021recognizing} targets emotion cause extraction within conversations, and Affect in Tweets from SemEval-2018 remains a standard for intensity modeling. Together, these corpora benchmark discriminative classification, intensity estimation, and causal reasoning across single-utterance and contextual settings.

Recent work (2022–2025) emphasizes context-rich, task-oriented, and ambiguity-aware evaluation aligned with LLM capabilities. EmoWOZ\cite{feng2021emowoz} introduces emotional variation in task-oriented dialogues, probing whether systems detect shifts such as anger versus neutrality in service conversations. New evaluations target uncertainty and mixed affect; \cite{hong2025aer} propose an Ambiguous Emotion Dataset with high annotator disagreement to test whether models can recognize uncertain or blended emotions, and report results across multiple standard datasets. Surveys such as \cite{chen2024recent} document a broadening agenda from classification to emotionally aware response generation and Theory-of-Mind assessments, while reinforcing that discriminative emotion recognition remains a foundational testbed. Across this landscape, datasets differ in granularity (categorical vs. dimensional), domain (tweets, Reddit, dialogues, narratives), and annotation scheme (single- vs. multi-label, intensity, cause), and many are used to probe LLMs’ strengths and weaknesses under context, subjectivity, and ambiguity. At the same time, multimodal image/video resources, together with ERC counterparts such as MELD, underscore an ongoing trend toward richer, context-sensitive, and comprehensive evaluation.

\subsection{LLM methods of Emotion Recognition}

\paragraph{Prompting and Adaptive Emotional Reasoning} Prompt- and reasoning-centric approaches use lightweight controls to elicit latent affective abilities in general-purpose LLMs. EmotionPrompt injects affective cues into instructions, improving induction-style tasks, broad capability benchmarks, and human-rated generative quality \cite{li2023large}. Emotional Chain-of-Thought aligns intermediate reasoning with human emotional guidelines, increasing harmlessness and positivity \cite{li2024enhancing}. Task-adaptive long reasoning (Emotion-o1) adjusts chain length to difficulty and jointly rewards accuracy, depth, diversity, and logical consistency, improving advanced affective tasks such as sarcasm detection \cite{song2025emotion}. For noisy speech–text pipelines, Revise–Reason–Recognize combines emotion-specific prompts (acoustic, linguistic, psychological) with ASR correction to maintain robustness \cite{li2025revise}. Training-free in-context learning that pairs image-similarity retrieval with chain-of-thought enables context-aware visual emotion understanding without retraining \cite{lei2024large}. Reinforcement learning with verifiable rewards improves explainability, accuracy, and out-of-distribution robustness for omnimodal emotion recognition, while attributing modality contributions \cite{zhao2025r1}.

\paragraph{Instruction Tuning and Parameter-Efficient Specialization} Parameter-efficient customization enables cost-effective specialization of LLMs for affective computing. DialogueLLM performs instruction tuning on multimodal dialogues and injects visual context as knowledge, reaching state-of-the-art results on emotion recognition in conversation (ERC) with modest compute \cite{zhang2023dialoguellm}. Adapter-based approaches such as P-Tuning v2 and LoRA allow LLMs to surpass dedicated baselines across multiple emotion datasets, demonstrating transferability and efficiency \cite{peng2024customising}. In low-resource settings, knowledge-augmented few-shot learning that couples contrastive embedding training with prompt-based self-prediction enhances sentiment and affect analysis \cite{yang2024advancing}. For multi-label scenarios, ambiguity-aware prompting enables reliable modeling of overlapping emotions, particularly when dialogue and speech cues are available \cite{hong2025aer}. Extending beyond recognition, fine-tuned LLMs can infer emotion-regulation strategies from observed behavior—outperforming Bayesian baselines even without post-interview data—highlighting potential for coaching and therapeutic applications \cite{muller2024recognizing}. Complementarily, cross-context fusion with LoRA and targeted domain adaptation further advances continuous affect prediction in challenging multimodal benchmarks \cite{yi2023exploring}.

\paragraph{Multimodal and Omni-Modal Emotion Integration} Multimodal instruction tuning is accelerating perception-rich emotional intelligence in LLMs. Emotion-LLaMA integrates audio–visual–text encoders with instruction tuning on MERR to couple recognition with reasoning, achieving state-of-the-art results across multiple corpora and zero-shot video settings \cite{cheng2024emotion}. Omni-Emotion advances video MLLMs with fine-grained facial and acoustic modeling, including micro-expressions, and curates high-quality, human-reviewed datasets for both recognition and explanation \cite{yang2025omni}. AffectGPT combines pre-fusion multimodal alignment with training on MER-Caption to support open-vocabulary emotion captioning, evaluated via MER-UniBench \cite{lian2025affectgpt}. On the visual front, EmoVIT pioneers affect-oriented visual instruction tuning by generating emotion-specific instructions, excelling at classification and affective reasoning \cite{xie2024emovit}. Face-centric EMO-LLaMA further leverages instruction data and facial priors (global/local features, demographics) to deliver SOTA-comparable FER, covering micro-expressions and audio–vision fusion \cite{xing2024emo}. For speech-centric applications, EMOVA provides end-to-end omni-modality with disentangled speech tokenization and controllable style for expressive spoken dialogue \cite{chen2025emova}, while SECap moves beyond discrete labels to natural-language emotion captions through HuBERT and Q-Former interfaces to LLaMA \cite{xu2024secap}. From a privacy standpoint, DEEMO demonstrates strong recognition and reasoning using de-identified video/audio and non-facial body cues, reducing identity exposure \cite{li2025deemo}. For compositional affect, LVLMs adapted via two-stage tuning—basic emotions followed by compound optimization—achieve SOTA with zero-shot generalization \cite{yu2025compound}, and remain effective for context-aware detection using training-free retrieval or light fine-tuning \cite{lei2024large}. Complementary visual–affective modeling with multi-perspective projection and EmoPrompt further strengthens nuanced emotion reasoning in MLLMs \cite{yang2024emollm}.

\paragraph{Benchmarking Emotional Intelligence: Strengths, Gaps, and Risks} A rapidly maturing evaluation ecosystem is revealing strengths, limitations, and risks. Grounded in psychological theory and spanning English and Chinese, EmoBench assesses emotion understanding and application, showing LLMs—despite progress—still fall short of average human performance \cite{sabour2024emobench}. EmoBench-M extends evaluation to multimodal settings and indicates MLLMs continue to lag humans on core emotional intelligence scenarios \cite{huemobench}. From a reliability perspective, EmotionHallucer audits “emotion hallucinations” across perceptual and knowledge dimensions, finds widespread errors, and proposes a mitigation framework with measurable gains \cite{xing2025emotionhallucer}. With respect to empathy, EmotionQueen probes key, mixed, and implicit events as well as intention recognition, revealing strong performance on explicit tasks but persistent limits for implicit affect \cite{chen2024emotionqueen}. For multimodal emotion understanding, MER-UniBench offers a dedicated benchmark \cite{lian2025affectgpt}, while AEB and the EmoLLMs suite standardize multi-task affective evaluation and annotation \cite{liu2024emollms}. Large-scale comparisons further show that LLMs can surpass humans in empathy ratings, though performance varies by emotion \cite{welivita2024large}; psychometric tests often place top LLMs at or above average human EQ, yet with mechanisms distinct from human reasoning and uneven skill profiles across EI branches \cite{wang2023emotional}\cite{vzorinab2024emotional}. In image-only emotion recognition, specialized CNNs still hold a slight edge over general LLMs, although LLMs remain practical under data scarcity \cite{nadeem2024vision}. Collectively, these benchmarks chart clear progress while highlighting open challenges in multimodality, implicit emotion inference, and hallucination control.

\paragraph{Interactive Systems and Emotional Support} End-user systems highlight promise and caveats. Embodied “Virtual Humans” couple LLMs with realistic avatars and explicit psychological constructs (e.g., personality, mood) to steer affective valence in semi-guided dialogue, achieving high naturalness and realism, although arousal control remains difficult \cite{llanes2024developing}. In child-facing settings, a state machine–guided chatbot elicits sharing of personal events and emotions and is perceived as a “close friend” in laboratory studies \cite{seo2024chacha}. For psychotherapy support, fusing emotion-aware embeddings with LLMs and retrieving context from sessions improves empathy, coherence, and fluency \cite{rasool2025emotion}. Beyond clinical use, LLM-generated arguments are as persuasive as human ones and show cognitive effort and moral language—implications for civic education and risks of misinformation \cite{carrasco2024large}. Nonetheless, in emotional-support conversations current LLMs exhibit strategy biases and miscalibrated preferences that hinder effectiveness, underscoring the need for external oversight and bias mitigation before reliable deployment \cite{kang2024can}. Methodologically, ambiguity-aware prompting and behavior-level strategy recognition broaden coverage of real-world affective ambiguity and regulation \cite{hong2025aer}\cite{muller2024recognizing}.

\paragraph{Mechanistic Affective Modeling and Causal Extraction} Mechanistic and theory-driven accounts are increasingly aligning model behavior with affective science. At the representation level, converging evidence for emotion-selective neuron groups in LLMs—together with ablation studies showing cross-layer compensation—implies distributed, model-dependent circuitry for affect \cite{lee2025large}. From a cognitive-science perspective, a comprehensive survey maps advances in LLM-based emotion cognition onto the sensation–perception–attention pipeline and situates techniques ranging from contrastive learning to theory-of-mind-style reasoning \cite{chen2024recent}. For causal explanation, emotion-cause triplet extraction benefits from multimodal, multi-scale heterogeneous graphs that foreground causal context and utterance-pair communication \cite{liang2025m3hg}. On the data side, hybrid human–AI labeling pipelines show GPT-4 can flag low-quality annotations, improving reliability and efficiency in affect datasets while remaining perceptually distinct from human raters \cite{niu2025rethinking}. Taken together with audits of hallucination and generalization, these strands chart paths toward theory-grounded, trustworthy emotional intelligence, including improved handling of ambiguity and regulation \cite{zhang2024refashioning}\cite{hong2025aer}\cite{muller2024recognizing}.

Overall, emotion recognition with large models now follows a coherent trajectory: lightweight prompting delivers immediate gains; instruction tuning with parameter-efficient adapters consolidates capability; multimodal and omni-modal architectures capture real-world signals; and theory-grounded benchmarks surface blind spots in implicit understanding and hallucination. Coupling explainable reinforcement learning, privacy-preserving design, and hybrid human–AI annotation improves reliability, while retrieval grounding and bias mitigation safeguard sensitive deployments. Selecting methods by data budget, latency, and modality turns a diverse toolkit into a robust, accurate, and responsible strategy for affective AI.

\subsection{Key Challenges of Emotion Recognition}

Despite significant progress, emotion recognition using LLMs faces several key challenges, primarily stemming from the inherent nature of human emotions and the limitations of current models. One of the most prominent challenges is the ambiguity and subjectivity of emotional expression. Human emotions are complex and often nuanced, making it difficult to assign a single, definitive label. Textual expressions can be interpreted differently by different individuals or even by the same individual under varying contexts. This is reflected in low inter-annotator agreement in many emotion datasets. LLMs, while powerful, can struggle to capture this inherent ambiguity, often providing a single emotion label that might not fully represent the subtlety of the expressed feeling. Forcing complex, mixed emotions into discrete, predefined categories is an oversimplification that can lead to models trained on an incomplete or skewed representation of reality. The AER-LLM study specifically addresses this by focusing on recognizing ambiguous emotions, but it remains a core challenge for the field. The subjective experience of emotion is also shaped by cultural and individual factors, which are often not adequately accounted for in datasets or models.

Another significant challenge is the context-dependency of emotions. The meaning of a word or phrase, and the emotion it conveys, can drastically change depending on the surrounding text, the speaker's intent, and the broader situational context. While LLMs are adept at capturing some level of context, fully understanding long-range dependencies and subtle contextual cues that disambiguate emotional meaning remains difficult. For example, sarcasm, irony, or humor can completely invert the apparent emotional valence of a statement, and detecting these figurative language uses is a challenge in itself. Furthermore, the reliability and bias in datasets pose a major hurdle. Many emotion recognition datasets are created using majority voting from multiple annotators, which can obscure the inherent ambiguity and lead to a "flattened" representation of emotions. Datasets may also suffer from biases related to the demographics of the annotators or the sources of the text (e.g., specific social media platforms), leading to models that perform well on similar data but generalize poorly to new domains or populations. The "observer effect," where the act of being monitored alters a user's emotional expression, can also compromise data fidelity, especially in real-world settings.

The interpretability and explainability of LLM-based emotion recognition systems also present challenges. While LLMs can achieve high accuracy, understanding why a model made a particular emotional prediction is often difficult. This "black box" nature can be problematic, especially in sensitive applications like mental health monitoring or human-robot interaction, where trust and transparency are crucial. Research into "emotion neurons" attempts to shed light on the internal representations of emotions within LLMs, but this is still an emerging area. Moreover, ethical considerations are paramount. The deployment of emotion recognition systems raises concerns about privacy, surveillance, and the potential for misuse, particularly if the systems are not robust or fair across different demographic groups . Ensuring that these technologies are developed and deployed responsibly is a critical ongoing challenge. Finally, resource limitations for low-resource languages and the computational cost of training and deploying large LLMs can also hinder the widespread adoption and further development of sophisticated emotion recognition systems . Addressing these multifaceted challenges requires interdisciplinary collaboration and continued innovation in model architecture, dataset creation, and evaluation methodologies.

The successes in emotion recognition lay a foundation for tackling other subjective phenomena. One such phenomenon that heavily intersects with sentiment and emotion is sarcasm, which we discuss next, as it often flips sentiment and adds complexity to emotion recognition tasks.

\section{Sarcasm Detection}

\subsection{Task Definition of Sarcasm Detection}

Sarcasm detection is a specialized task within natural language understanding that focuses on identifying whether a piece of text is intended to be sarcastic. Sarcasm is a form of figurative language where the speaker or writer says the opposite of what they truly mean, often for humorous, ironic, or critical effect. The core challenge in sarcasm detection lies in the discrepancy between the literal meaning of the words and the intended, often opposite, meaning. This discrepancy is signaled through contextual cues, tone of voice (in spoken language), linguistic patterns, or shared knowledge between the communicator and the audience. For example, the statement "Oh, great, another meeting!" is likely sarcastic if the speaker is known to dislike meetings or if the context suggests a negative sentiment towards meetings. Accurately detecting sarcasm is crucial for a deeper understanding of sentiment and opinion, as misinterpreting a sarcastic statement as literal can lead to a misunderstanding of the speaker's intent. The task is particularly relevant in analyzing social media text, product reviews, and online discussions, where sarcasm is frequently employed. The output of a sarcasm detection system is typically a binary label (sarcastic or not sarcastic), although some approaches may attempt to identify the target of the sarcasm or its underlying sentiment.

\subsection{Dataset of Sarcasm Detection}

Research on sarcasm and irony detection has been driven by datasets drawn largely from social media and curated benchmarks. Early Twitter corpora collected via distant supervision with hashtags such as \#sarcasm (e.g., the widely used \textbf{Ptáček Twitter corpus}) established a scalable but noisy paradigm. \textbf{SemEval-2018 Task 3} on Irony Detection in English Tweets standardized evaluation with both binary and fine-grained labels (e.g., Non-irony, Verbal Irony, Situational Irony), and many studies use its binary setting for comparability. Outside social media, the \textbf{News Headlines dataset} contrasts satirical headlines from The Onion with genuine headlines from HuffPost, offering a style-specific but domain-limited benchmark frequently used to test transfer.

Context-rich and multimodal resources broadened coverage. The Reddit Self-Annotated Corpus (\textbf{SARC})\cite{khodak2018corpus} leverages user markers (e.g., “/s”), provides both balanced and large unbalanced splits with hundreds of thousands of comments, and supplies conversational structure (parent/child and thread hierarchy) critical for pragmatic cues. CASCADE\cite{zhou2025cascade} focuses explicitly on dialogue context by labeling the final utterance in multi-turn discussions as sarcastic or not. MUStARD\cite{castro2019towards} and its extension MUStARD++\cite{bhosale2023sarcasm} compile sarcastic and non-sarcastic dialogue snippets from TV shows with aligned video/audio and transcripts; despite being multimodal, their text transcripts are widely used for text-only experiments targeting conversational sarcasm.

Recent datasets emphasize annotation quality, multilinguality, and speech. \textbf{SemEval-2022 Task 6 }(iSarcasmEval) addresses the noise of hashtag supervision by asking original authors to annotate intended sarcasm in English and Arabic and to provide literal rephrasings for sarcastic tweets, yielding smaller but high-fidelity pairs that support supervised learning and analysis of meaning contrast. In 2025, PodSarc introduced a large spoken sarcasm benchmark from a podcast, pairing audio with transcripts and using LLM-assisted labels that were human-validated. Concurrently, LLM-based studies have evaluated across “widely used benchmark datasets” spanning tweets, forums, and dialogues (e.g., those above). For instance, \cite{lee2024pragmatic} report state-of-the-art results on SemEval-2018 and MUStARD using text-only prompting, while broader assessments such as multi-agent approaches like \cite{li2025leveraging} underscore the continuing role of these benchmarks.

Together, these datasets cover short quips, threaded conversations, news headlines, and audiovisual dialogue, with diverse annotation methodologies (self-annotation, author intent, expert/crowd labels). They reveal design trade-offs: distant supervision scales but is noisy; author-intent labels and paired rephrasings increase reliability but reduce size; conversational and multimodal context improves ecological validity. This variety enables comprehensive evaluation of sarcasm detectors and LLMs, while highlighting persistent challenges of domain shift, context dependence, and pragmatic nuance.

\subsection{LLM methods of Sarcasm Detection}

\paragraph{Prompted Reasoning for Sarcasm} Pragmatic and metacognitive prompting enrich LLMs’ capacity to infer implied meanings and reconcile contextual mismatches, delivering state-of-the-art performance \cite{lee2024pragmatic}. In a complementary direction, SarcasmCue formalizes cue-centric reasoning—contradictions, graphs, bags, and tensors—showing that non-sequential cue aggregation can boost smaller models, whereas stronger models benefit more from structured chains and graphs \cite{yao2407sarcasm}. Moreover, chain-of-thought rationales aid entity-level sentiment in news and can transfer to sarcasm; however, their effectiveness is inconsistent and improves with self-consistency \cite{kuila2024deciphering}. Fine-tuning still outperforms zero-shot prompting for sarcasm on large GPT variants \cite{gole10sarcasm}, and LLaMA‑3 often succeeds on shorter inputs yet struggles as length increases \cite{mai2024llama}. More broadly, benchmarking indicates that supervised PLMs surpass LLMs on sarcasm \cite{zhang2024sarcasmbench}, and scaling alone does not guarantee pragmatic competence—sarcasm continues to lag metaphor under psychiatric-style probing \cite{yakura2024evaluating}. Additionally, specialized fine-tuning on iSarcasmEval with efficient PEFT/QLoRA yields strong gains, underscoring the importance of target-domain supervision and explicit intention cues \cite{heraldi2024effective}.

\paragraph{Multimodal Understanding and OOD Robustness} Generative, instruction-following multimodal systems that retrieve demonstrations sidestep overfitting-prone fusion stacks and improve out-of-distribution generalization on RedEval while achieving in-domain SOTA \cite{tang2024leveraging}. On the reasoning front, MiDRE blends internal incongruity reasoning with external LVLM rationales and adaptively weights them to outperform prior methods \cite{jana2025think}; DMDP injects deep, modality-disentangled prompts for few-shot settings and cross-dataset generalization \cite{jana2025dual}; and CofiPara first uses LMM-generated rationales to train coarse sarcasm and then targets fine-grained sarcasm entities \cite{lin2024cofipara}. Complementarily, EilMoB extracts emotion-aware textual incongruity from image–text pairs and bridges modalities to exploit cross-source tensions \cite{zhao2025eilmob}. In parallel, agentic VLLM pipelines that triangulate superficial form, semantics, and sentiment consistently lift zero-shot performance on MMSD2.0 \cite{wang2024s3}. However, evaluations expose a seeing–understanding gap: high perceptual accuracy coexists with sizable sarcasm-comprehension errors rooted in pragmatic and affective reasoning deficits \cite{zhang2025mlms}, and explanation quality does not reliably track model scale \cite{amal2025well}. Taken together, a unified benchmark (MHSDB) underscores that robust multimodal fusion and carefully chosen integration strategies are pivotal for nuanced humor and sarcasm \cite{dong2025mhsdb}.

\paragraph{Multi-Agent Orchestration Pipelines} Decomposing the challenge of complex sarcasm into coordinated expert roles enhances both robustness and interpretability. In this paradigm, Commander-GPT dispatches focused sub-tasks—keyword extraction, sentiment estimation, and cross-modal verification—to specialized LLMs and fuses their outputs through a coordinating controller \cite{zhang2025commander}. From a complementary angle, CAF-I formalizes irony via multi-agent collaboration that separates context, semantics, and rhetoric, then aggregates them with a decision agent and iterative feedback \cite{liu2025caf}. Through the lens of deliberation, LDGNet stages debates among LLM “debaters” and employs a learned judge to surface latent world knowledge, producing reliable sentiment decisions across both in-domain and OOD settings \cite{zhou2025ldgnet}. Extending these ideas to audio-only conditions, LLM-guided annotation pipelines with human-in-the-loop gating introduce new speech sarcasm resources (PodSarc) and enable competitive bimodal systems, bringing agentic supervision to low-visibility modalities \cite{li2025leveraging}. Taken together, systematic decomposition, structured interaction, and principled adjudication emerge as a coherent design pattern for pragmatic inference under uncertainty.

\paragraph{Commonsense, Incongruity, and Knowledge Alignment} Commonsense-centered approaches treat emotional incongruity as a primary signal for sarcasm. In this framing, EICR combines retrieval-augmented LLMs, dependency-graph refinement, adaptive reasoning skeletons, and adversarial contrastive learning to isolate sentiment-inconsistent subgraphs while suppressing spurious correlations \cite{qiu2025detecting}. From a temporal perspective, KA-LLM models evolving events by building dynamic knowledge graphs over topic–target pairs and aligning them with hybrid objectives, thereby explaining how sarcasm triggers shift over time \cite{xiang2025dynamic}. On the multimodal front, recent methods extract or synthesize textual “incongruity carriers” to narrow modality gaps—EilMoB’s emotion-aware incongruity modeling and CofiPara’s rationale-guided pretraining exemplify this trend \cite{zhao2025eilmob}\cite{lin2024cofipara}. Complementarily, external rationales generated by LVLMs, though often noisy, supply useful cues that move beyond shallow captions and steer incongruity resolution \cite{jana2025think}. Looking back, earlier contextual paradigms—such as CASCADE’s incorporation of user and discussion features—anticipated today’s knowledge-infused strategies and, at the same time, highlight enduring difficulties with implicit sarcasm that lacks overt cues \cite{zhou2023evaluation}\cite{sabera2025comparative}.

\paragraph{Nuances, Augmentation, and Linguistic Variety} Language nuance remains pivotal. At the domain level, a nuclear-industry study shows topic-specific LLMs struggle with sarcasm while general-domain models perform better; robustness improves with adversarial text augmentation and targeted sarcasm removal, whereas emojis tend to amplify rather than flip sentiment \cite{bhargava2025impact}. Across language varieties, the BESSTIE benchmark finds degraded transfer from inner-circle to outer-circle English—especially for sarcasm—underscoring the need for variety-specific resources and adaptation \cite{srirag2024besstie}. In cross-lingual settings, Indonesian sarcasm experiments indicate that fine-tuned PLMs outperform zero-shot LLMs, and naive augmentation does not remedy class imbalance \cite{suhartono2024idsarcasm}. On the evaluation side, out-of-distribution suites like RedEval and explanation audits expose metric pitfalls—embedding-based scores can assign high similarity to contradictory explanations—calling for more reliable assessment protocols \cite{tang2024leveraging}\cite{amal2025well}. From an optimization standpoint, dynamic adjustment during multi-task fine-tuning (DAO) stabilizes learning across heterogeneous sentiment subtasks, a strategy well-suited to sarcasm’s imbalanced, multi-objective regimes \cite{ding2024dynamic}.

\paragraph{Benchmarks and Evaluation Practices} Recent resources are making evaluation more comprehensive and equitable. In head-to-head comparisons, SarcasmBench assesses LLMs and PLMs across datasets and prompts, finding GPT-4 the strongest among LLMs yet still behind supervised PLMs; moreover, few-shot instruction-only prompting often outperforms chain-of-thought \cite{zhang2024sarcasmbench}. At the multimodal scale, MHSDB standardizes humor and sarcasm evaluation across languages and modalities, with fusion approaches consistently beating unimodal baselines \cite{dong2025mhsdb}. From a sociolinguistic angle, BESSTIE systematically probes English varieties to reveal persistent equity gaps \cite{srirag2024besstie}. Methodologically, MORE-based audits show that automatic metrics can misjudge explanation faithfulness \cite{amal2025well}, while Visual Room tasks disentangle perception from pragmatic comprehension to quantify the “understanding gap” \cite{zhang2025mlms}. Even so, core datasets like SARC 2.0 pol-bal remain valuable for contrasting tuned and zero-shot paradigms across successive GPT generations \cite{gole2024sarcasm}. Historically, Reddit-based studies continue to highlight the centrality of conversational context and user history \cite{zhou2023evaluation}. Meanwhile, new speech corpora such as PodSarc expand modality coverage \cite{li2025leveraging}, and regional resources like IdSarcasm sustain non-English evaluation \cite{suhartono2024idsarcasm}.

\paragraph{Lessons From Sentiment and ABSA} Progress in sentiment analysis offers transferable tools for sarcasm detection. At the modeling level, instruction-tuned financial LLMs show that small supervised instruction sets can imbue numeracy and domain fluency beyond generic chat models \cite{zhang2023instruct}\cite{inserte2024large}, while at the optimization level, dynamic adaptive strategies improve the stability of multi-task fine-tuning across diverse sentiment objectives \cite{ding2024dynamic}. At the reasoning level, chain-of-thought rationales strengthen entity-specific sentiment decisions and, when paired with self-consistency, suggest prompt-engineering routes for sarcasm \cite{kuila2024deciphering}. At the task granularity level, ABSA comparisons highlight domain sensitivity and the value of strong PLMs/LLMs (DeBERTa, PaLM, GPT) for fine-grained aspect judgments—capabilities adjacent to pinpointing sarcasm targets \cite{mughal2024comparative}. In applied contexts, ChatGPT aligns closely with human ABSA in hospitality \cite{agua2025large}, design-aware position encoding enriches generative ABSA with implicit knowledge \cite{han2025design}, and prompt-engineered sentiment analysis can discriminate subtle clinical language in fibromyalgia screening \cite{venerito2024large}. From an evaluation perspective, broader audits indicate that LLMs still lag humans on sentiment, humor, and metaphor, yet remain sensitive to prompt improvements \cite{yazhou2024can}. From a domain perspective, case studies on nuclear discourse show that sarcasm and sentiment intertwine with policy frames, topicality, and stylistic signals, motivating joint modeling and specialized augmentation \cite{kwon2024sentiment}\cite{bhargava2025impact}.

Overall, recent advances in sarcasm detection leveraging Large Language Models (LLMs) reveal a clear trajectory from single-task fine-tuning toward prompt-engineered, reasoning-aware, and multi-agent/multimodal systems. Benchmarks and empirical studies consistently show that purely scaling models does not guarantee pragmatic comprehension, particularly when sarcasm hinges on cultural, contextual, or emotional incongruity. Techniques such as pragmatic metacognitive prompting, structured cue reasoning, commonsense integration, and dynamic knowledge alignment improve robustness, while agentic frameworks and modality-bridging architectures enable richer interpretation across text, image, and audio. Evaluation work underscores persistent gaps across varieties, languages, and OOD scenarios, urging more equitable, context-aware resources. Cross-pollination from sentiment and ABSA research, along with nuanced handling of topic-specific language cues, suggests that future sarcasm detection will benefit from domain adaptation, explicit reasoning steps, and hybrid integration of statistical, commonsense, and multimodal signals to approach human-like interpretive capability.

\subsection{Key Challenges of Sarcasm Detection}

Sarcasm detection remains a challenging task for LLMs due to several inherent difficulties. One primary challenge is the heavy reliance on context. The interpretation of an utterance as sarcastic often depends on a wide array of contextual factors, including world knowledge, shared understanding between interlocutors, the speaker's typical style, and the specific situation. LLMs, despite their extensive pre-training, may still struggle to access and integrate all relevant contextual information, especially if it's not explicitly stated in the immediate text. For example, understanding a sarcastic comment about a recent event requires knowledge of that event. Another significant challenge is the subtlety and variability of sarcastic cues. Sarcasm can be expressed in many different ways, and the cues can be very subtle, such as a slight change in word choice, a particular sentence structure, or even the absence of expected emotional markers. These cues can be difficult for models to learn, especially when they are sparse or overlap with non-sarcastic language patterns. The ambiguity between sarcasm and other forms of figurative language like irony, humor, or hyperbole also poses a challenge. Distinguishing these closely related concepts can be difficult even for humans, and models may misclassify one for the other.

Furthermore, dataset bias and quality are ongoing concerns. Many sarcasm detection datasets are created from specific sources like social media, which may not be representative of sarcasm in other domains or genres. The annotation process itself can be subjective, and inter-annotator agreement is not always high, leading to noisy labels. Sarcasm is also highly culture-dependent; what is considered sarcastic in one culture may not be in another, or the cues might differ. LLMs trained on data primarily from one cultural context may not generalize well to others. The lack of vocal or visual cues in text-based sarcasm is another hurdle. In spoken communication, tone of voice, facial expressions, and body language provide crucial signals for sarcasm. Text-based models must rely solely on linguistic cues, making the task inherently harder. Finally, adversarial attacks, where subtle changes are made to a text to fool a model into misclassifying sarcasm, highlight the brittleness of some current approaches. Addressing these challenges requires continued research into more context-aware, robust, and nuanced LLM architectures and training methodologies.

We now turn to humor detection, a related challenge that overlaps with sarcasm—sarcasm is a form of humor, though not all humor is sarcastic, and both involve non-literal intent—yet it introduces distinct demands on background knowledge and linguistic creativity. These added complexities are precisely what current LLMs are being tested on.

\section{Humor Detection}

\subsection{Task Definition of Humor Detection}

Humor detection is the task of automatically determining whether a piece of text is intended to be humorous, with some variants also rating its degree or categorizing the humor type; broader humor understanding includes explaining why something is funny and generating jokes. The task is challenging because humor is subjective and culturally contingent, often relies on non-literal intent and linguistic creativity—puns, wordplay, sarcasm, incongruity (overlapping with sarcasm and metaphor), exaggeration, and cultural references—and may require external knowledge and context to “get” the joke. Unlike sentiment analysis, cues are subtler and more context-dependent. Applications include improving human–computer interaction through appropriate responses, filtering or recommending humorous content, and analyzing social dynamics in online communities. While post-ChatGPT LLMs can produce jokes, explain simple ones, and often flag humorous intent, truly human-like humor comprehension still demands advanced reasoning and commonsense.

\subsection{Dataset of Humor Detection}

Work on humor detection has relied on short-form, web-scraped resources that seeded early modeling. Widely used collections include the Short Jokes dataset and the 160,000 Jokes dataset from \textbf{Kaggle}, \textbf{Pun of the Day}, a 16k one-liners collection, and large \textbf{Reddit} jokes dumps (e.g., from r/Jokes). These corpora are mostly one-liners or brief anecdotes and are often paired with non-jokes sampled from other sources to form binary detection sets; variants target specific subtypes such as roast/insult humor. Beyond English, multilingual resources emerged, notably the Spanish \textbf{HAHA} (Humor Analysis) datasets at IberLEF (with humor presence and funniness ratings), \textbf{Hinglish puns collections}, and Chinese releases such as \textbf{CHumor 1.0}. Early multimodal and conversational angles came from \textbf{UR-FUNNY} (humor in TED-talk conversations) and \textbf{MUStARD} (multimodal sarcasm, overlapping with humorous cues). Researchers also began to exploit aligned or minimally contrastive pairs to better capture what “makes” text funny, including The Onion satire with human-edited “serious” counterparts (the Unfunny Corpus) and aligned topic-matched joke/non-joke pairs.

From 2020 to 2021, shared tasks consolidated high-quality, carefully annotated benchmarks. \textbf{SemEval-2020 Task 7} introduced Humicroedit, in which single-word edits turn news headlines humorous, supporting pairwise ranking and analysis of humor-inducing transformations. \textbf{SemEval-2021 Task 7} (HaHackathon) released a large English dataset—primarily tweets and short texts—with multiple annotators per item, covering humor detection (yes/no), funniness rating, and offense in humor; the “Humor and Offense” (\textbf{HAHO}) references typically refer to this split. These tasks highlighted subjectivity via dense annotation and established standard evaluation settings. In parallel, \textbf{Twitter hashtag} (\#humor) collections and several Kaggle humor-detection datasets (e.g., \textbf{HahahaClf}) offered additional short-text benchmarks.

Since 2022, datasets have broadened in modality, context, and language while being used to probe large language models (LLMs). Studies have evaluated LLMs (e.g., GPT-3) on SemEval-2021, Humicroedit, Pun-of-the-Day and Hinglish pun sets, and Reddit joke vs non-joke discrimination, often finding humor—especially wordplay—remains challenging. Conversational and workplace-context corpora (e.g., \textbf{WRIME} and other dialogue resources, including dinner-party dialogues with humorous turns) test whether models recognize humor in context and alongside social variables such as appropriateness and offense. Aligned-pair designs gained traction: beyond Humicroedit, satire–serious headline pairs and other minimally contrastive text pairs have been shown to help models learn portable humor cues, with reports that classifiers trained on such pairs generalize well across datasets. These efforts underscore how dataset choices—humor type (puns, satire, insults), text length, conversational context, and offensiveness—strongly shape detection performance.

Recent directions extend humor detection into new settings and modalities. Targeted datasets for emotionally supportive dialogues have been proposed\cite{quan2025can}, separating humor generation in a specified style from humor recognition in context (e.g., recognizing a counselor’s gentle joke and its appropriateness). In vision-and-language, HumorDB\cite{jain2024ai} has been introduced for graphical humor, providing images (cartoons, photos) labeled for funniness, funniness ratings, and minimally contrastive pairs that differ only in a humor-bearing element; initial results show vision–language models perform above chance yet below human levels. Parallel “co-creativity” evaluations—such as AI-assisted meme captioning—link detection to generation and collaborative use. Together with established resources like Kaggle short-joke corpora, Reddit jokes, Pun of the Day, Humicroedit, UR-FUNNY, MUStARD, SemEval-2021, HAHA (Spanish), Hinglish puns, CHumor 1.0, WRIME, and aligned satire–serious pairs, this expanding ecosystem supports robust, comparative evaluation of humor detection across text, dialogue, and images.

\subsection{LLM methods of Humor Detection}

\paragraph{Benchmarking and Dataset-Driven Approaches} Large-scale, high-quality datasets remain the foundation for advancing humor recognition. Chumor 1.0 and 2.0 \cite{he2024chumor}\cite{he2024chumor}\cite{he2025chumor}, sourced from Ruo Zhi Ba, target culturally nuanced Chinese humor, revealing state-of-the-art (SOTA) LLMs perform only slightly above chance, with human explanations far superior. TalkFunny\cite{chen2024talk} extends this by capturing explainable humor responses with chain-of-humor annotations, enabling evaluation of conversational humor comprehension. HumorBench\cite{narad2025llms} and HumorDB\cite{jain2024ai} introduce English cartoon-caption and visual humor datasets, respectively, exposing persistent performance gaps between human and model understanding. MHSDB\cite{dong2025mhsdb} and YesBut\cite{hu2024cracking} widen scope to multimodal/multilingual humor, showing fusion of modalities consistently outperforms unimodal baselines, but contradiction-based narrative humor remains elusive. These benchmarks underscore that data realism, cultural specificity, and multi-modality are crucial for robust humor understanding evaluation.

\paragraph{Prompting and Few-Shot Baselines} Even without task-specific fine-tuning, prompt-based strategies provide informative baselines for humor detection. From a low-resource perspective, few-shot prompting of GPT-4 and Gemini on Croatian tweets \cite{bago2025few} yields LLM–human agreement on par with human–human agreement, indicating feasibility for rapid dataset bootstrapping. At the modality level, multimodal prompting that incorporates speech audio \cite{baluja2024text} improves explanations of phonetic humor by recovering prosodic cues that purely text-based models miss. In workplace settings, \cite{shafiei2025not} shows that current LLMs misjudge contextual appropriateness, underscoring the gap between surface-level humor detection and situational awareness. These studies position prompting as a lightweight entry point to humor recognition while revealing the brittleness of context-sensitive judgments.

\paragraph{Fine-Tuning and Representation Learning} When domain alignment is critical, supervised adaptation surpasses prompting. From a task-alignment perspective, CYUT’s CLEF JOKER submission \cite{wu2024humour} fine-tuned LLaMA 3 and RoBERTa for humor-genre classification, outperforming zero-shot GPT-4, though test-set generalization lagged. From a representation-learning angle, \cite{epron2024orpailleur} leveraged hidden LLM representations with cross-validation to achieve competitive accuracy without fine-tuning, while noting ambiguous class boundaries. In language-specific settings, specialized Chinese humor models such as CFunModel \cite{yu2025cfunmodel} integrate multi-task learning on aggregated humor corpora, surpassing general-purpose LLMs on recognition benchmarks. Taken together, these studies suggest that controlled adaptation improves humor sensitivity, provided overfitting is managed.

\paragraph{Cultural, Linguistic, and Translation-Aware Methods} Humor recognition becomes more challenging when linguistic or cultural gaps arise. From a cross-lingual perspective, Jokes or Gibberish? \cite{pituxcoosuvarnjokes} examines humor preservation in English–Thai translation and finds that explanation-augmented prompting (GPT-Ex) yields the highest joke retention, particularly for idioms and cultural references. From a cross-cultural angle, \cite{guo2025cross} quantifies humor intensity in Chinese and English family jokes via ambiguity, sentiment, and incongruity indicators, revealing divergences in humor structure. In slang-heavy Chinese contexts, DuanzAI \cite{rohn2024duanzai} boosts LLM comprehension of slang-based humor through phonetic matching and pinyin–hanzi disambiguation. These approaches highlight the value of embedding cultural and linguistic priors into recognition systems for culturally situated humor.

\paragraph{Multimodal Humor Understanding} Humor often spans text, images, and audio, necessitating multimodal processing. From a representation perspective, ClassicMemes-50-Templates \cite{deng2025large} and MemeMind \cite{booij2025mememind} address meme classification and explanation using vision–language embeddings. From the knowledge integration angle, BottleHumor \cite{hwang2025bottlehumor} leverages the information bottleneck to iteratively distill relevant world knowledge for multimodal humor explanation. In terms of fusion strategies, MHSDB \cite{dong2025mhsdb} and YesBut \cite{hu2024cracking} show that multimodal feature fusion outperforms unimodal approaches; however, models still struggle to comprehend implicit contradictions conveyed in visuals.

\paragraph{Safety, Ethics, and Robustness} Beyond accuracy, humor recognition intersects with safety and ethical concerns. HumorReject\cite{wu2025humorreject} fine-tunes LLMs to respond to harmful prompts with indirect refusals, decoupling safety from denial templates. \cite{mirowski2024robot} found that safety filters often erase minority perspectives in comedic contexts, reinforcing hegemonic norms; they advocate artist-centric alignment. Red-teaming LVLMs revealed that dark-humor prompts can bypass safety tuning, generating toxic or insulting content. These works highlight the need to blend cultural sensitivity, adversarial testing, and humor-aware refusal strategies when deploying humor recognition or interaction systems.

Humor detection is converging on a data–model–alignment playbook: realistic, culturally grounded multimodal benchmarks; lightweight prompting for quick gains; targeted adaptation and representation reuse for domain fit; and culture- and translation-aware priors. Multimodal fusion and knowledge-centric approaches advance explanation and situational awareness, while humor-aware refusals and adversarial evaluation integrate safety into deployment. Ultimately, humor serves as a sharp probe of commonsense, pragmatics, and cross-modal reasoning, paving a practical path from benchmarking to robust, culturally sensitive interaction.

\subsection{Key Challenges of Humor Detection}

Humor detection presents challenges for LLMs, primarily due to the subjective and culturally dependent nature of humor. What one individual or culture finds hilarious, another might find offensive, confusing, or not funny. This subjectivity makes it difficult to create applicable humor detection models and to establish a clear "ground truth" for training data. Annotators may disagree on whether a particular text is humorous, leading to noisy labels and potentially biased models. Another challenge is the diversity and complexity of humor. Humor can manifest in countless forms, including puns, satire, irony, slapstick, absurdity, and observational humor, each with its own linguistic and cognitive mechanisms. LLMs may struggle to learn a unified representation that captures all these varied types effectively. For example, understanding a pun requires phonological and semantic knowledge, while understanding satire requires awareness of social norms and critique.

The reliance on implicit meaning, common sense, and world knowledge is another hurdle. Many jokes rely on unstated premises, cultural references, or an understanding of typical scenarios that are then subverted. LLMs, despite their vast training data, may not always possess the depth of world knowledge or the ability to make the subtle inferences required to "get" a joke. The incongruity-resolution theory of humor suggests that humor arises from the perception of an incongruity that is then resolved in a playful or unexpected way. Modeling this cognitive process of identifying and resolving incongruity is a complex task for LLMs. Furthermore, humor is often context-dependent. A statement might be funny in one context but not in another. Capturing and representing the relevant context, especially in short texts or isolated utterances, can be difficult. Finally, evaluating humor detection systems is challenging. Standard metrics like accuracy may not fully capture a model's ability to understand humor, especially if the test data is biased or if the humor is particularly subtle. Developing more nuanced evaluation methods that can assess a model's deeper understanding of humor is an ongoing research area.

In summary, LLMs have made noticeable progress in recognizing humor and even explaining certain kinds of jokes, but they are far from truly understanding all humor as humans do. They tend to be formulaic and miss subtle context. Humor detection research with LLMs is pushing boundaries by introducing style-aware evaluation, co-creativity studies, and multimodal humor tasks. These efforts highlight both the capability and the limits of current models. The insights gained (e.g., need for multi-step reasoning and context modeling) echo those in sarcasm and metaphor tasks. We next look at stance detection, another task requiring subjective understanding, where LLMs are proving useful, especially in zero-shot and multi-agent settings.

\section{Stance Detection}

\subsection{Task Definition of Stance Detection}

Stance detection is a core NLP task that determines an author’s position toward an explicitly specified target (e.g., a person, policy, product, or proposition), typically assigning labels such as Favor/Support, Against/Oppose, Neutral/Neither, and in some frameworks Query when the text asks about the target without a clear position. Unlike sentiment analysis, which captures overall positive/negative tone, stance is inherently target-dependent: the same text can take different stances depending on the target and may diverge from its sentiment (e.g., “The policy is harsh but necessary” is negative in tone yet supportive in stance). This target-conditioned nature—and the need to capture both explicit signals and implicit, context-driven cues—makes stance detection challenging but vital for political discourse analysis, rumor and misinformation tracking, public health and opinion monitoring, argument mining, and social media moderation. With the rise of online communication, its importance has grown, and modern large language models, combining broad world knowledge with pragmatic inference, are increasingly effective for few- and zero-shot stance detection on emerging topics.

\subsection{Dataset of Stance Detection}

Recent surveys of stance detection for the LLM era emphasize that progress is driven by diverse, well-annotated datasets spanning social media, news, forums, and debate platforms, and covering targets such as political actors, ideologies, policies, products, and public-health topics. Annotation typically presents a text with a target and asks human annotators to label stance; inter-annotator agreement is a key quality signal. Dataset characteristics—size, class balance, target specificity, explicit vs. implicit stance, and modality—strongly shape model design and performance. Cross-lingual and multimodal resources are increasingly common, reflecting the realities of online discourse.

Early work established core benchmarks and task variants. \textbf{SemEval-2016 Task 6} on Twitter included five targets (Atheism, Climate Change, Feminism, Hillary Clinton, Legalization of Abortion) with Favor/Against/None labels, showing that tweets can convey stance without explicitly naming the target. Related lines include rumor-related stance tasks with conversational labels such as support, deny, query, and discuss (as in SemEval), and the Fake News Challenge (\textbf{FNC-1}), which frames stance between headlines and articles as agree, disagree, discuss, or unrelated. Broader-document settings appear in debate/forum corpora such as PERSPECTRum\cite{chen2019seeing} and in argumentation resources like \textbf{ArgMin}, as well as stance in news commentaries. Multilingual threads include datasets around specific political contexts (e.g., stance on Catalan independence). Mohammad’s releases surrounding the SemEval stance task consolidated early public resources and practices for evaluation.

As the field matured, a subsequent phase emphasized cross-target and event-driven generalization. P-STANCE\cite{li2021p} collects tweets expressing stance toward U.S. political figures and is designed for cross-target evaluation (training on one figure, testing on another), probing target transfer. COVID-19\cite{glandt2021stance} datasets capture stance toward fast-evolving public-health topics (e.g., masking), adding noise, sarcasm, and shifting narratives typical of crisis-time social media. These corpora broadened domains and targets while keeping tweet-scale inputs well matched to general-purpose language models.

More recently, datasets increasingly meet LLM-oriented needs: broader topical coverage, more languages, and richer signals. VAST\cite{allaway2020zero} (Varied Stance Topics) extends target breadth and domains to test open-domain stance recognition. VaxxStance\cite{agerri2021vaxxstance} focuses on vaccine-related stance, supporting research on public-health discourse and implicit attitudes. MAWQIF\cite{alturayeif2022mawqif} expands Arabic stance resources, advancing cross-lingual and low-resource evaluation. Recent corpora also explore multimodality (text plus images or user/profile cues) and finer-grained or implicit labels, making annotation harder and class balance more uneven. These datasets are central to benchmarking zero-shot and cross-target capabilities of LLMs and to studying robustness in realistic, multilingual settings, including misinformation and rapidly evolving events. Together, the earlier benchmarks (e.g., SemEval-2016, FNC-1, debate/news datasets) and the latest resources (e.g., VAST, VaxxStance, MAWQIF) provide complementary testbeds for stance detection, enabling systematic comparison of methods while highlighting persistent challenges such as implicit stance, domain shift, and cross-lingual transfer.

\subsection{LLM methods of Stance Detection}

\paragraph{Symbolic and Logic-Augmented Reasoning} Within Symbolic and Logic-Augmented Reasoning, fusing symbolic constraints with LLMs yields more interpretable and consistent stance decisions. From a rule-encoding perspective, FOLAR encodes First-Order Logic (FOL) rules elicited by Chain-of-Thought into a Logic Tensor Network and applies multi-decision fusion to curb bias \cite{dai2025large}. From the lens of rationale unification, LogiMDF consolidates divergent LLM rationales via a Logical Fusion Schema and models them with a multi-view hypergraph network to reconcile inconsistencies \cite{zhang2025logic}. From the angle of prompt-based knowledge integration, prompt-tuned fusion frameworks leverage multi-prompt learning and explanation-guided supervision to incorporate LLM-acquired knowledge, strengthening reasoning while filtering noise \cite{ding2024leveraging}\cite{ding2024distantly}. For cross-target transfer, performance improves when LLMs surface target-oriented analytical perspectives and natural language explanations that are then fused into the predictor \cite{ding2024cross}. At the memory-augmentation level, semi-parametric “experienced experts” dynamically retrieve domain-specialized memories to stabilize reasoning and reduce hallucinations \cite{wang2024deem}. In sum, symbolic and logic-augmented fusion enhances the interpretability and consistency of stance reasoning.

\paragraph{Chain-of-Thought and Explicit Rationales} Across tasks, explicit reasoning consistently improves zero- and few-shot stance detection. From a methodological perspective, Chain-of-Stance decomposes the decision process into stance-aware steps and delivers substantial gains without task-specific fine-tuning \cite{ma2024chain}, while Stance Reasoner performs zero-shot inference by generating background-grounded reasoning chains that steer the final stance, enhancing both interpretability and generalization \cite{taranukhin2024stance}. In practical terms, CoT-derived explanations can supervise downstream models or calibrate prompt tuning, achieving strong performance at modest cost \cite{ding2024leveraging}\cite{ding2024distantly}. From an annotation perspective, GPT-4’s zero-shot CoT emerges as a competitive and economical alternative to few-shot prompting \cite{liyanage2024gpt}. From a knowledge discovery standpoint, CoT also serves as a knowledge elicitation tool for logic extraction and cross-target analysis pipelines \cite{ding2024cross}\cite{dai2025large}. Chain-of-Thought and explicit rationales constitute a unifying paradigm that enhances accuracy, data efficiency, and interpretability while enabling scalable knowledge discovery in stance detection.

\paragraph{Multi-Agent Collaboration and Consistency} Collaborative agent architectures integrate multi-faceted knowledge and enforce cross-agent consistency to improve decision quality. In this vein, COLA orchestrates role-infused expert agents—linguistic, domain, and social-media specialists—that debate and consolidate a stance, providing explainability without requiring additional training data \cite{lan2024stance}. Extending structured deliberation, ZSMD sets up support-versus-oppose debaters augmented with background knowledge and introduces a referee to resolve disagreements, thereby improving zero-shot performance and capturing nuance \cite{ma2025exploring}. On the efficiency front, CoVer amortizes LLM reasoning over batches and employs a small model to verify logical consistency and aggregate predictions, substantially reducing LLM calls while maintaining state-of-the-art results \cite{yan2025collaborative}. Multi-Agent Collaboration and Consistency—through role specialization, structured debate with arbitration, and systematic consistency checking—yield a unified, cost-conscious pipeline that enhances accuracy, robustness, and explainability in stance assessment.

\paragraph{Knowledge Injection and Retrieval Augmentation} Injecting structured knowledge helps bridge target–text gaps and stabilizes zero-shot and cross-target settings. At the representation level, prompted LLMs extract target–text relations that are fed into a generation model and coupled with prototypical contrastive alignment to strengthen decoding \cite{zhang2024llm}. From a retrieval perspective, retrieval-augmented pipelines ground tweet–claim relations with evidence and LLM reasoning to enhance stance truthfulness \cite{zhu2025ratsd}. On the training side, multi-task fine-tuning with debate data and knowledge retrieval complements LLM semantics and boosts zero-shot performance \cite{fan2025enhancing}. For low-resource transfer, knowledge can be infused and aligned from diverse sources \cite{yan2024collaborative}. In terms of data coverage, synthetic open-domain stance corpora generated by ChatGPT expand coverage while remaining cost-effective \cite{zhao2024zerostance}. Knowledge Injection and Retrieval Augmentation operate as synergistic levers that ground representations and decoding with explicit evidence, improve truthfulness, and extend generalization across domains, targets, and resource levels at cost.

\paragraph{Fine-Tuning, Reinforcement, and Data Curation} Modern LLMs—including midsize models fine-tuned on public datasets—now surpass prior benchmarks and offer strong efficiency–accuracy tradeoffs \cite{gul2024stance}. Methodologically, reinforcement tuning with hybrid rewards can surface high-quality LLM-annotated examples and enable joint stance detection and rumor verification under label scarcity \cite{yang2024reinforcement}. From a data-annotation standpoint, LLMs themselves are viable labelers: few-shot and zero-shot chain-of-thought GPT-4 labeling approaches approximate supervised baselines at lower cost \cite{liyanage2024gpt}. From a domain-adaptation angle, domain-specific corpora such as $\delta$-Stance show that while proprietary LLMs capture polarity, supervised fine-tuning is essential for modeling intensity and supports cross-domain transfer \cite{gupta2025stance}. In terms of data augmentation, synthetic open-domain datasets complement human labels and improve generalization to unseen targets \cite{zhao2024zerostance}. Fine-Tuning, Reinforcement, and Data Curation function as mutually reinforcing pillars, yielding systems that are accurate, sample-efficient, and robust across domains and targets.

\paragraph{Multimodal and Multilingual Stance Understanding} From a cross-lingual perspective, VLMs underuse visual cues and over-rely on textual content, with performance strongly shaped by language support and model size \cite{vasilakes2025exploring}. With respect to multimodal conversational use, complexity increases; new datasets and MLLM architectures that learn joint text-image representations achieve state-of-the-art results yet still reveal substantial headroom \cite{niu2024multimodal}. In terms of multi-turn dialogue, progress remains limited: even specialized attention mechanisms yield only modest gains on recent benchmarks, highlighting unresolved issues in long-range dependencies and dialogue-role modeling \cite{niu2024challenge}. From an application and safety standpoint, stance-driven generation systems demonstrate downstream utility and safety-aware content creation in advocacy contexts \cite{wang2024memecraft}. Overall, the field is advancing but unevenly, with improved representation learning tempered by persistent challenges in cross-lingual grounding, long-context reasoning, and controllable safe generation.

\paragraph{User-Level Stance and Political Bias} From a theory-driven standpoint, agendas call for shifting from message-level to user-level modeling, integrating psychological features and LLM-inferred attributes to better capture stance formation \cite{bhattacharya2025rethinking}. Methodologically, unsupervised pipelines that map user timelines to socio-political statements via LLM-based NLI generalize across elections and cultures, approaching supervised scores \cite{gambini2024evaluating}. Political bias remains consequential: LLMs skew liberal and are sensitive to demographic cues, underscoring careful prompt design \cite{pit2024whose}. At the dataset level, effects dominate performance variance in political stance tasks, and target ambiguity exacerbates errors, calling for clearer target specification and robust prompting \cite{ng2025examining}. Overall, advances will hinge on user-centered modeling, bias-aware prompting, and clearer targets.

A comprehensive survey inventories LLM-driven stance detection across learning regimes, modalities, and target relations, mapping applications (misinformation, politics, health, moderation) and open challenges (implicit stance, bias, explainability, low-resource, real-time, compute). The emerging toolbox—logic-infused reasoning, multi-agent collaboration, knowledge retrieval/injection, and data-centric supervision—charts a coherent path toward interpretable, scalable, and generalizable stance detection systems that transfer across targets, modalities, and users.

\subsection{Key Challenges of Stance Detection}

Despite the significant progress enabled by Large Language Models (LLMs) in stance detection, several key challenges persist, limiting the robustness and general applicability of current systems. For instance, implicit stance expression, cultural biases in training data, and the computational costs associated with LLMs. Implicit stance expression is a major hurdle because individuals often convey their opinions indirectly, using sarcasm, irony, or subtle linguistic cues that are difficult for models to interpret accurately without a deep understanding of context and world knowledge. LLMs, despite their advanced capabilities, can still struggle with such nuanced language, leading to misclassification. Cultural biases present in the vast corpora used to pretrain LLMs can also propagate into stance detection models, causing them to perform differently across various demographic groups or cultural contexts. This can lead to unfair or inaccurate predictions, particularly when dealing with sensitive topics or diverse user bases. Addressing these biases requires careful dataset curation, debiasing techniques, and culturally-aware model development.

Another challenge is the computational expense of training and deploying LLMs, especially for real-time applications or resource-constrained environments. While parameter-efficient fine-tuning methods offer relief, the inference latency and hardware requirements for state-of-the-art LLMs can be prohibitive. Furthermore, the dynamic nature of language and the emergence of slang, neologisms, and discourse patterns mean that models can quickly become outdated if not continuously updated or retrained. The evaluation of stance detection models presents challenges, as human annotators may disagree on the stance label for ambiguous texts, making it difficult to establish a ground truth. Developing evaluation metrics that can capture the nuances of stance and account for inter-annotator disagreement is an ongoing area of research. These challenges highlight the need for further research in areas like explainable stance reasoning, low-resource adaptation, and the development of real-time deployment frameworks for LLM-based stance detection systems.

Stance detection benefits from LLMs’ language mastery and reasoning. The post-ChatGPT era has yielded methodologies – sequential reasoning, multi-agent frameworks – that have pushed stance detection performance to new highs, even in zero-shot settings. Yet, challenges like implicit stance and model bias ensure it remains an active field. The innovations, such as structured reasoning about opinions, could be relevant to other subjective tasks, since stance co-occurs with sentiment, emotion, and figurative language. One such overlapping task is metaphor recognition, which we examine next, where LLMs are used to detect when language is used non-literally.

\section{Metaphor Recognition}

\subsection{Task Definition of Metaphor Recognition}

Metaphor recognition is a task in natural language processing that involves identifying and interpreting metaphorical language within text. Metaphors are figurative expressions where a concept (the "target" or "tenor") is understood in terms of another, often unrelated, concept (the "source" or "vehicle"). For example, in the phrase "time is money," the abstract concept of "time" (target) is conceptualized through the more concrete concept of "money" (source), implying that time is a valuable resource that can be spent, saved, or wasted. The task of metaphor recognition typically involves two sub-tasks: metaphor identification (detecting whether a word or phrase is used metaphorically) and metaphor interpretation (understanding the meaning conveyed by the metaphor, often by identifying the mapping between source and target domains). This is essential for deeper natural language understanding, as metaphors are pervasive in everyday communication, literature, and specialized domains like law and medicine. Understanding metaphors enables LLMs to grasp nuanced meanings, infer speaker intent, and produce more coherent, context-appropriate text. This is plausible because LLMs possess broad semantic knowledge: they “know” word meanings and can detect expectation-violating usages, much like humans rely on Selectional Preference Violation as a cue for metaphor. Post-ChatGPT models have been applied to metaphor tasks to see if they grasp abstract figurative language and even to generate metaphors.

\subsection{Dataset of Metaphor Recognition}

Research on metaphor Recognition has been anchored by several foundational datasets created before 2022. The VU Amsterdam Metaphor Corpus\cite{steen2010method} (VUA/VUAMC) remains the primary benchmark: it provides token-level metaphor annotations across genres (news, fiction, academic, conversation) following MIP/MIPVU, covers all parts of speech, and contains tens of thousands of labels; it also underpinned the 2018 and 2020 shared tasks. Verb-focused resources include TroFi\cite{birke2005clustering} (~3k sentences for ~50 target verbs labeled as literal vs metaphorical) and MOH-X\cite{mohammad2016metaphor} (647 verb instances), which are frequently paired for evaluation. Other established sets target specific constructions or populations: LCC datasets\cite{mohler2016introducing} emphasize adjective–noun metaphors; the TOEFL metaphor dataset\cite{klebanov2018corpus} annotates second-language learner sentences; and the Stab news corpus marks sentence-level metaphor presence. While mainstream benchmarks frame the problem as word-level tagging, some studies also consider multi-word or idiomatic metaphors.

Recently, the landscape expanded toward evaluation of deeper interpretation and LLM robustness. The Metaphor Understanding Challenge Dataset (MUNCH) \cite{tong2024metaphor} is a notable LLM-oriented benchmark that couples naturally occurring metaphorical sentences from four genres with over 10,000 human-written apt paraphrases and 1,500 inapt paraphrases, enabling tests that distinguish genuine metaphor understanding from lexical overlap; it also spans varying levels of novelty and is openly available. In Chinese, a recent shared-task-style resource (“Task 9”)\cite{che2016semeval} provides 34,463 metaphorical sentences annotated with tenor, vehicle, and ground, plus two 500-sentence validation sets aligned with the test format, supporting component-level analysis beyond binary identification. Complementing this, CMDAG\cite{shao2024cmdag} is a 28K-sentence Chinese literary corpus annotated for tenor, vehicle, and ground that uniquely leverages grounds as chain-of-thought to steer metaphor generation; code is available. Newer datasets also explore metaphor novelty and multilingual coverage (e.g., Russian). Consistent with these trends, recent methods \cite{lin2025dual} continue to report results on VUA/VUAMC and the smaller verb datasets (MOH-X, TroFi), underscoring their role as standard testbeds.

Complementing core resources, domain-specific corpora have been curated to probe generalization, including classical, metaphor-rich texts such as the Bhagavad Gita and the Sermon on the Mount\cite{chandra2024large}. Multimodal efforts link language to gesture or vision to study how metaphors align with nonverbal cues, and some datasets combine figurative categories (e.g., hyperbole with metaphor\cite{zheng2025enhancing}) to encourage unified modeling. Finally, a subset of tasks explicitly targets concurrent or multi-word metaphors, though word-level tagging remains the dominant formulation. Together, these datasets enable both traditional identification and deeper interpretation, and they offer varied genres, languages, and novelty levels for comprehensive evaluation.

\subsection{LLM methods of Metaphor Recognition}

\paragraph{Theory-Guided Prompting Pipelines} Recent advances have shifted metaphor recognition from superficial multimodal fusion to cognitively grounded prompting and scaffolding. On the cognitive side, Chain-of-Cognition prompting encourages models to reason about source–target mappings and cross-modal associations rather than merely combining modalities \cite{zhang2025multimodal}. On the instructional side, Theory-guided Scaffolding Instruction operationalizes metaphor theory through staged questions and a knowledge graph, yielding interpretable decisions and enabling recovery when models falter \cite{tian2024theory}. From a decision-making perspective, Dual-Perspective Metaphor Detection integrates implicit datastore cues with explicit theory-driven prompts and self-judgment to improve reliability and explainability \cite{lin2025dual}. In Chinese multimodal scenarios, a Chain-of-Thought bi-level optimizer approximates human cognition by modeling hierarchical mappings \cite{zhang2025towards}. As for context, lightweight injection proves effective: introducing hypothetical scenarios before proverbs markedly improves word-level detection \cite{goren2024context}. Meanwhile, few-shot GPT-3, although exhibiting partial knowledge of mappings, suffers from source hallucination and overreliance on lexical triggers—underscoring the need for guided reasoning \cite{wachowiak2023does}. Taken together, these advances point toward more human-like, reliable, and interpretable metaphor recognition.

\paragraph{Multimodal Recognition With Imaginative Bridges} Multimodal metaphor understanding hinges on aligning heterogeneous cues through cognitively plausible bridges. Beyond simple caption fusion, Chain-of-Cognition prompting elicits textual entity relations and ties them to visual evidence for cross-domain mapping \cite{zhang2025multimodal}. Under low-resource conditions, imaginative frames grounded in Conceptual Metaphor Theory stimulate cross-modal association and enable data-efficient, retrieval-augmented reasoning \cite{tian2025imara}. On the efficiency front, CDGLT introduces controlled “concept drift” via SLERP-perturbed CLIP embeddings and tunes only LayerNorms to bridge literal–figurative gaps at low cost \cite{qian2025concept}. In Chinese settings, CM3D and a Chain-of-Thought mapping model provide annotated domains and interpretable alignment signals \cite{zhang2025towards}. On the generation and supervision side, a co-creation pipeline that expands linguistic metaphors into textual entailments for diffusion models yields high-quality visual metaphors and intrinsic/extrinsic evaluation signals useful for recognition supervision \cite{chakrabarty2023spy}. Empirically, however, II-Bench shows MLLMs lag humans on implied meanings—especially for abstract or sentiment-laden images—spotlighting sentiment reasoning as a key bottleneck \cite{liu2024ii}. Targeted sentiment-aware reasoning may help close this gap.

\paragraph{Pretraining, Corpora, and Benchmarks} Dedicated resources and pretraining schemes are catalyzing measurable progress. On the pretraining front, MetaPro 2.0 couples a large paraphrase-rich VMC-P corpus with Anomalous Language Modeling, markedly improving identification and literal paraphrasing of figurative expressions \cite{mao2024metapro}. On the diagnostic side, MUNCH differentiates genuine interpretation from mere lexical similarity using apt/inapt paraphrases across genres, revealing persistent LLM mapping gaps \cite{tong2024metaphor}. On the resource side, the Figurative Archive aggregates Italian metaphors with psycholinguistic ratings and corpus metrics, enabling controlled studies of familiarity and concreteness effects \cite{bressler2025figurative}. In Chinese contexts, CMDAG contributes a large corpus with tenors, vehicles, and grounds; supervising with grounds as Chain-of-Thought improves generative quality and yields explicit features reusable for recognition \cite{shao2024cmdag}. For multimodal Chinese, CM3D brings cross-modal mappings into the ecosystem \cite{zhang2025towards}. On the evaluation front, domain-specific studies—from religious texts showing cross-translation consistency \cite{chandra2024large} to large-scale analyses of the Book of Songs revealing cognitive variation \cite{bao2025exploring}—stress cross-cultural robustness. Taken together, these corpora and training paradigms target anomalous language head-on and help standardize evaluation. Continued multilingual expansion will further strengthen generalization.

\paragraph{Context, Domain, and Emotion as Signals} Contextualization, domain adaptation, and affective cues are decisive for recognition. On contextualization, prompted contexts close the abstraction gap in proverb-level detection \cite{goren2024context}. In translation, domain adaptation reduces metaphor errors; literary-adapted NMT and LLMs compare favorably with commercial MT despite a 64–80\% accuracy ceiling \cite{karakanta2025metaphors}. Methodologically, multi-agent reasoning maps culturally laden Traditional Chinese Medicine metaphors to Western medical concepts, illustrating cross-paradigm grounding as a route to reliable mapping \cite{tang2025metaphor}. On the affect side, emotion knowledge helps disambiguate figurative devices: modeling bidirectional dynamics between hyperbole and metaphor with emotion-aware features delivers large F1 gains and reduces type confusion \cite{zheng2025enhancing}. For interdisciplinary reading and political discourse, dialogic/on-demand metaphor generation and prompt-engineered analyses make opaque jargon accessible while preserving critical reflection \cite{yarmand2025towards}\cite{meng2025large}. Across traditions, religious and classical corpora expose invariances and differences that can calibrate domain-sensitive recognizers \cite{chandra2024large}\cite{bao2025exploring}. Together, these directions chart a practical path toward robust metaphor recognition.

\paragraph{Creativity-Aware Signals For Recognition} Recognition improves when systems internalize analogy structure and creativity constraints. At scale, larger LMs better separate metaphors from faulty analogies via perplexity, yet metaphorical fluency remains hard—evidence of structural awareness without full creative competence \cite{boisson2024metaphors}. On data generation and supervision, human–AI co-creation of visual metaphors surfaces entailments and image–text correspondences that supervise detectors and stress-test cross-domain alignment \cite{chakrabarty2023spy}. From an assessment perspective, automatic scoring of metaphor creativity supplies scalar signals aligned with human judgments, a promising auxiliary objective for recognition models \cite{distefano2023automatic}. In tooling, authoring systems that scaffold extended metaphor creation make explicit coherence, extension, and revision steps—signals robust recognizers should verify \cite{kim2023metaphorian}. For evaluation, comparative studies of novel literary metaphors map divergences between human and model interpretations, yielding granular error taxonomies for training and analysis \cite{ichien2024interpretation}. Together, these strands point toward recognition that is structurally grounded and creatively aware.

\paragraph{Limits, Biases, and Reliability Controls} Empirical audits caution against overconfidence: GPT-3 often hallucinates source domains, mislabels literal as figurative (and the reverse), and overrelies on lexical cues at the expense of context \cite{wachowiak2023does}. In vision–language settings, MLLMs trail humans on implied meanings, and sentiment hints can artificially inflate scores—evidence of shallow affective reasoning \cite{liu2024ii}. Conceptually, critics urge reframing “hallucination” as “confabulation,” which better captures context-shaped fabrication in figurative inference \cite{smith2023hallucination}; related links to absolute metaphors and psychosis theories point to structural blind spots in token-based reasoning \cite{heimann2025extimate}. From a comparative cognition standpoint, analyses reveal domain preferences and biases that depart from human metaphor usage, highlighting fairness and generalization risks \cite{mao2025comparative}. On the reliability front, self-judgment with theory-guided prompts \cite{lin2025dual}, emotion-informed multitask training \cite{zheng2025enhancing}, and scaffolded stepwise support \cite{tian2024theory} offer pragmatic controls that raise the floor for trustworthy recognition. Continued stress-testing across domains will be essential.

Metaphor recognition is shifting from shallow fusion to grounded, theory-guided pipelines: chain-of-cognition prompts, scaffolded instruction, and self-judgment align source–target mappings with visual evidence for interpretable decisions. Imaginative multimodal bridges and low-cost adapters improve cross-domain alignment, while corpora and anomalous-language pretraining standardize evaluation and boost generalization. Domain and emotion signals reduce errors; creativity-aware supervision adds structural discipline. Yet LLMs still hallucinate and miss implied sentiment. Next, unifying cognitive scaffolds with retrieval, sentiment-aware modules, and uncertainty—backed by multilingual, domain-rich benchmarks—promises more robust, human-like understanding.

\subsection{Key Challenges of Metaphor Recognition}

Metaphor recognition presents key challenges for Large Language Models (LLMs), primarily stemming from the nuanced, context-dependent, and culturally specific nature of metaphorical language. One challenge, as highlighted by experiments with the MUNCH dataset, is that LLMs may struggle to perform full metaphor interpretation, sometimes relying on lexical similarity rather than genuine cross-domain mapping. This means that even if an LLM correctly identifies a word as metaphorical, it might not accurately understand the intended meaning or the specific way the source domain illuminates the target domain. The interpretation of novel metaphors, which are creative and not part of common parlance, is particularly difficult because LLMs primarily learn from existing text corpora and may not have encountered these specific figurative uses before.

Another significant challenge is the ambiguity in distinguishing metaphorical usage from literal usage, especially for polysemous words (words with multiple meanings). LLMs need to disambiguate word senses based on context, which can be complex when both literal and metaphorical interpretations are plausible. Furthermore, the "grounds" of a metaphor—the specific attributes or features that are mapped from the source domain to the target domain—are often implicit and require deep world knowledge and reasoning to infer. For example, understanding why "time is money" involves understanding cultural values placed on both time and money as resources. LLMs may lack this nuanced understanding or struggle to articulate the specific grounds of a metaphor. The paper on Chinese metaphor recognition also implicitly points to the challenge of adapting LLM methods to different languages, as metaphorical constructions and common mappings can vary significantly across linguistic and cultural boundaries. The need for high-quality, large-scale annotated datasets covering diverse types of metaphors and languages remains a practical challenge for training and evaluating robust metaphor recognition systems. Finally, evaluating the quality of metaphor interpretation by LLMs is non-trivial, as it often requires human judgment and can be subjective. Developing objective and reliable evaluation metrics that capture the depth of understanding is an ongoing research problem. The systematic review on figurative language processing likely discusses these and other challenges in more detail, providing a broader perspective on the limitations of current LLMs in this area.

All considered, LLMs have set new state-of-art in metaphor detection, bringing accuracy up and providing human-readable explanations. This is a significant step for figurative language understanding. The integration of explicit metaphor theory into LLM reasoning is a prime example of combining old linguistic insights with new model capabilities. Such synergy could be a model for other tasks. Finally, we will discuss intent detection and aesthetic evaluation, the last two tasks in our survey, before moving to cross-task analysis and future directions.

\section{Comparative Analysis and Insights}

\subsection{Similarities and Differences among Subjective Language Tasks}

The subjective language tasks discussed—sentiment analysis, emotion recognition, sarcasm detection, humor detection, stance detection, metaphor recognition, user intent detection, and aesthetics identification—share fundamental similarities, yet they also exhibit distinct characteristics that define their challenges and required LLM capabilities. A core similarity is their inherent subjectivity; these tasks involve interpreting language that reflects personal perspectives, feelings, opinions, or evaluations rather than objective facts. This means they are all highly context-dependent and often require understanding implicit meanings, cultural nuances, and speaker intent. For instance, sarcasm, humor, and metaphor all rely on a discrepancy between literal and intended meaning, which LLMs must infer. Similarly, sentiment, emotion, and stance are often conveyed indirectly. Consequently, all these tasks benefit from LLMs' ability to capture deep contextual understanding and semantic relationships.

However, there are also significant differences. Granularity and scope of interpretation vary: sentiment analysis typically deals with broad polarity (positive/negative/neutral), while emotion recognition aims for more specific affective states (joy, anger, etc.). Stance detection focuses on a position towards a target, which can be distinct from general sentiment. Metaphor and sarcasm detection involve identifying specific figurative language constructs. Humor detection targets a particular communicative intent (to amuse). User intent detection is about identifying a goal or purpose, which may or may not be explicitly emotional or evaluative. Aesthetics identification deals with judgments of beauty or artistic merit, a highly abstract and culturally variable concept. The nature of the "target" also differs: stance detection is explicitly target-dependent (e.g., stance towards a policy), whereas sentiment or emotion might be more general or directed at an unspecified entity. The type of reasoning required can also vary; metaphor interpretation often involves analogical reasoning, sarcasm detection requires recognizing incongruity and often negative intent, while humor detection might involve understanding punchlines or absurdity. These differences necessitate specialized approaches or fine-tuning for each task, even when leveraging general-purpose LLMs.

\subsection{Towards Unified Subjective Language Modeling: Potential of Multi-task LLM}

The shared characteristics among subjective language tasks, such as their reliance on context, implicit meaning, and nuanced interpretation, suggest a significant potential for unified subjective language modeling using multi-task LLMs. Instead of training separate models for sentiment analysis, emotion recognition, sarcasm detection, etc., a single, powerful LLM could be trained to perform all these tasks simultaneously or to share representations and knowledge across them. The underlying hypothesis is that understanding one aspect of subjectivity (e.g., emotion) can inform the understanding of others (e.g., sarcasm or humor). For example, recognizing that a statement conveys negative emotion might be a crucial cue for identifying sarcasm if the literal meaning is positive. A multi-task LLM could learn these inter-task relationships implicitly by being exposed to diverse subjective phenomena during training. This approach could lead to more robust and generalizable models, as knowledge acquired from one task with abundant data (e.g., sentiment analysis) could potentially benefit tasks with scarcer data (e.g., aesthetics identification).

The development of such unified models faces challenges, including the need for large-scale, multi-task datasets where texts are annotated for multiple subjective attributes simultaneously. Designing effective multi-task learning architectures and training strategies that allow for positive knowledge transfer without negative interference (where learning one task harms another) is also critical. Furthermore, the diverse output spaces of these tasks (e.g., categorical labels for sentiment, free-text descriptions for aesthetics) require flexible model architectures. However, the potential benefits are substantial. A unified model could offer a more holistic understanding of subjective language, capturing the interplay between different facets of human expression. It could also be more efficient in terms of development and deployment compared to maintaining multiple specialized models. Research in this direction is actively exploring how to best leverage the capabilities of large foundational models for a broad spectrum of subjective understanding tasks, aiming for AI systems that can comprehend the richness and complexity of human subjectivity in a more integrated manner.

\subsection{Multi-task Fusion vs. Single-task Fine-tuning: Which Is More Effective?}

The debate between multi-task fusion (training a single model on multiple tasks) and single-task fine-tuning (fine-tuning a pre-trained LLM separately for each task) is central to developing effective subjective language understanding systems. Each approach has its advantages and disadvantages, and the optimal choice depends on circumstances, including data availability, task similarity, and computational resources. Single-task fine-tuning allows for specialization; the model can be optimized extensively for the nuances of a particular task, potentially leading to higher performance on that benchmark if sufficient task-specific data is available. This approach is straightforward to implement and is widely used. However, it can lead to a proliferation of models if many subjective tasks need to be handled, and it may not leverage the commonalities between related subjective phenomena.

Multi-task fusion, on the other hand, aims to train a single model that can perform across a range of tasks. The advantage is knowledge transfer: learning patterns useful for one task (e.g., detecting negative sentiment) might help in another (e.g., detecting sarcasm, which often involves negative sentiment). This can be particularly beneficial for tasks with limited labeled data, as the model can leverage information from richer tasks. Multi-task learning can also lead to more generalizable and robust representations that capture broader aspects of subjectivity. However, multi-task fusion is more complex to design and train. Challenges include negative transfer (where learning one task interferes with another), imbalanced task difficulties or data sizes, and the need for careful weighting of task losses during training. The effectiveness of multi-task fusion often depends on the relatedness of the tasks; tasks that share underlying linguistic or cognitive mechanisms are more likely to benefit from joint training. Recent trends show a growing interest in exploring multi-task learning paradigms with LLMs, often by extending pre-trained models with shared encoders and task-specific heads, or by using prompts to guide the model towards different tasks. The ultimate goal is to find a balance that harnesses the power of shared learning while preserving task-specific performance.

\section{Challenges and Open Issues}

\subsection{Technical Challenges}

Despite the impressive capabilities of LLMs, several technical challenges persist in the domain of subjective language understanding. A primary challenge is modeling ambiguity and nuance. Subjective language is often inherently ambiguous, with meanings that can shift based on subtle contextual cues, speaker intent, or cultural background. LLMs, while adept at pattern recognition, can still struggle to capture these fine-grained distinctions, sometimes producing interpretations that are too literal or that miss the underlying subtlety. For example, distinguishing between sarcasm and genuine praise, or understanding a metaphor that relies on uncommon cultural knowledge, remains difficult. Handling implicit meaning is another significant hurdle. Much of subjective expression is not explicitly stated but rather implied through tone, figurative language, or shared understanding. LLMs need to go beyond surface-level semantics to infer these implicit meanings accurately. This requires not only vast amounts of training data but also sophisticated reasoning capabilities.

The reliability and bias in training data also pose major technical challenges. LLMs learn from the data they are trained on, and if this data contains biases (e.g., cultural, gender, or racial biases) or reflects subjective annotations with low inter-annotator agreement, these issues can be amplified and perpetuated by the model. This can lead to unfair, inaccurate, or stereotypical outputs when dealing with subjective content. Developing techniques for debiasing models and datasets, and for creating more reliable and diverse annotations, is crucial. Furthermore, the computational cost and resource intensiveness of training and deploying large LLMs limit their accessibility and practical application, especially for real-time systems or in resource-constrained environments. While parameter-efficient fine-tuning methods and model compression techniques offer some relief, achieving state-of-the-art performance often still requires substantial resources. Finally, the "black box" nature of many LLMs makes it difficult to interpret their reasoning processes, especially for complex subjective judgments. Improving the explainability and interpretability of LLMs is essential for building trust and for debugging models when they make errors in subjective understanding.

\subsection{Ethical and Societal Implications}

The application of LLMs to subjective language understanding raises significant ethical and societal implications that must be carefully considered. One major concern is the perpetuation and amplification of biases. LLMs trained on large, unfiltered datasets from the internet can learn and replicate societal biases related to gender, race, ethnicity, religion, and other sensitive attributes. When these models are used for tasks like sentiment analysis, emotion recognition, or content moderation, they may produce biased or unfair outcomes, leading to discrimination or the marginalization of certain groups. For example, a sentiment analysis model might misinterpret expressions from a particular dialect or cultural group due to a lack of representation in its training data. Privacy concerns are also paramount, especially when LLMs are used to analyze personal communications or expressions of emotion. The data used to train these models, or the data they process in deployment, might contain sensitive personal information. Ensuring that this data is handled securely and ethically, with appropriate consent and anonymization, is critical.

Another issue is the potential for manipulation and misuse. LLMs that can understand and generate subjective language could be used to create persuasive or manipulative content, such as misinformation, propaganda, or targeted scams. The ability to mimic human-like empathy or opinion could be exploited to deceive users or influence public opinion in unethical ways. The impact on human communication and creativity is also a concern. Over-reliance on AI for tasks like writing, artistic judgment, or emotional support could potentially diminish human skills in these areas or lead to a homogenization of expression. Furthermore, the deployment of subjective language understanding systems in areas like hiring, loan applications, or criminal justice raises serious questions about fairness, accountability, and due process, especially if the decision-making processes of these systems are not transparent or are found to be biased. Addressing these ethical and societal challenges requires a multi-faceted approach involving researchers, policymakers, and industry stakeholders to develop robust ethical guidelines, fairness-aware algorithms, and appropriate regulatory frameworks.

\subsection{Future Directions}

The field of subjective language understanding with LLMs is rapidly evolving, and several promising future directions are emerging. One key direction is the development of more robust and nuanced LLMs that can better handle ambiguity, implicit meaning, and cultural context. This includes advancing models' reasoning capabilities, perhaps by integrating symbolic reasoning or commonsense knowledge bases, to enable deeper understanding beyond surface-level patterns. Research into multimodal subjective understanding will also continue to grow, as human communication is inherently multimodal (text, speech, vision). LLMs that can effectively integrate information from multiple modalities will be better equipped to interpret complex subjective expressions, such as sarcasm in a video or the emotional tone of an audiovisual narrative. Another important direction is improving the fairness, transparency, and interpretability of LLMs. This involves developing techniques for detecting and mitigating biases in models and datasets, as well as creating methods to explain why an LLM made a particular subjective judgment, which is crucial for building trust and accountability.

The exploration of personalized and adaptive subjective language models is a promising avenue. Future LLMs might adapt their understanding of subjectivity to individual users' communication styles, preferences, and cultural backgrounds, leading to more empathetic and effective interactions. There is a need for better evaluation benchmarks and metrics designed for subjective tasks. Current benchmarks focus on accuracy, which may not fully capture a model's ability to understand subtle nuances or handle diverse perspectives. Developing holistic evaluation frameworks that consider fairness, robustness, and human-aligned judgment will be crucial. Finally, the ethical development and deployment of subjective language understanding systems will remain a critical focus. This includes establishing clear ethical guidelines, promoting responsible AI practices, and fostering interdisciplinary collaboration to ensure that these powerful technologies are used for beneficial and equitable purposes. The integration of insights from linguistics, cognitive science, psychology, and social sciences will continue to be vital for advancing the field in a human-centric manner.

\section{Conclusion}

\subsection{Summary of Findings}

This survey has provided a comprehensive overview of the state of subjective language understanding with Large Language Models (LLMs), covering key tasks such as sentiment analysis, emotion recognition, sarcasm detection, humor detection, stance detection, metaphor recognition, user intent detection, and aesthetics identification. We have seen that LLMs, with their capabilities in contextual understanding and semantic representation, have advanced the performance on these tasks compared to traditional methods. The evolution of language models, particularly the advent of Transformer-based architectures and large-scale pre-training, has been instrumental in this progress. Various approaches, including prompt-based learning, supervised fine-tuning (SFT), and reasoning-based methods, are being employed to adapt LLMs to the nuances of subjective language. Multi-model and multimodal LLMs are also emerging as powerful tools for handling the complexities of subjective expression, especially when it involves multiple sources of information.

However, the survey also highlights that challenges remain. The inherent subjectivity, ambiguity, and context-dependency of language pose difficulties for LLMs. Tasks like sarcasm detection, metaphor interpretation, and aesthetics identification, which require understanding of implicit meaning and cultural nuances, are particularly challenging. Issues such as bias in training data, the computational cost of LLMs, and the "black box" nature of their decision-making processes need to be addressed. The comparative analysis revealed both similarities and differences among subjective tasks, pointing towards the potential for unified modeling approaches but also the need for task-specific considerations. Ethical and societal implications, including privacy concerns and the potential for misuse, underscore the importance of responsible development and deployment of these technologies.

\subsection{Subjective Language Understanding as a Key Direction for Future LLM Research}

Subjective language understanding stands as a critical and challenging frontier for future LLM research. As LLMs become increasingly integrated into human-facing applications, their ability to accurately perceive, interpret, and respond to the rich tapestry of human emotions, opinions, intentions, and figurative expressions becomes paramount for creating truly intelligent and empathetic AI systems. The nuances of subjective language—sarcasm, humor, metaphor, aesthetic judgment—are fundamental to human communication and social interaction. Mastering these aspects will enable LLMs to move beyond mere information processing to engage in more natural, meaningful, and contextually aware dialogues. Future research in this area will not only push the boundaries of NLP but also contribute to a deeper understanding of human cognition and language itself. The development of LLMs that can robustly handle subjectivity will unlock new possibilities in areas such as personalized education, mental health support, creative arts, and cross-cultural communication, making AI a more valuable and trustworthy partner in various aspects of human life.

The challenges inherent in subjective language understanding, such as ambiguity, context-dependency, and cultural variability, necessitate innovation in LLM architectures, training methodologies, and evaluation techniques. Future research should focus on enhancing LLMs' reasoning capabilities, ability to integrate commonsense and world knowledge, and capacity to learn from limited or noisy data. Exploring multimodal approaches that combine text with other sensory inputs will be crucial for a holistic understanding of subjective expression. Furthermore, addressing issues of bias, fairness, and interpretability will be essential for building trustworthy and ethically sound subjective language understanding systems. The insights gained from tackling subjective language will likely also benefit other areas of AI, leading to more robust and human-like machine intelligence overall.

\subsection{Calling for a Unified Research Framework and Evaluation Benchmarks}

To systematically advance the field of subjective language understanding with LLMs, there is a pressing need for a unified research framework and standardized evaluation benchmarks. Currently, research efforts are often fragmented, with different studies using varied datasets, evaluation metrics, and experimental setups, making it difficult to compare results and track progress effectively. A unified framework would provide common definitions, taxonomies, and methodological guidelines for studying subjective language. This would facilitate collaboration, reproducibility, and the sharing of insights across different research groups and tasks. Such a framework should encompass the diverse aspects of subjectivity, from affective states and opinions to figurative language and aesthetic judgments, acknowledging their interconnections while also respecting their unique characteristics.

Crucially, the development of comprehensive and challenging evaluation benchmarks is essential. These benchmarks should go beyond simple accuracy metrics and aim to assess LLMs' abilities to handle nuance, ambiguity, context-dependency, and cultural diversity. They should include datasets that represent a wide range of subjective phenomena, languages, and domains, including carefully curated adversarial examples to test model robustness. Human evaluation, involving diverse annotators, should be an integral part of these benchmarks to provide a more holistic assessment of model performance aligned with human judgment. Furthermore, benchmarks should be designed to probe not only the "what" (e.g., correct classification) but also the "why" (e.g., model's reasoning process, where feasible). By establishing such a unified framework and robust benchmarks, the research community can accelerate progress towards LLMs that can truly understand and engage with the complexities of human subjective experience.

\bibliographystyle{plainnat}
\bibliography{main}

\begin{thebibliography}{100}

\bibitem{achiam2023gpt}
Josh Achiam, Steven Adler, Sandhini Agarwal, Lama Ahmad, Ilge Akkaya, Florencia~Leoni Aleman, Diogo Almeida, Janko Altenschmidt, Sam Altman, Shyamal Anadkat, et~al.
\newblock Gpt-4 technical report.
\newblock {\em arXiv preprint arXiv:2303.08774}, 2023.

\bibitem{agerri2021vaxxstance}
Rodrigo Agerri, Roberto Centeno, Mar{\i}a Espinosa, Joseba~Fernandez de~Landa, and Alvaro Rodrigo.
\newblock Vaxxstance: A dataset for cross-lingual stance detection on vaccines.
\newblock 2021.

\bibitem{agua2025large}
Mariana {\'A}gua, Nuno Ant{\'o}nio, Marco~P Carrasco, and Carimo Rassal.
\newblock Large language models powered aspect-based sentiment analysis for enhanced customer insights.
\newblock {\em Tourism \& Management Studies}, 21(1):1--19, 2025.

\bibitem{ahmed2024leveraging}
Rabbia Ahmed, Sadaf~Abdul Rauf, and Seemab Latif.
\newblock Leveraging large language models and prompt settings for context-aware financial sentiment analysis.
\newblock In {\em 2024 5th International Conference on Advancements in Computational Sciences (ICACS)}, pages 1--9. IEEE, 2024.

\bibitem{allaway2020zero}
Emily Allaway and Kathleen McKeown.
\newblock Zero-shot stance detection: A dataset and model using generalized topic representations.
\newblock {\em arXiv preprint arXiv:2010.03640}, 2020.

\bibitem{alturayeif2022mawqif}
Nora~Saleh Alturayeif, Hamzah~Abdullah Luqman, and Moataz Aly~Kamaleldin Ahmed.
\newblock Mawqif: A multi-label arabic dataset for target-specific stance detection.
\newblock In {\em Proceedings of the Seventh Arabic Natural Language Processing Workshop (WANLP)}, pages 174--184, 2022.

\bibitem{amal2025well}
Ikhlasul Amal and Annisa~Nur Ramadhani.
\newblock How well do vision-language models explain sarcasm? an evaluation of multimodal explanation quality for social media posts.
\newblock {\em Artificial Intelligence Systems and Its Applications}, 1(1):31--55, 2025.

\bibitem{arora2024intent}
Gaurav Arora, Shreya Jain, and Srujana Merugu.
\newblock Intent detection in the age of llms.
\newblock {\em arXiv preprint arXiv:2410.01627}, 2024.

\bibitem{bago2025few}
Petra Bago and Nikola Bakari{\'c}.
\newblock Few-shot prompting, full-scale confusion: Evaluating large language models for humor detection in croatian tweets.
\newblock In {\em Proceedings of the 10th Workshop on Slavic Natural Language Processing (Slavic NLP 2025)}, pages 9--16, 2025.

\bibitem{baluja2024text}
Ashwin Baluja.
\newblock Text is not all you need: Multimodal prompting helps llms understand humor.
\newblock {\em arXiv preprint arXiv:2412.05315}, 2024.

\bibitem{bao2025exploring}
Hui Bao, Kai He, Yige Wang, and Zeyu Gao.
\newblock Exploring cognitive difference in poetry collection via large language models and metaphors: A case study of the book of songs.
\newblock {\em Cognitive Computation}, 17(3):106, 2025.

\bibitem{bhargava2025impact}
Naman Bhargava, Mohammed~I Radaideh, O~Hwang Kwon, Aditi Verma, and Majdi~I Radaideh.
\newblock On the impact of language nuances on sentiment analysis with large language models: Paraphrasing, sarcasm, and emojis.
\newblock {\em arXiv preprint arXiv:2504.05603}, 2025.

\bibitem{bhattacharya2025rethinking}
Prasanta Bhattacharya, Hong Zhang, Yiming Cao, Wei Gao, Brandon~Siyuan Loh, Joseph~JP Simons, and Liang~Ze Wong.
\newblock Rethinking stance detection: A theoretically-informed research agenda for user-level inference using language models.
\newblock {\em arXiv preprint arXiv:2502.02074}, 2025.

\bibitem{bhosale2023sarcasm}
Swapnil Bhosale, Abhra Chaudhuri, Alex Lee~Robert Williams, Divyank Tiwari, Anjan Dutta, Xiatian Zhu, Pushpak Bhattacharyya, and Diptesh Kanojia.
\newblock Sarcasm in sight and sound: Benchmarking and expansion to improve multimodal sarcasm detection.
\newblock {\em arXiv preprint arXiv:2310.01430}, 2023.

\bibitem{birke2005clustering}
Julia Birke.
\newblock A clustering approach for the unsupervised recognition of nonliteral language.
\newblock 2005.

\bibitem{boisson2024metaphors}
Joanne Boisson, Asahi Ushio, Hsuvas Borkakoty, Kiamehr Rezaee, Dimosthenis Antypas, Zara Siddique, Nina White, and Jose Camacho-Collados.
\newblock How are metaphors processed by language models? the case of analogies.
\newblock In {\em Proceedings of the 28th Conference on Computational Natural Language Learning}, pages 365--387, 2024.

\bibitem{booij2025mememind}
Diederik Booij.
\newblock Mememind framework: Leveraging large language models for meme classification and xai.
\newblock Master's thesis, 2025.

\bibitem{bressler2025figurative}
Maddalena Bressler, Veronica Mangiaterra, Paolo Canal, Federico Frau, Fabrizio Luciani, Biagio Scalingi, Chiara Barattieri di~San Pietro, Chiara Battaglini, Chiara Pompei, Fortunata Romeo, et~al.
\newblock Figurative archive: an open dataset and web-based application for the study of metaphor.
\newblock {\em arXiv preprint arXiv:2503.00444}, 2025.

\bibitem{brown2020language}
Tom Brown, Benjamin Mann, Nick Ryder, Melanie Subbiah, Jared~D Kaplan, Prafulla Dhariwal, Arvind Neelakantan, Pranav Shyam, Girish Sastry, Amanda Askell, et~al.
\newblock Language models are few-shot learners.
\newblock {\em Advances in neural information processing systems}, 33:1877--1901, 2020.

\bibitem{carrasco2024large}
Carlos Carrasco-Farre.
\newblock Large language models are as persuasive as humans, but how? about the cognitive effort and moral-emotional language of llm arguments.
\newblock {\em arXiv preprint arXiv:2404.09329}, 2024.

\bibitem{castro2019towards}
Santiago Castro, Devamanyu Hazarika, Ver{\'o}nica P{\'e}rez-Rosas, Roger Zimmermann, Rada Mihalcea, and Soujanya Poria.
\newblock Towards multimodal sarcasm detection (an \_obviously\_ perfect paper).
\newblock {\em arXiv preprint arXiv:1906.01815}, 2019.

\bibitem{chakrabarty2023spy}
Tuhin Chakrabarty, Arkadiy Saakyan, Olivia Winn, Artemis Panagopoulou, Yue Yang, Marianna Apidianaki, and Smaranda Muresan.
\newblock I spy a metaphor: Large language models and diffusion models co-create visual metaphors.
\newblock {\em arXiv preprint arXiv:2305.14724}, 2023.

\bibitem{chandra2024large}
Rohitash Chandra, Abhishek Tiwari, Naman Jain, and Sushrut Badhe.
\newblock Large language models for metaphor detection: Bhagavad gita and sermon on the mount.
\newblock {\em IEEE Access}, 12:84452--84469, 2024.

\bibitem{che2016semeval}
Wanxiang Che, Yanqiu Shao, Ting Liu, and Yu~Ding.
\newblock Semeval-2016 task 9: Chinese semantic dependency parsing.
\newblock In {\em Proceedings of the 10th International Workshop on Semantic Evaluation (SemEval-2016)}, pages 1074--1080, 2016.

\bibitem{chen2025emova}
Kai Chen, Yunhao Gou, Runhui Huang, Zhili Liu, Daxin Tan, Jing Xu, Chunwei Wang, Yi~Zhu, Yihan Zeng, Kuo Yang, et~al.
\newblock Emova: Empowering language models to see, hear and speak with vivid emotions.
\newblock In {\em Proceedings of the Computer Vision and Pattern Recognition Conference}, pages 5455--5466, 2025.

\bibitem{chen2019seeing}
Sihao Chen, Daniel Khashabi, Wenpeng Yin, Chris Callison-Burch, and Dan Roth.
\newblock Seeing things from a different angle: Discovering diverse perspectives about claims.
\newblock {\em arXiv preprint arXiv:1906.03538}, 2019.

\bibitem{ChenLZZ25}
Weisi Chen, Wulong Liu, Jiaxin Zheng, and Xu~Zhang.
\newblock Leveraging large language model as news sentiment predictor in stock markets: a knowledge-enhanced strategy.
\newblock {\em Discov. Comput.}, 28(1):74, 2025.

\bibitem{chen2024emotionqueen}
Yuyan Chen, Hao Wang, Songzhou Yan, Sijia Liu, Yueze Li, Yi~Zhao, and Yanghua Xiao.
\newblock Emotionqueen: A benchmark for evaluating empathy of large language models.
\newblock {\em arXiv preprint arXiv:2409.13359}, 2024.

\bibitem{chen2024recent}
Yuyan Chen and Yanghua Xiao.
\newblock Recent advancement of emotion cognition in large language models.
\newblock {\em arXiv preprint arXiv:2409.13354}, 2024.

\bibitem{chen2024talk}
Yuyan Chen, Yichen Yuan, Panjun Liu, Dayiheng Liu, Qinghao Guan, Mengfei Guo, Haiming Peng, Bang Liu, Zhixu Li, and Yanghua Xiao.
\newblock Talk funny! a large-scale humor response dataset with chain-of-humor interpretation.
\newblock In {\em Proceedings of the AAAI Conference on Artificial Intelligence}, volume~38, pages 17826--17834, 2024.

\bibitem{cheng2024emotion}
Zebang Cheng, Zhi-Qi Cheng, Jun-Yan He, Kai Wang, Yuxiang Lin, Zheng Lian, Xiaojiang Peng, and Alexander Hauptmann.
\newblock Emotion-llama: Multimodal emotion recognition and reasoning with instruction tuning.
\newblock {\em Advances in Neural Information Processing Systems}, 37:110805--110853, 2024.

\bibitem{chiang2023vicuna}
Wei-Lin Chiang, Zhuohan Li, Ziqing Lin, Ying Sheng, Zhanghao Wu, Hao Zhang, Lianmin Zheng, Siyuan Zhuang, Yonghao Zhuang, Joseph~E Gonzalez, et~al.
\newblock Vicuna: An open-source chatbot impressing gpt-4 with 90\%* chatgpt quality.
\newblock {\em See https://vicuna. lmsys. org (accessed 14 April 2023)}, 2(3):6, 2023.

\bibitem{CHIU2025102632}
I-Chan Chiu and Mao-Wei Hung.
\newblock Finance-specific large language models: Advancing sentiment analysis and return prediction with llama 2.
\newblock {\em Pacific-Basin Finance Journal}, 90:102632, 2025.

\bibitem{chowdhery2023palm}
Aakanksha Chowdhery, Sharan Narang, Jacob Devlin, Maarten Bosma, Gaurav Mishra, Adam Roberts, Paul Barham, Hyung~Won Chung, Charles Sutton, Sebastian Gehrmann, et~al.
\newblock Palm: Scaling language modeling with pathways.
\newblock {\em Journal of Machine Learning Research}, 24(240):1--113, 2023.

\bibitem{dai2025large}
Genan Dai, Jiayu Liao, Sicheng Zhao, Xianghua Fu, Xiaojiang Peng, Hu~Huang, and Bowen Zhang.
\newblock Large language model enhanced logic tensor network for stance detection.
\newblock {\em Neural Networks}, 183:106956, 2025.

\bibitem{demszky2020goemotions}
Dorottya Demszky, Dana Movshovitz-Attias, Jeongwoo Ko, Alan Cowen, Gaurav Nemade, and Sujith Ravi.
\newblock Goemotions: A dataset of fine-grained emotions.
\newblock {\em arXiv preprint arXiv:2005.00547}, 2020.

\bibitem{deng2025large}
Shiling Deng, Serge Belongie, and Peter~Ebert Christensen.
\newblock Large vision-language models for knowledge-grounded data annotation of memes.
\newblock {\em arXiv preprint arXiv:2501.13851}, 2025.

\bibitem{deng2023llms}
Xiang Deng, Vasilisa Bashlovkina, Feng Han, Simon Baumgartner, and Michael Bendersky.
\newblock Llms to the moon? reddit market sentiment analysis with large language models.
\newblock In {\em Companion Proceedings of the ACM Web Conference 2023}, pages 1014--1019, 2023.

\bibitem{dettmers2023qlora}
Tim Dettmers, Artidoro Pagnoni, Ari Holtzman, and Luke Zettlemoyer.
\newblock Qlora: Efficient finetuning of quantized llms.
\newblock {\em Advances in neural information processing systems}, 36:10088--10115, 2023.

\bibitem{devlin2019bert}
Jacob Devlin, Ming-Wei Chang, Kenton Lee, and Kristina Toutanova.
\newblock Bert: Pre-training of deep bidirectional transformers for language understanding.
\newblock In {\em Proceedings of the 2019 conference of the North American chapter of the association for computational linguistics: human language technologies, volume 1 (long and short papers)}, pages 4171--4186, 2019.

\bibitem{ding2024cross}
Daijun Ding, Rong Chen, Liwen Jing, Bowen Zhang, Xu~Huang, Li~Dong, Xiaowen Zhao, and Ge~Song.
\newblock Cross-target stance detection by exploiting target analytical perspectives.
\newblock In {\em ICASSP 2024-2024 IEEE International Conference on Acoustics, Speech and Signal Processing (ICASSP)}, pages 10651--10655. IEEE, 2024.

\bibitem{ding2024distantly}
Daijun Ding, Genan Dai, Cheng Peng, Xiaojiang Peng, Bowen Zhang, and Hu~Huang.
\newblock Distantly supervised explainable stance detection via chain-of-thought supervision.
\newblock {\em Mathematics}, 12(7):1119, 2024.

\bibitem{ding2024leveraging}
Daijun Ding, Xianghua Fu, Xiaojiang Peng, Xiaomao Fan, Hu~Huang, and Bowen Zhang.
\newblock Leveraging chain-of-thought to enhance stance detection with prompt-tuning.
\newblock {\em Mathematics}, 12(4):568, 2024.

\bibitem{ding2024dynamic}
Hongcheng Ding, Xuanze Zhao, Ruiting Deng, Shamsul~Nahar Abdullah, Deshinta~Arrova Dewi, and Zixiao Jiang.
\newblock Dynamic adaptive optimization for effective sentiment analysis fine-tuning on large language models.
\newblock {\em arXiv preprint arXiv:2408.11856}, 2024.

\bibitem{abs-2405-05496}
Xuanwen Ding, Jie Zhou, Liang Dou, Qin Chen, Yuanbin Wu, Chengcai Chen, and Liang He.
\newblock Boosting large language models with continual learning for aspect-based sentiment analysis.
\newblock {\em CoRR}, abs/2405.05496, 2024.

\bibitem{distefano2023automatic}
PV~DiStefano, JD~Patterson, and R~Beaty.
\newblock Automatic scoring of metaphor creativity with large language models. psyarxiv, 2023.

\bibitem{dong2025mhsdb}
Zhongren Dong, Donghao Wang, Ciqiang Chen, Dong-Yan Huang, and Zixing Zhang.
\newblock Mhsdb: A comprehensive benchmark for multimodal humor and sarcasm detection leveraging foundation models.
\newblock In {\em ICASSP 2025-2025 IEEE International Conference on Acoustics, Speech and Signal Processing (ICASSP)}, pages 1--5. IEEE, 2025.

\bibitem{dubey2024llama}
Abhimanyu Dubey, Abhinav Jauhri, Abhinav Pandey, Abhishek Kadian, Ahmad Al-Dahle, Aiesha Letman, Akhil Mathur, Alan Schelten, Amy Yang, Angela Fan, et~al.
\newblock The llama 3 herd of models.
\newblock {\em arXiv e-prints}, pages arXiv--2407, 2024.

\bibitem{elman1990finding}
Jeffrey~L Elman.
\newblock Finding structure in time.
\newblock {\em Cognitive science}, 14(2):179--211, 1990.

\bibitem{epron2024orpailleur}
Pierre Epron, Ga{\"e}l Guibon, and Miguel Couceiro.
\newblock Orpailleur \& synalp at clef 2024 task 2: Good old cross validation for large language models yields the best humorous detection.
\newblock In {\em Working Notes of the Conference and Labs of the Evaluation Forum (CLEF 2024)}, volume 3740, pages 1841--1856. CEUR-WS. org, 2024.

\bibitem{fan2025enhancing}
Qinlong Fan, Jicang Lu, Yepeng Sun, Qiankun Pi, and Shouxin Shang.
\newblock Enhancing zero-shot stance detection via multi-task fine-tuning with debate data and knowledge augmentation.
\newblock {\em Complex \& Intelligent Systems}, 11(2):151, 2025.

\bibitem{feng2021emowoz}
Shutong Feng, Nurul Lubis, Christian Geishauser, Hsien-chin Lin, Michael Heck, Carel van Niekerk, and Milica Ga{\v{s}}i{\'c}.
\newblock Emowoz: A large-scale corpus and labelling scheme for emotion recognition in task-oriented dialogue systems.
\newblock {\em arXiv preprint arXiv:2109.04919}, 2021.

\bibitem{gambini2024evaluating}
Margherita Gambini, Caterina Senette, Tiziano Fagni, and Maurizio Tesconi.
\newblock Evaluating large language models for user stance detection on x (twitter).
\newblock {\em Machine Learning}, 113(10):7243--7266, 2024.

\bibitem{gautam2025survey}
Himanshu Gautam, Abhishek Gaur, and Dharmendra~Kumar Yadav.
\newblock A survey on the impact of pre-trained language models in sentiment classification task.
\newblock {\em International Journal of Data Science and Analytics}, pages 1--39, 2025.

\bibitem{glandt2021stance}
Kyle Glandt, Sarthak Khanal, Yingjie Li, Doina Caragea, and Cornelia Caragea.
\newblock Stance detection in covid-19 tweets.
\newblock In {\em Proceedings of the 59th annual meeting of the association for computational linguistics and the 11th international joint conference on natural language processing (long papers)}, volume~1, 2021.

\bibitem{gole10sarcasm}
M~Gole, WP~Nwadiugwu, and A~Miranskyy.
\newblock On sarcasm detection with openai gpt-based models. 2023.
\newblock {\em URL: https://doi. org/10.48550/arXiv}, 2312.

\bibitem{gole2024sarcasm}
Montgomery Gole, Williams-Paul Nwadiugwu, and Andriy Miranskyy.
\newblock On sarcasm detection with openai gpt-based models.
\newblock In {\em 2024 34th International Conference on Collaborative Advances in Software and COmputiNg (CASCON)}, pages 1--6. IEEE, 2024.

\bibitem{goren2024context}
Gamze Goren and Carlo Strapparava.
\newblock Context matters: Enhancing metaphor recognition in proverbs.
\newblock In {\em Proceedings of the 2024 Joint International Conference on Computational Linguistics, Language Resources and Evaluation (LREC-COLING 2024)}, pages 3825--3830, 2024.

\bibitem{gul2024stance}
Ilker G{\"u}l, R{\'e}mi Lebret, and Karl Aberer.
\newblock Stance detection on social media with fine-tuned large language models.
\newblock {\em arXiv preprint arXiv:2404.12171}, 2024.

\bibitem{guo2025deepseek}
Daya Guo, Dejian Yang, Haowei Zhang, Junxiao Song, Ruoyu Zhang, Runxin Xu, Qihao Zhu, Shirong Ma, Peiyi Wang, Xiao Bi, et~al.
\newblock Deepseek-r1: Incentivizing reasoning capability in llms via reinforcement learning.
\newblock {\em arXiv preprint arXiv:2501.12948}, 2025.

\bibitem{guo2025cross}
Yihui Guo.
\newblock A cross-cultural study of humor intensity in chinese and english family jokes: A large language model-based approach.
\newblock {\em International Journal of Linguistics Studies}, 5(2):01--10, 2025.

\bibitem{gupta2025stance}
Ankita Gupta, Douglas Rice, and Brendan O’Connor.
\newblock -stance: A large-scale real world dataset of stances in legal argumentation.
\newblock In {\em Proceedings of the 63rd Annual Meeting of the Association for Computational Linguistics (Volume 1: Long Papers)}, pages 31450--31467, 2025.

\bibitem{han2025design}
Yi~Han and Mohsen Moghaddam.
\newblock Design knowledge as attention emphasizer in large language model-based sentiment analysis.
\newblock {\em Journal of Computing and Information Science in Engineering}, 25(2):021007, 2025.

\bibitem{He2024.03.19.24304544}
Lu~He, Samaneh Omranian, Susan McRoy, and Kai Zheng.
\newblock Using large language models for sentiment analysis of health-related social media data: empirical evaluation and practical tips.
\newblock {\em medRxiv}, 2024.

\bibitem{he2024chumor}
Ruiqi He, Yushu He, Longju Bai, Jiarui Liu, Zhenjie Sun, Zenghao Tang, He~Wang, Hanchen Xia, and Naihao Deng.
\newblock Chumor 1.0: A truly funny and challenging chinese humor understanding dataset from ruo zhi ba.
\newblock {\em arXiv preprint arXiv:2406.12754}, 2024.

\bibitem{he2025chumor}
Ruiqi He, Yushu He, Longju Bai, Jiarui Liu, Zhenjie Sun, Zenghao Tang, He~Wang, Hanchen Xia, Rada Mihalcea, and Naihao Deng.
\newblock Chumor 2.0: Towards better benchmarking chinese humor understanding from (ruo zhi ba).
\newblock In {\em Findings of the Association for Computational Linguistics: ACL 2025}, pages 21799--21818, 2025.

\bibitem{heimann2025extimate}
Marc Heimann and Anne-Friederike H{\"u}bener.
\newblock The extimate core of understanding: absolute metaphors, psychosis and large language models.
\newblock {\em AI \& SOCIETY}, 40(3):1265--1276, 2025.

\bibitem{HellwigFW25}
Nils~Constantin Hellwig, Jakob Fehle, and Christian Wolff.
\newblock Exploring large language models for the generation of synthetic training samples for aspect-based sentiment analysis in low resource settings.
\newblock {\em Expert Syst. Appl.}, 261:125514, 2025.

\bibitem{heraldi2024effective}
Fachry~Dennis Heraldi and Zakhralativa Ruskanda.
\newblock Effective intended sarcasm detection using fine-tuned llama 2 large language models.
\newblock In {\em 2024 11th International Conference on Advanced Informatics: Concept, Theory and Application (ICAICTA)}, pages 1--6. IEEE, 2024.

\bibitem{hong2025aer}
Xin Hong, Yuan Gong, Vidhyasaharan Sethu, and Ting Dang.
\newblock Aer-llm: Ambiguity-aware emotion recognition leveraging large language models.
\newblock In {\em ICASSP 2025-2025 IEEE International Conference on Acoustics, Speech and Signal Processing (ICASSP)}, pages 1--5. IEEE, 2025.

\bibitem{hu2022lora}
Edward~J Hu, Yelong Shen, Phillip Wallis, Zeyuan Allen-Zhu, Yuanzhi Li, Shean Wang, Lu~Wang, Weizhu Chen, et~al.
\newblock Lora: Low-rank adaptation of large language models.
\newblock {\em ICLR}, 1(2):3, 2022.

\bibitem{huemobench}
H~Hu, Y~Zhou, L~You, H~Xu, Q~Wang, Z~Lian, FR~Yu, F~Ma, and L~Cui.
\newblock Emobench-m: Benchmarking emotional intelligence for multimodal large language models (2025).

\bibitem{hu2024cracking}
Zhe Hu, Tuo Liang, Jing Li, Yiren Lu, Yunlai Zhou, Yiran Qiao, Jing Ma, and Yu~Yin.
\newblock Cracking the code of juxtaposition: Can ai models understand the humorous contradictions.
\newblock {\em Advances in Neural Information Processing Systems}, 37:47166--47188, 2024.

\bibitem{hurst2024gpt}
Aaron Hurst, Adam Lerer, Adam~P Goucher, Adam Perelman, Aditya Ramesh, Aidan Clark, AJ~Ostrow, Akila Welihinda, Alan Hayes, Alec Radford, et~al.
\newblock Gpt-4o system card.
\newblock {\em arXiv preprint arXiv:2410.21276}, 2024.

\bibitem{hwang2025bottlehumor}
EunJeong Hwang, Peter West, and Vered Shwartz.
\newblock Bottlehumor: Self-informed humor explanation using the information bottleneck principle.
\newblock {\em arXiv preprint arXiv:2502.18331}, 2025.

\bibitem{ichien2024interpretation}
Nicholas Ichien, Du~Stamenkoviƒ{\'a}, Keith Holyoak, et~al.
\newblock Interpretation of novel literary metaphors by humans and gpt-4.
\newblock In {\em Proceedings of the Annual Meeting of the Cognitive Science Society}, volume~46, 2024.

\bibitem{inserte2024large}
Pau~Rodriguez Inserte, Mariam Nakhl{\'e}, Raheel Qader, Gaetan Caillaut, and Jingshu Liu.
\newblock Large language model adaptation for financial sentiment analysis.
\newblock {\em arXiv preprint arXiv:2401.14777}, 2024.

\bibitem{jain2024ai}
Veedant Jain, Felipe dos Santos~Alves Feitosa, and Gabriel Kreiman.
\newblock Is ai fun? humordb: a curated dataset and benchmark to investigate graphical humor.
\newblock {\em arXiv preprint arXiv:2406.13564}, 2024.

\bibitem{jana2025dual}
Soumyadeep Jana, Abhrajyoti Kundu, and Sanasam~Ranbir Singh.
\newblock Dual modality-aware gated prompt tuning for few-shot multimodal sarcasm detection.
\newblock {\em arXiv preprint arXiv:2507.04468}, 2025.

\bibitem{jana2025think}
Soumyadeep Jana, Abhrajyoti Kundu, and Sanasam~Ranbir Singh.
\newblock Think twice before you judge: Mixture of dual reasoning experts for multimodal sarcasm detection.
\newblock {\em arXiv preprint arXiv:2507.04458}, 2025.

\bibitem{jiang2023mistral7b}
Albert~Q. Jiang, Alexandre Sablayrolles, Arthur Mensch, Chris Bamford, Devendra~Singh Chaplot, Diego de~las Casas, Florian Bressand, Gianna Lengyel, Guillaume Lample, Lucile Saulnier, Lélio~Renard Lavaud, Marie-Anne Lachaux, Pierre Stock, Teven~Le Scao, Thibaut Lavril, Thomas Wang, Timothée Lacroix, and William~El Sayed.
\newblock Mistral 7b, 2023.

\bibitem{jiang2024mixtral}
Albert~Q Jiang, Alexandre Sablayrolles, Antoine Roux, Arthur Mensch, Blanche Savary, Chris Bamford, Devendra~Singh Chaplot, Diego de~las Casas, Emma~Bou Hanna, Florian Bressand, et~al.
\newblock Mixtral of experts.
\newblock {\em arXiv preprint arXiv:2401.04088}, 2024.

\bibitem{kang2024can}
Dongjin Kang, Sunghwan Kim, Taeyoon Kwon, Seungjun Moon, Hyunsouk Cho, Youngjae Yu, Dongha Lee, and Jinyoung Yeo.
\newblock Can large language models be good emotional supporter? mitigating preference bias on emotional support conversation.
\newblock {\em arXiv preprint arXiv:2402.13211}, 2024.

\bibitem{karakanta2025metaphors}
Alina Karakanta, Mayra Nas, and Aletta~G Dorst.
\newblock Metaphors in literary machine translation: Close but no cigar?
\newblock {\em Proceedings of Machine Translation Summit XX Volume}, 1:276--286, 2025.

\bibitem{kheiri2023sentimentgpt}
Kiana Kheiri and Hamid Karimi.
\newblock Sentimentgpt: Exploiting gpt for advanced sentiment analysis and its departure from current machine learning.
\newblock {\em arXiv preprint arXiv:2307.10234}, 2023.

\bibitem{khodak2018corpus}
Mikhail Khodak, Nikunj Saunshi, and Kiran Vodrahalli.
\newblock A large self-annotated corpus for sarcasm.
\newblock In {\em Proceedings of the Linguistic Resource and Evaluation Conference (LREC)}, 2018.

\bibitem{kim2023metaphorian}
Jeongyeon Kim, Sangho Suh, Lydia~B Chilton, and Haijun Xia.
\newblock Metaphorian: Leveraging large language models to support extended metaphor creation for science writing.
\newblock In {\em Proceedings of the 2023 ACM Designing Interactive Systems Conference}, pages 115--135, 2023.

\bibitem{kirtac2025large}
Kemal Kirtac and Guido Germano.
\newblock Large language models in finance: what is financial sentiment?
\newblock {\em arXiv preprint arXiv:2503.03612}, 2025.

\bibitem{klebanov2018corpus}
Beata~Beigman Klebanov, Chee~Wee Leong, and Michael Flor.
\newblock A corpus of non-native written english annotated for metaphor.
\newblock In {\em Proceedings of the 2018 Conference of the North American Chapter of the Association for Computational Linguistics: Human Language Technologies, Volume 2 (Short Papers)}, pages 86--91, 2018.

\bibitem{kuila2024deciphering}
Alapan Kuila and Sudeshna Sarkar.
\newblock Deciphering political entity sentiment in news with large language models: Zero-shot and few-shot strategies.
\newblock {\em arXiv preprint arXiv:2404.04361}, 2024.

\bibitem{kwon2024sentiment}
O~Hwang Kwon, Katie Vu, Naman Bhargava, Mohammed~I Radaideh, Jacob Cooper, Veda Joynt, and Majdi~I Radaideh.
\newblock Sentiment analysis of the united states public support of nuclear power on social media using large language models.
\newblock {\em Renewable and Sustainable Energy Reviews}, 200:114570, 2024.

\bibitem{lan2024stance}
Xiaochong Lan, Chen Gao, Depeng Jin, and Yong Li.
\newblock Stance detection with collaborative role-infused llm-based agents.
\newblock In {\em Proceedings of the international AAAI conference on web and social media}, volume~18, pages 891--903, 2024.

\bibitem{abs-2306-14096}
Yinyu Lan, Yanru Wu, Wang Xu, Weiqiang Feng, and Youhao Zhang.
\newblock Chinese fine-grained financial sentiment analysis with large language models.
\newblock {\em CoRR}, abs/2306.14096, 2023.

\bibitem{lee2025large}
Jaewook Lee, Woojin Lee, Oh-Woog Kwon, and Harksoo Kim.
\newblock Do large language models have “emotion neurons”? investigating the existence and role.
\newblock In {\em Findings of the Association for Computational Linguistics: ACL 2025}, pages 15617--15639, 2025.

\bibitem{lee2024pragmatic}
Joshua Lee, Wyatt Fong, Alexander Le, Sur Shah, Kevin Han, and Kevin Zhu.
\newblock Pragmatic metacognitive prompting improves llm performance on sarcasm detection.
\newblock {\em arXiv preprint arXiv:2412.04509}, 2024.

\bibitem{lei2024large}
Yuxuan Lei, Dingkang Yang, Zhaoyu Chen, Jiawei Chen, Peng Zhai, and Lihua Zhang.
\newblock Large vision-language models as emotion recognizers in context awareness.
\newblock {\em arXiv preprint arXiv:2407.11300}, 2024.

\bibitem{li2023large}
Cheng Li, Jindong Wang, Yixuan Zhang, Kaijie Zhu, Wenxin Hou, Jianxun Lian, Fang Luo, Qiang Yang, and Xing Xie.
\newblock Large language models understand and can be enhanced by emotional stimuli.
\newblock {\em arXiv preprint arXiv:2307.11760}, 2023.

\bibitem{li2025deemo}
Deng Li, Bohao Xing, Xin Liu, Baiqiang Xia, Bihan Wen, and Heikki K{\"a}lvi{\"a}inen.
\newblock Deemo: De-identity multimodal emotion recognition and reasoning.
\newblock {\em arXiv preprint arXiv:2504.19549}, 2025.

\bibitem{li2021p}
Yingjie Li, Tiberiu Sosea, Aditya Sawant, Ajith~Jayaraman Nair, Diana Inkpen, and Cornelia Caragea.
\newblock P-stance: A large dataset for stance detection in political domain.
\newblock In {\em Findings of the association for computational linguistics: ACL-IJCNLP 2021}, pages 2355--2365, 2021.

\bibitem{li2025revise}
Yuanchao Li, Yuan Gong, Chao-Han~Huck Yang, Peter Bell, and Catherine Lai.
\newblock Revise, reason, and recognize: Llm-based emotion recognition via emotion-specific prompts and asr error correction.
\newblock In {\em ICASSP 2025-2025 IEEE International Conference on Acoustics, Speech and Signal Processing (ICASSP)}, pages 1--5. IEEE, 2025.

\bibitem{li2024enhancing}
Zaijing Li, Gongwei Chen, Rui Shao, Yuquan Xie, Dongmei Jiang, and Liqiang Nie.
\newblock Enhancing emotional generation capability of large language models via emotional chain-of-thought.
\newblock {\em arXiv preprint arXiv:2401.06836}, 2024.

\bibitem{li2025leveraging}
Zhu Li, Yuqing Zhang, Xiyuan Gao, Shekhar Nayak, and Matt Coler.
\newblock Leveraging large language models for sarcastic speech annotation in sarcasm detection.
\newblock {\em arXiv preprint arXiv:2506.00955}, 2025.

\bibitem{lian2025affectgpt}
Zheng Lian, Haoyu Chen, Lan Chen, Haiyang Sun, Licai Sun, Yong Ren, Zebang Cheng, Bin Liu, Rui Liu, Xiaojiang Peng, et~al.
\newblock Affectgpt: A new dataset, model, and benchmark for emotion understanding with multimodal large language models.
\newblock {\em arXiv preprint arXiv:2501.16566}, 2025.

\bibitem{liang2025m3hg}
Qiao Liang, Ying Shen, Tiantian Chen, and Lin Zhang.
\newblock M3hg: Multimodal, multi-scale, and multi-type node heterogeneous graph for emotion cause triplet extraction in conversations.
\newblock In {\em Findings of the Association for Computational Linguistics: ACL 2025}, pages 11416--11431, 2025.

\bibitem{lin2024cofipara}
Hongzhan Lin, Zixin Chen, Ziyang Luo, Mingfei Cheng, Jing Ma, and Guang Chen.
\newblock Cofipara: a coarse-to-fine paradigm for multimodal sarcasm target identification with large multimodal models.
\newblock {\em arXiv preprint arXiv:2405.00390}, 2024.

\bibitem{lin2025dual}
Yujie Lin, Jingyao Liu, Yan Gao, Ante Wang, and Jinsong Su.
\newblock A dual-perspective metaphor detection framework using large language models.
\newblock In {\em ICASSP 2025-2025 IEEE International Conference on Acoustics, Speech and Signal Processing (ICASSP)}, pages 1--5. IEEE, 2025.

\bibitem{liu2024large}
Yang Liu, Xichou Zhu, Zhou Shen, Yi~Liu, Min Li, Yujun Chen, Benzi John, Zhenzhen Ma, Tao Hu, Zhi Li, et~al.
\newblock Do large language models possess sensitive to sentiment?
\newblock {\em arXiv preprint arXiv:2409.02370}, 2024.

\bibitem{liu2019roberta}
Yinhan Liu, Myle Ott, Naman Goyal, Jingfei Du, Mandar Joshi, Danqi Chen, Omer Levy, Mike Lewis, Luke Zettlemoyer, and Veselin Stoyanov.
\newblock Roberta: A robustly optimized bert pretraining approach.
\newblock {\em arXiv preprint arXiv:1907.11692}, 2019.

\bibitem{liu2024emollms}
Zhiwei Liu, Kailai Yang, Qianqian Xie, Tianlin Zhang, and Sophia Ananiadou.
\newblock Emollms: A series of emotional large language models and annotation tools for comprehensive affective analysis.
\newblock In {\em Proceedings of the 30th ACM SIGKDD Conference on Knowledge Discovery and Data Mining}, pages 5487--5496, 2024.

\bibitem{liu2025caf}
Ziqi Liu, Ziyang Zhou, and Mingxuan Hu.
\newblock Caf-i: A collaborative multi-agent framework for enhanced irony detection with large language models.
\newblock {\em arXiv preprint arXiv:2506.08430}, 2025.

\bibitem{liu2024ii}
Ziqiang Liu, Feiteng Fang, Xi~Feng, Xeron Du, Chenhao Zhang, Noah Wang, Qixuan Zhao, Liyang Fan, CHENGGUANG GAN, Hongquan Lin, et~al.
\newblock Ii-bench: An image implication understanding benchmark for multimodal large language models.
\newblock {\em Advances in Neural Information Processing Systems}, 37:46378--46480, 2024.

\bibitem{liyanage2024gpt}
Chandreen~R Liyanage, Ravi Gokani, and Vijay Mago.
\newblock Gpt-4 as an x data annotator: Unraveling its performance on a stance classification task.
\newblock {\em PloS one}, 19(8):e0307741, 2024.

\bibitem{llanes2024developing}
Jose Llanes-Jurado, Luc{\'\i}a G{\'o}mez-Zaragoz{\'a}, Maria~Eleonora Minissi, Mariano Alca{\~n}iz, and Javier Mar{\'\i}n-Morales.
\newblock Developing conversational virtual humans for social emotion elicitation based on large language models.
\newblock {\em Expert Systems with Applications}, 246:123261, 2024.

\bibitem{ma2025exploring}
Junxia Ma, Changjiang Wang, Lu~Rong, Bo~Wang, and Yaoli Xu.
\newblock Exploring multi-agent debate for zero-shot stance detection: A novel approach.
\newblock {\em Applied Sciences}, 15(9):4612, 2025.

\bibitem{ma2024chain}
Junxia Ma, Changjiang Wang, Hanwen Xing, Dongming Zhao, and Yazhou Zhang.
\newblock Chain of stance: Stance detection with large language models.
\newblock In {\em CCF International Conference on Natural Language Processing and Chinese Computing}, pages 82--94. Springer, 2024.

\bibitem{mahendran2025comparative}
Manish~Barath Mahendran, Aswin~Kumar Gokul, Poornima Lakshmi, and S~Pavithra.
\newblock Comparative advances in financial sentiment analysis: A review of bert, finbert, and large language models.
\newblock In {\em 2025 3rd International Conference on Intelligent Data Communication Technologies and Internet of Things (IDCIoT)}, pages 39--45. IEEE, 2025.

\bibitem{mai2024llama}
Zhelu Mai, Jinran Zhang, Zhuoer Xu, and Zhaomin Xiao.
\newblock Is llama 3 good at sarcasm detection? a comprehensive study.
\newblock In {\em Proceedings of the 2024 7th International Conference on Machine Learning and Machine Intelligence (MLMI)}, pages 141--145, 2024.

\bibitem{mao2025comparative}
Rui Mao, Guanyi Chen, Xiao Li, Mengshi Ge, and Erik Cambria.
\newblock A comparative analysis of metaphorical cognition in chatgpt and human minds.
\newblock {\em Cognitive Computation}, 17(1):35, 2025.

\bibitem{mao2024metapro}
Rui Mao, Kai He, Claudia Ong, Qian Liu, and Erik Cambria.
\newblock Metapro 2.0: Computational metaphor processing on the effectiveness of anomalous language modeling.
\newblock In {\em Findings of the Association for Computational Linguistics ACL 2024}, pages 9891--9908, 2024.

\bibitem{meng2025large}
Haohan Meng, Xiaoyu Li, and Jinhua Sun.
\newblock Large language models prompt engineering as a method for embodied cognitive linguistic representation: a case study of political metaphors in trump’s discourse.
\newblock {\em Frontiers in Psychology}, 16:1591408, 2025.

\bibitem{mikolov2013efficient}
Tomas Mikolov, Kai Chen, Greg Corrado, and Jeffrey Dean.
\newblock Efficient estimation of word representations in vector space.
\newblock {\em arXiv preprint arXiv:1301.3781}, 2013.

\bibitem{mirowski2024robot}
Piotr Mirowski, Juliette Love, Kory Mathewson, and Shakir Mohamed.
\newblock A robot walks into a bar: Can language models serve as creativity supporttools for comedy? an evaluation of llms’ humour alignment with comedians.
\newblock In {\em Proceedings of the 2024 ACM Conference on Fairness, Accountability, and Transparency}, pages 1622--1636, 2024.

\bibitem{mohammad2016metaphor}
Saif Mohammad, Ekaterina Shutova, and Peter Turney.
\newblock Metaphor as a medium for emotion: An empirical study.
\newblock In {\em Proceedings of the fifth joint conference on lexical and computational semantics}, pages 23--33, 2016.

\bibitem{mohler2016introducing}
Michael Mohler, Mary Brunson, Bryan Rink, and Marc Tomlinson.
\newblock Introducing the lcc metaphor datasets.
\newblock In {\em Proceedings of the Tenth International Conference on Language Resources and Evaluation (LREC'16)}, pages 4221--4227, 2016.

\bibitem{mughal2024comparative}
Nimra Mughal, Ghulam Mujtaba, Sarang Shaikh, Aveenash Kumar, and Sher~Muhammad Daudpota.
\newblock Comparative analysis of deep natural networks and large language models for aspect-based sentiment analysis.
\newblock {\em IEEE Access}, 12:60943--60959, 2024.

\bibitem{muller2024recognizing}
Philipp M{\"u}ller, Alexander Heimerl, Sayed~Muddashir Hossain, Lea Siegel, Jan Alexandersson, Patrick Gebhard, Elisabeth Andr{\'e}, and Tanja Schneeberger.
\newblock Recognizing emotion regulation strategies from human behavior with large language models.
\newblock In {\em 2024 12th International Conference on Affective Computing and Intelligent Interaction (ACII)}, pages 210--218. IEEE, 2024.

\bibitem{mun2025leveraging}
Yejoon Mun and Namhyoung Kim.
\newblock Leveraging large language models for sentiment analysis and investment strategy development in financial markets.
\newblock {\em Journal of Theoretical and Applied Electronic Commerce Research}, 20(2):77, 2025.

\bibitem{nadeem2024vision}
Mohammad Nadeem, Shahab~Saquib Sohail, Laeeba Javed, Faisal Anwer, Abdul Khader~Jilani Saudagar, and Khan Muhammad.
\newblock Vision-enabled large language and deep learning models for image-based emotion recognition.
\newblock {\em Cognitive Computation}, 16(5):2566--2579, 2024.

\bibitem{narad2025llms}
Reuben Narad, Siddharth Suresh, Jiayi Chen, Pine~SL Dysart-Bricken, Bob Mankoff, Robert Nowak, Jifan Zhang, and Lalit Jain.
\newblock Which llms get the joke? probing non-stem reasoning abilities with humorbench.
\newblock {\em arXiv preprint arXiv:2507.21476}, 2025.

\bibitem{ng2025examining}
Lynnette Hui~Xian Ng, Iain~J Cruickshank, and Roy Lee.
\newblock Examining the influence of political bias on large language model performance in stance classification.
\newblock In {\em Proceedings of the International AAAI Conference on Web and Social Media}, volume~19, pages 1315--1328, 2025.

\bibitem{niu2024multimodal}
Fuqiang Niu, Zebang Cheng, Xianghua Fu, Xiaojiang Peng, Genan Dai, Yin Chen, Hu~Huang, and Bowen Zhang.
\newblock Multimodal multi-turn conversation stance detection: A challenge dataset and effective model.
\newblock In {\em Proceedings of the 32nd ACM international conference on multimedia}, pages 3867--3876, 2024.

\bibitem{niu2024challenge}
Fuqiang Niu, Min Yang, Ang Li, Baoquan Zhang, Xiaojiang Peng, and Bowen Zhang.
\newblock A challenge dataset and effective models for conversational stance detection.
\newblock {\em arXiv preprint arXiv:2403.11145}, 2024.

\bibitem{niu2025rethinking}
Minxue Niu, Yara El-Tawil, Amrit Romana, and Emily~Mower Provost.
\newblock Rethinking emotion annotations in the era of large language models.
\newblock {\em IEEE Transactions on Affective Computing}, 2025.

\bibitem{ouyang2022training}
Long Ouyang, Jeffrey Wu, Xu~Jiang, Diogo Almeida, Carroll Wainwright, Pamela Mishkin, Chong Zhang, Sandhini Agarwal, Katarina Slama, Alex Ray, et~al.
\newblock Training language models to follow instructions with human feedback.
\newblock {\em Advances in neural information processing systems}, 35:27730--27744, 2022.

\bibitem{peng2024customising}
Liyizhe Peng, Zixing Zhang, Tao Pang, Jing Han, Huan Zhao, Hao Chen, and Bj{\"o}rn~W Schuller.
\newblock Customising general large language models for specialised emotion recognition tasks.
\newblock In {\em ICASSP 2024-2024 IEEE International Conference on Acoustics, Speech and Signal Processing (ICASSP)}, pages 11326--11330. IEEE, 2024.

\bibitem{pit2024whose}
Pagnarasmey Pit, Xingjun Ma, Mike Conway, Qingyu Chen, James Bailey, Henry Pit, Putrasmey Keo, Watey Diep, and Yu-Gang Jiang.
\newblock Whose side are you on? investigating the political stance of large language models.
\newblock {\em arXiv preprint arXiv:2403.13840}, 2024.

\bibitem{pituxcoosuvarnjokes}
Mondheera Pituxcoosuvarn and Yohei Murakami.
\newblock Jokes or gibberish? humor retention in translation with neural machine translation vs. large language model.
\newblock {\em Humor Retention in Translation with Neural Machine Translation vs. Large Language Model}.

\bibitem{poria2021recognizing}
Soujanya Poria, Navonil Majumder, Devamanyu Hazarika, Deepanway Ghosal, Rishabh Bhardwaj, Samson Yu~Bai Jian, Pengfei Hong, Romila Ghosh, Abhinaba Roy, Niyati Chhaya, et~al.
\newblock Recognizing emotion cause in conversations.
\newblock {\em Cognitive Computation}, 13(5):1317--1332, 2021.

\bibitem{qian2025concept}
Wenhao Qian, Zhenzhen Hu, Zijie Song, and Jia Li.
\newblock Concept drift guided layernorm tuning for efficient multimodal metaphor identification.
\newblock In {\em Proceedings of the 2025 International Conference on Multimedia Retrieval}, pages 1100--1108, 2025.

\bibitem{qiu2025detecting}
Ziqi Qiu, Jianxing Yu, Yufeng Zhang, Hanjiang Lai, Yanghui Rao, Qinliang Su, and Jian Yin.
\newblock Detecting emotional incongruity of sarcasm by commonsense reasoning.
\newblock In {\em Proceedings of the 31st International Conference on Computational Linguistics}, pages 9062--9073, 2025.

\bibitem{quan2025can}
Kexin Quan, Pavithra Ramakrishnan, and Jessie Chin.
\newblock Can ai take a joke—or make one? a study of humor generation and recognition in llms.
\newblock In {\em Proceedings of the 2025 Conference on Creativity and Cognition}, pages 431--437, 2025.

\bibitem{radaideh2025fairness}
Mohammed~I Radaideh, O~Hwang Kwon, and Majdi~I Radaideh.
\newblock Fairness and social bias quantification in large language models for sentiment analysis.
\newblock {\em Knowledge-Based Systems}, page 113569, 2025.

\bibitem{radford2023robust}
Alec Radford, Jong~Wook Kim, Tao Xu, Greg Brockman, Christine McLeavey, and Ilya Sutskever.
\newblock Robust speech recognition via large-scale weak supervision.
\newblock In {\em International conference on machine learning}, pages 28492--28518. PMLR, 2023.

\bibitem{radford2018improving}
Alec Radford, Karthik Narasimhan, Tim Salimans, Ilya Sutskever, et~al.
\newblock Improving language understanding by generative pre-training.
\newblock 2018.

\bibitem{radford2019language}
Alec Radford, Jeffrey Wu, Rewon Child, David Luan, Dario Amodei, Ilya Sutskever, et~al.
\newblock Language models are unsupervised multitask learners.
\newblock {\em OpenAI blog}, 1(8):9, 2019.

\bibitem{raffel2020exploring}
Colin Raffel, Noam Shazeer, Adam Roberts, Katherine Lee, Sharan Narang, Michael Matena, Yanqi Zhou, Wei Li, and Peter~J Liu.
\newblock Exploring the limits of transfer learning with a unified text-to-text transformer.
\newblock {\em Journal of machine learning research}, 21(140):1--67, 2020.

\bibitem{rashkin2018towards}
Hannah Rashkin, Eric~Michael Smith, Margaret Li, and Y-Lan Boureau.
\newblock Towards empathetic open-domain conversation models: A new benchmark and dataset.
\newblock {\em arXiv preprint arXiv:1811.00207}, 2018.

\bibitem{rasool2025emotion}
Abdur Rasool, Muhammad~Irfan Shahzad, Hafsa Aslam, Vincent Chan, and Muhammad~Ali Arshad.
\newblock Emotion-aware embedding fusion in large language models (flan-t5, llama 2, deepseek-r1, and chatgpt 4) for intelligent response generation.
\newblock {\em AI}, 6(3):56, 2025.

\bibitem{rohn2024duanzai}
Yesian Rohn.
\newblock Duanzai: Slang-enhanced llm with prompt for humor understanding.
\newblock {\em arXiv preprint arXiv:2405.15818}, 2024.

\bibitem{abs-2411-02666}
Kangrui Ruan, Xinyang Wang, and Xuan Di.
\newblock From twitter to reasoner: Understand mobility travel modes and sentiment using large language models.
\newblock {\em CoRR}, abs/2411.02666, 2024.

\bibitem{sabera2025comparative}
Darrel~Nathaniel Sabera and Dinar~Ajeng Kristiyanti.
\newblock Comparative analysis of large language model as feature extraction methods in sarcasm detection using classification algorithms.
\newblock In {\em 2025 4th International Conference on Electronics Representation and Algorithm (ICERA)}, pages 352--357. IEEE, 2025.

\bibitem{sabour2024emobench}
Sahand Sabour, Siyang Liu, Zheyuan Zhang, June~M Liu, Jinfeng Zhou, Alvionna~S Sunaryo, Juanzi Li, Tatia Lee, Rada Mihalcea, and Minlie Huang.
\newblock Emobench: Evaluating the emotional intelligence of large language models.
\newblock {\em arXiv preprint arXiv:2402.12071}, 2024.

\bibitem{seo2024chacha}
Woosuk Seo, Chanmo Yang, and Young-Ho Kim.
\newblock Chacha: leveraging large language models to prompt children to share their emotions about personal events.
\newblock In {\em Proceedings of the 2024 CHI Conference on Human Factors in Computing Systems}, pages 1--20, 2024.

\bibitem{seshakagari2025dynamic}
Haranadha Reddy~Busireddy Seshakagari, Aravindan Umashankar, T~Harikala, L~Jayasree, and Jeffrey Severance.
\newblock Dynamic financial sentiment analysis and market forecasting through large language models.
\newblock {\em International Journal of Human Computations and Intelligence}, 4(1):397--410, 2025.

\bibitem{shafiei2025not}
Mohammadamin Shafiei and Hamidreza Saffari.
\newblock Not all jokes land: Evaluating large language models understanding of workplace humor.
\newblock {\em arXiv preprint arXiv:2506.01819}, 2025.

\bibitem{shao2024cmdag}
Yujie Shao, Xinrong Yao, Xingwei Qu, Chenghua Lin, Shi Wang, Stephen~W Huang, Ge~Zhang, and Jie Fu.
\newblock Cmdag: A chinese metaphor dataset with annotated grounds as cot for boosting metaphor generation.
\newblock {\em arXiv preprint arXiv:2402.13145}, 2024.

\bibitem{ShaoYCWG25}
Zhiqi Shao, Xusheng Yao, Feng Chen, Ze~Wang, and Junbin Gao.
\newblock Revisiting time-varying dynamics in stock market forecasting: {A} multi-source sentiment analysis approach with large language model.
\newblock {\em Decis. Support Syst.}, 190:114362, 2025.

\bibitem{abs-2310-18025}
Paul~F. Simmering and Paavo Huoviala.
\newblock Large language models for aspect-based sentiment analysis.
\newblock {\em CoRR}, abs/2310.18025, 2023.

\bibitem{smith2023hallucination}
Andrew~L Smith, Felix Greaves, and Trishan Panch.
\newblock Hallucination or confabulation? neuroanatomy as metaphor in large language models.
\newblock {\em PLOS Digital Health}, 2(11):e0000388, 2023.

\bibitem{smith2022using}
Shaden Smith, Mostofa Patwary, Brandon Norick, Patrick LeGresley, Samyam Rajbhandari, Jared Casper, Zhun Liu, Shrimai Prabhumoye, George Zerveas, Vijay Korthikanti, et~al.
\newblock Using deepspeed and megatron to train megatron-turing nlg 530b, a large-scale generative language model.
\newblock {\em arXiv preprint arXiv:2201.11990}, 2022.

\bibitem{song2025emotion}
Changhao Song, Yazhou Zhang, and Peng Zhang.
\newblock Emotion-o1: Adaptive long reasoning for emotion understanding in llms.
\newblock {\em arXiv preprint arXiv:2505.22548}, 2025.

\bibitem{srirag2024besstie}
Dipankar Srirag, Aditya Joshi, Jordan Painter, and Diptesh Kanojia.
\newblock Besstie: A benchmark for sentiment and sarcasm classification for varieties of english.
\newblock {\em arXiv preprint arXiv:2412.04726}, 2024.

\bibitem{steen2010method}
Gerard~J Steen, Aletta~G Dorst, Tina Krennmayr, Anna~A Kaal, and J~Berenike Herrmann.
\newblock A method for linguistic metaphor identification.
\newblock 2010.

\bibitem{StigallKANP24}
William Stigall, Md~Abdullah Al~Hafiz Khan, Dinesh~Chowdary Attota, Francis Nweke, and Yong Pei.
\newblock Large language models performance comparison of emotion and sentiment classification.
\newblock In {\em Proceedings of the 2024 {ACM} Southeast Conference, {ACM} {SE} 2024, Marietta, GA, USA, April 18-20, 2024}, pages 60--68. {ACM}, 2024.

\bibitem{suhartono2024idsarcasm}
Derwin Suhartono, Wilson Wongso, and Alif~Tri Handoyo.
\newblock Idsarcasm: Benchmarking and evaluating language models for indonesian sarcasm detection.
\newblock {\em IEEE Access}, 12:87323--87332, 2024.

\bibitem{tang2024leveraging}
Binghao Tang, Boda Lin, Haolong Yan, and Si~Li.
\newblock Leveraging generative large language models with visual instruction and demonstration retrieval for multimodal sarcasm detection.
\newblock In {\em Proceedings of the 2024 Conference of the North American Chapter of the Association for Computational Linguistics: Human Language Technologies (Volume 1: Long Papers)}, pages 1732--1742, 2024.

\bibitem{tang2025metaphor}
Jiacheng Tang, Nankai Wu, Fan Gao, Chengxiao Dai, Mengyao Zhao, and Xinjie Zhao.
\newblock From metaphor to mechanism: How llms decode traditional chinese medicine symbolic language for modern clinical relevance.
\newblock {\em arXiv preprint arXiv:2503.02760}, 2025.

\bibitem{taori2023alpaca}
Rohan Taori, Ishaan Gulrajani, Tianyi Zhang, Yann Dubois, Xuechen Li, Carlos Guestrin, Percy Liang, and Tatsunori~B Hashimoto.
\newblock Alpaca: A strong, replicable instruction-following model.
\newblock {\em Stanford Center for Research on Foundation Models. https://crfm. stanford. edu/2023/03/13/alpaca. html}, 3(6):7, 2023.

\bibitem{taranukhin2024stance}
Maksym Taranukhin, Vered Shwartz, and Evangelos Milios.
\newblock Stance reasoner: Zero-shot stance detection on social media with explicit reasoning.
\newblock {\em arXiv preprint arXiv:2403.14895}, 2024.

\bibitem{team2023gemini}
Gemini Team, Rohan Anil, Sebastian Borgeaud, Jean-Baptiste Alayrac, Jiahui Yu, Radu Soricut, Johan Schalkwyk, Andrew~M Dai, Anja Hauth, Katie Millican, et~al.
\newblock Gemini: a family of highly capable multimodal models.
\newblock {\em arXiv preprint arXiv:2312.11805}, 2023.

\bibitem{team2024gemini}
Gemini Team, Petko Georgiev, Ving~Ian Lei, Ryan Burnell, Libin Bai, Anmol Gulati, Garrett Tanzer, Damien Vincent, Zhufeng Pan, Shibo Wang, et~al.
\newblock Gemini 1.5: Unlocking multimodal understanding across millions of tokens of context.
\newblock {\em arXiv preprint arXiv:2403.05530}, 2024.

\bibitem{team2024gemma}
Gemma Team, Morgane Riviere, Shreya Pathak, Pier~Giuseppe Sessa, Cassidy Hardin, Surya Bhupatiraju, L{\'e}onard Hussenot, Thomas Mesnard, Bobak Shahriari, Alexandre Ram{\'e}, et~al.
\newblock Gemma 2: Improving open language models at a practical size.
\newblock {\em arXiv preprint arXiv:2408.00118}, 2024.

\bibitem{tian2025imara}
Yuan Tian, Minzheng Wang, Nan Xu, and Wenji Mao.
\newblock Imara: An imaginative frame augmented method for low-resource multimodal metaphor detection and explanation.
\newblock In {\em Findings of the Association for Computational Linguistics: NAACL 2025}, pages 3953--3967, 2025.

\bibitem{tian2024theory}
Yuan Tian, Nan Xu, and Wenji Mao.
\newblock A theory guided scaffolding instruction framework for llm-enabled metaphor reasoning.
\newblock In {\em Proceedings of the 2024 Conference of the North American Chapter of the Association for Computational Linguistics: Human Language Technologies (Volume 1: Long Papers)}, pages 7731--7748, 2024.

\bibitem{tigges2023linear}
Curt Tigges, Oskar~John Hollinsworth, Atticus Geiger, and Neel Nanda.
\newblock Linear representations of sentiment in large language models.
\newblock {\em arXiv preprint arXiv:2310.15154}, 2023.

\bibitem{tong2024metaphor}
Xiaoyu Tong, Rochelle Choenni, Martha Lewis, and Ekaterina Shutova.
\newblock Metaphor understanding challenge dataset for llms.
\newblock {\em arXiv preprint arXiv:2403.11810}, 2024.

\bibitem{touvron2023llama}
Hugo Touvron, Thibaut Lavril, Gautier Izacard, Xavier Martinet, Marie-Anne Lachaux, Timoth{\'e}e Lacroix, Baptiste Rozi{\`e}re, Naman Goyal, Eric Hambro, Faisal Azhar, et~al.
\newblock Llama: Open and efficient foundation language models.
\newblock {\em arXiv preprint arXiv:2302.13971}, 2023.

\bibitem{touvron2023llama2}
Hugo Touvron, Louis Martin, Kevin Stone, Peter Albert, Amjad Almahairi, Yasmine Babaei, Nikolay Bashlykov, Soumya Batra, Prajjwal Bhargava, Shruti Bhosale, et~al.
\newblock Llama 2: Open foundation and fine-tuned chat models.
\newblock {\em arXiv preprint arXiv:2307.09288}, 2023.

\bibitem{vasilakes2025exploring}
Jake Vasilakes, Carolina Scarton, and Zhixue Zhao.
\newblock Exploring vision language models for multimodal and multilingual stance detection.
\newblock {\em arXiv preprint arXiv:2501.17654}, 2025.

\bibitem{vaswani2017attention}
Ashish Vaswani, Noam Shazeer, Niki Parmar, Jakob Uszkoreit, Llion Jones, Aidan~N Gomez, {\L}ukasz Kaiser, and Illia Polosukhin.
\newblock Attention is all you need.
\newblock {\em Advances in neural information processing systems}, 30, 2017.

\bibitem{venerito2024large}
Vincenzo Venerito and Florenzo Iannone.
\newblock Large language model-driven sentiment analysis for facilitating fibromyalgia diagnosis.
\newblock {\em RMD open}, 10(2), 2024.

\bibitem{vzorinab2024emotional}
Gleb~D Vzorinab, Alexey~M Bukinichac, Anna~V Sedykha, Irina~I Vetrovab, and Elena~A Sergienkob.
\newblock The emotional intelligence of the gpt-4 large language model.
\newblock {\em Psychology in Russia: State of the art}, 17(2):85--99, 2024.

\bibitem{wachowiak2023does}
Lennart Wachowiak and Dagmar Gromann.
\newblock Does gpt-3 grasp metaphors? identifying metaphor mappings with generative language models.
\newblock In {\em Proceedings of the 61st annual meeting of the association for computational linguistics (volume 1: Long papers)}, pages 1018--1032, 2023.

\bibitem{wang2024memecraft}
Han Wang and Roy Ka-Wei Lee.
\newblock Memecraft: Contextual and stance-driven multimodal meme generation.
\newblock In {\em Proceedings of the ACM Web Conference 2024}, pages 4642--4652, 2024.

\bibitem{WANG2025115802}
Hongyu Wang, Weiqi Hua, Jinqing Peng, and Maomao Hu.
\newblock Public sentiment analysis of data center energy consumption using social media data and large language models.
\newblock {\em Energy and Buildings}, 341:115802, 2025.

\bibitem{wang2024s3}
Peng Wang, Yongheng Zhang, Hao Fei, Qiguang Chen, Yukai Wang, Jiasheng Si, Wenpeng Lu, Min Li, and Libo Qin.
\newblock S3 agent: unlocking the power of vllm for zero-shot multi-modal sarcasm detection.
\newblock {\em ACM Transactions on Multimedia Computing, Communications and Applications}, 2024.

\bibitem{wang2024deem}
Xiaolong Wang, Yile Wang, Sijie Cheng, Peng Li, and Yang Liu.
\newblock Deem: Dynamic experienced expert modeling for stance detection.
\newblock {\em arXiv preprint arXiv:2402.15264}, 2024.

\bibitem{wang2023emotional}
Xuena Wang, Xueting Li, Zi~Yin, Yue Wu, and Jia Liu.
\newblock Emotional intelligence of large language models.
\newblock {\em Journal of Pacific Rim Psychology}, 17:18344909231213958, 2023.

\bibitem{wei2022chain}
Jason Wei, Xuezhi Wang, Dale Schuurmans, Maarten Bosma, Fei Xia, Ed~Chi, Quoc~V Le, Denny Zhou, et~al.
\newblock Chain-of-thought prompting elicits reasoning in large language models.
\newblock {\em Advances in neural information processing systems}, 35:24824--24837, 2022.

\bibitem{welivita2024large}
Anuradha Welivita and Pearl Pu.
\newblock Are large language models more empathetic than humans?
\newblock {\em arXiv preprint arXiv:2406.05063}, 2024.

\bibitem{Wu_2025}
Chengyan Wu, Bolei Ma, Zheyu Zhang, Ningyuan Deng, Yanqing He, and Yun Xue.
\newblock Evaluating zero-shot multilingual aspect-based sentiment analysis with large language models.
\newblock {\em International Journal of Machine Learning and Cybernetics}, June 2025.

\bibitem{wu2023sect}
Lifang Wu, Lehao Xing, Ge~Shi, Sinuo Deng, and Jie Yang.
\newblock Sect: Sentiment-enriched continual training for image sentiment analysis.
\newblock In {\em International Conference on Image and Graphics}, pages 93--105. Springer, 2023.

\bibitem{wu2024humour}
Shih-Hung Wu, Yu-Feng Huang, and Tsz-Yeung Lau.
\newblock Humour classification by fine-tuning llms: Cyut at clef 2024 joker lab subtask humour classification according to genre and technique.
\newblock In {\em Working Notes of the Conference and Labs of the Evaluation Forum (CLEF 2024). CEUR Workshop Proceedings}, pages 1933--1947, 2024.

\bibitem{wu2025humorreject}
Zihui Wu, Haichang Gao, Jiacheng Luo, and Zhaoxiang Liu.
\newblock Humorreject: Decoupling llm safety from refusal prefix via a little humor.
\newblock {\em arXiv preprint arXiv:2501.13677}, 2025.

\bibitem{xiang2025dynamic}
Mengyu Xiang, Yuxuan Song, Qiudan Li, Shu Wu, and Daniel~Dajun Zeng.
\newblock Dynamic detection of sarcasm topic-target pairs via llm-based knowledge alignment.
\newblock In {\em Companion Proceedings of the ACM on Web Conference 2025}, pages 1422--1425, 2025.

\bibitem{xie2024emovit}
Hongxia Xie, Chu-Jun Peng, Yu-Wen Tseng, Hung-Jen Chen, Chan-Feng Hsu, Hong-Han Shuai, and Wen-Huang Cheng.
\newblock Emovit: Revolutionizing emotion insights with visual instruction tuning.
\newblock In {\em Proceedings of the IEEE/CVF Conference on Computer Vision and Pattern Recognition}, pages 26596--26605, 2024.

\bibitem{xing2025emotionhallucer}
Bohao Xing, Xin Liu, Guoying Zhao, Chengyu Liu, Xiaolan Fu, and Heikki K{\"a}lvi{\"a}inen.
\newblock Emotionhallucer: Evaluating emotion hallucinations in multimodal large language models.
\newblock {\em arXiv preprint arXiv:2505.11405}, 2025.

\bibitem{xing2024emo}
Bohao Xing, Zitong Yu, Xin Liu, Kaishen Yuan, Qilang Ye, Weicheng Xie, Huanjing Yue, Jingyu Yang, and Heikki K{\"a}lvi{\"a}inen.
\newblock Emo-llama: Enhancing facial emotion understanding with instruction tuning.
\newblock {\em arXiv preprint arXiv:2408.11424}, 2024.

\bibitem{xu2024secap}
Yaoxun Xu, Hangting Chen, Jianwei Yu, Qiaochu Huang, Zhiyong Wu, Shi-Xiong Zhang, Guangzhi Li, Yi~Luo, and Rongzhi Gu.
\newblock Secap: Speech emotion captioning with large language model.
\newblock In {\em Proceedings of the AAAI Conference on Artificial Intelligence}, volume~38, pages 19323--19331, 2024.

\bibitem{yakura2024evaluating}
Hiromu Yakura.
\newblock Evaluating large language models’ ability using a psychiatric screening tool based on metaphor and sarcasm scenarios.
\newblock {\em Journal of Intelligence}, 12(7):70, 2024.

\bibitem{yan2024collaborative}
Ming Yan, Tianyi~Zhou Joey, and W~Tsang Ivor.
\newblock Collaborative knowledge infusion for low-resource stance detection.
\newblock {\em Big Data Mining and Analytics}, 7(3):682--698, 2024.

\bibitem{yan2025collaborative}
Yu~Yan, Sheng Sun, Zixiang Tang, Teli Liu, and Min Liu.
\newblock Collaborative stance detection via small-large language model consistency verification.
\newblock {\em arXiv preprint arXiv:2502.19954}, 2025.

\bibitem{yang2024qwen2technicalreport}
An~Yang, Baosong Yang, Binyuan Hui, Bo~Zheng, Bowen Yu, Chang Zhou, Chengpeng Li, Chengyuan Li, Dayiheng Liu, Fei Huang, Guanting Dong, Haoran Wei, Huan Lin, Jialong Tang, Jialin Wang, Jian Yang, Jianhong Tu, Jianwei Zhang, Jianxin Ma, Jianxin Yang, Jin Xu, Jingren Zhou, Jinze Bai, Jinzheng He, Junyang Lin, Kai Dang, Keming Lu, Keqin Chen, Kexin Yang, Mei Li, Mingfeng Xue, Na~Ni, Pei Zhang, Peng Wang, Ru~Peng, Rui Men, Ruize Gao, Runji Lin, Shijie Wang, Shuai Bai, Sinan Tan, Tianhang Zhu, Tianhao Li, Tianyu Liu, Wenbin Ge, Xiaodong Deng, Xiaohuan Zhou, Xingzhang Ren, Xinyu Zhang, Xipin Wei, Xuancheng Ren, Xuejing Liu, Yang Fan, Yang Yao, Yichang Zhang, Yu~Wan, Yunfei Chu, Yuqiong Liu, Zeyu Cui, Zhenru Zhang, Zhifang Guo, and Zhihao Fan.
\newblock Qwen2 technical report, 2024.

\bibitem{yang2024advancing}
Haowei Yang, Yun Zi, Honglin Qin, Hongye Zheng, and Yuxiang Hu.
\newblock Advancing emotional analysis with large language models.
\newblock {\em Journal of Computer Science and Software Applications}, 4(3):8--15, 2024.

\bibitem{yang2025omni}
Qize Yang, Detao Bai, Yi-Xing Peng, and Xihan Wei.
\newblock Omni-emotion: Extending video mllm with detailed face and audio modeling for multimodal emotion analysis.
\newblock {\em arXiv preprint arXiv:2501.09502}, 2025.

\bibitem{yang2024emollm}
Qu~Yang, Mang Ye, and Bo~Du.
\newblock Emollm: Multimodal emotional understanding meets large language models.
\newblock {\em arXiv preprint arXiv:2406.16442}, 2024.

\bibitem{yang2024reinforcement}
Ruichao Yang, Wei Gao, Jing Ma, Hongzhan Lin, and Bo~Wang.
\newblock Reinforcement tuning for detecting stances and debunking rumors jointly with large language models.
\newblock {\em arXiv preprint arXiv:2406.02143}, 2024.

\bibitem{YangLGPWYI25}
Yunchu Yang, Jiaxuan Li, Jielong Guo, Patrick~Cheong{-}Iao Pang, Yapeng Wang, Xu~Yang, and Sio~Kei Im.
\newblock Performance evaluation and application potential of small large language models in complex sentiment analysis tasks.
\newblock {\em {IEEE} Access}, 13:49007--49017, 2025.

\bibitem{yao2407sarcasm}
B~Yao, Y~Zhang, Q~Li, and J~Qin.
\newblock Is sarcasm detection a step-by-step reasoning process in large language models?(2024).
\newblock {\em arXiv preprint arXiv:2407.12725}.

\bibitem{yao2025sarcasm}
Ben Yao, Yazhou Zhang, Qiuchi Li, and Jing Qin.
\newblock Is sarcasm detection a step-by-step reasoning process in large language models?
\newblock In {\em Proceedings of the AAAI Conference on Artificial Intelligence}, volume~39, pages 25651--25659, 2025.

\bibitem{yarmand2025towards}
Matin Yarmand, Courtney~N Reed, Udayan Tandon, Eric~B Hekler, Nadir Weibel, and April~Yi Wang.
\newblock Towards dialogic and on-demand metaphors for interdisciplinary reading.
\newblock In {\em Proceedings of the 2025 CHI Conference on Human Factors in Computing Systems}, pages 1--19, 2025.

\bibitem{yazhou2024can}
Zhang Yazhou, Wang Mengyao, Rong Lu, Yu~Yang, Zhao Dongming, and Qin Jing.
\newblock Can chatgpt be served as the sentiment expert? an evaluation of chatgpt on sentiment and metaphor analysis.
\newblock {\em Acta Scientiarum Naturalium Universitatis Pekinensis}, 60(1):43--52, 2024.

\bibitem{yi2023exploring}
Guofeng Yi, Yuguang Yang, Yu~Pan, Yuhang Cao, Jixun Yao, Xiang Lv, Cunhang Fan, Zhao Lv, Jianhua Tao, Shan Liang, et~al.
\newblock Exploring the power of cross-contextual large language model in mimic emotion prediction.
\newblock In {\em Proceedings of the 4th on Multimodal Sentiment Analysis Challenge and Workshop: Mimicked Emotions, Humour and Personalisation}, pages 19--26, 2023.

\bibitem{yongsatianchot2023s}
Nutchanon Yongsatianchot, Tobias Thejll-Madsen, and Stacy Marsella.
\newblock What’s next in affective modeling? large language models.
\newblock In {\em 2023 11th International Conference on Affective Computing and Intelligent Interaction Workshops and Demos (ACIIW)}, pages 1--7. IEEE, 2023.

\bibitem{yu2025compound}
Jun Yu and Xilong Lu.
\newblock Compound expression recognition via large vision-language models.
\newblock {\em arXiv preprint arXiv:2503.11241}, 2025.

\bibitem{yu2025cfunmodel}
Zhenghan Yu, Xinyu Hu, and Xiaojun Wan.
\newblock Cfunmodel: A" funny" language model capable of chinese humor generation and processing.
\newblock {\em arXiv preprint arXiv:2503.20417}, 2025.

\bibitem{zhang2025logic}
Bowen Zhang, Jun Ma, Xianghua Fu, and Genan Dai.
\newblock Logic augmented multi-decision fusion framework for stance detection on social media.
\newblock {\em Information Fusion}, page 103214, 2025.

\bibitem{zhang2023instruct}
Boyu Zhang, Hongyang Yang, and Xiao-Yang Liu.
\newblock Instruct-fingpt: Financial sentiment analysis by instruction tuning of general-purpose large language models.
\newblock {\em arXiv preprint arXiv:2306.12659}, 2023.

\bibitem{zhang2023enhancing}
Boyu Zhang, Hongyang Yang, Tianyu Zhou, Muhammad Ali~Babar, and Xiao-Yang Liu.
\newblock Enhancing financial sentiment analysis via retrieval augmented large language models.
\newblock In {\em Proceedings of the fourth ACM international conference on AI in finance}, pages 349--356, 2023.

\bibitem{zhang2025multimodal}
Dongyu Zhang, Xingyuan Lu, Mulin Zhuang, Senqi Yang, and Hongjun Chen.
\newblock Multimodal metaphor recognition based on chain-of-cognition prompting.
\newblock {\em Cognitive Systems Research}, 91:101356, 2025.

\bibitem{zhang2025towards}
Dongyu Zhang, Shengcheng Yin, Jingwei Yu, Zhiyao Wu, Zhen Li, Chengpei Xu, Xiaoxia Wang, and Feng Xia.
\newblock Towards multimodal metaphor understanding: A chinese dataset and model for metaphor mapping identification.
\newblock {\em arXiv preprint arXiv:2501.02434}, 2025.

\bibitem{zhang2025revisiting}
Ting Zhang, Ivana~Clairine Irsan, Ferdian Thung, and David Lo.
\newblock Revisiting sentiment analysis for software engineering in the era of large language models.
\newblock {\em ACM Transactions on Software Engineering and Methodology}, 34(3):1--30, 2025.

\bibitem{zhang2023sentiment}
Wenxuan Zhang, Yue Deng, Bing Liu, Sinno~Jialin Pan, and Lidong Bing.
\newblock Sentiment analysis in the era of large language models: A reality check.
\newblock {\em arXiv preprint arXiv:2305.15005}, 2023.

\bibitem{zhang2023dialoguellm}
Y~Zhang, M~Wang, Y~Wu, Li~Q Tiwari~Prayag, B~Wang, and J~Qin.
\newblock Dialoguellm: context and emotion knowledge-tuned large language models for emotion recognition in conversations. arxiv. org, 2023.

\bibitem{zhang2024sarcasmbench}
Y~Zhang, C~Zou, Z~Lian, P~Tiwari, and J~Qin.
\newblock Sarcasmbench: Towards evaluating large language models on sarcasm understanding (no. arxiv: 2408.11319). arxiv, 2024.

\bibitem{zhang2025pushing}
Yazhou Zhang, Mengyao Wang, Qiuchi Li, Prayag Tiwari, and Jing Qin.
\newblock Pushing the limit of llm capacity for text classification.
\newblock In {\em Companion Proceedings of the ACM on Web Conference 2025}, pages 1524--1528, 2025.

\bibitem{zhang2025mlms}
Yazhou Zhang, Chunwang Zou, Qimeng Liu, Lu~Rong, Ben Yao, Zheng Lian, Qiuchi Li, Peng Zhang, and Jing Qin.
\newblock Are mlms trapped in the visual room?
\newblock {\em arXiv preprint arXiv:2505.23272}, 2025.

\bibitem{zhang2025commander}
Yazhou Zhang, Chunwang Zou, Bo~Wang, and Jing Qin.
\newblock Commander-gpt: Fully unleashing the sarcasm detection capability of multi-modal large language models.
\newblock {\em arXiv preprint arXiv:2503.18681}, 2025.

\bibitem{zhang2024affective}
Yiqun Zhang, Xiaocui Yang, Xingle Xu, Zeran Gao, Yijie Huang, Shiyi Mu, Shi Feng, Daling Wang, Yifei Zhang, Kaisong Song, et~al.
\newblock Affective computing in the era of large language models: A survey from the nlp perspective.
\newblock {\em arXiv preprint arXiv:2408.04638}, 2024.

\bibitem{zhang2024llm}
Zhao Zhang, Yiming Li, Jin Zhang, and Hui Xu.
\newblock Llm-driven knowledge injection advances zero-shot and cross-target stance detection.
\newblock In {\em Proceedings of the 2024 Conference of the North American Chapter of the Association for Computational Linguistics: Human Language Technologies (Volume 2: Short Papers)}, pages 371--378, 2024.

\bibitem{zhang2024refashioning}
Zixing Zhang, Liyizhe Peng, Tao Pang, Jing Han, Huan Zhao, and Bj{\"o}rn~W Schuller.
\newblock Refashioning emotion recognition modeling: the advent of generalized large models.
\newblock {\em IEEE Transactions on Computational Social Systems}, 11(5):6690--6704, 2024.

\bibitem{zhao2024ez}
Chenye Zhao and Cornelia Caragea.
\newblock Ez-stance: A large dataset for english zero-shot stance detection.
\newblock In {\em Proceedings of the 62nd Annual Meeting of the Association for Computational Linguistics (Volume 1: Long Papers)}, pages 15697--15714, 2024.

\bibitem{zhao2024zerostance}
Chenye Zhao, Yingjie Li, Cornelia Caragea, and Yue Zhang.
\newblock Zerostance: Leveraging chatgpt for open-domain stance detection via dataset generation.
\newblock In {\em Findings of the Association for Computational Linguistics ACL 2024}, pages 13390--13405, 2024.

\bibitem{zhao2025eilmob}
Haochen Zhao, Yongxiu Xu, Xinkui Lin, Jiarui Lu, Hongbo Xu, and Yubin Wang.
\newblock Eilmob: Emotion-aware incongruity learning and modality bridging network for multi-modal sarcasm detection.
\newblock In {\em Proceedings of the 2025 International Conference on Multimedia Retrieval}, pages 1868--1876, 2025.

\bibitem{zhao2025r1}
Jiaxing Zhao, Xihan Wei, and Liefeng Bo.
\newblock R1-omni: Explainable omni-multimodal emotion recognition with reinforcement learning.
\newblock {\em arXiv preprint arXiv:2503.05379}, 2025.

\bibitem{zheng2025enhancing}
Li~Zheng, Sihang Wang, Hao Fei, Zuquan Peng, Fei Li, Jianming Fu, Chong Teng, and Donghong Ji.
\newblock Enhancing hyperbole and metaphor detection with their bidirectional dynamic interaction and emotion knowledge.
\newblock {\em arXiv preprint arXiv:2506.15504}, 2025.

\bibitem{zhou2025ldgnet}
Hengyang Zhou, Jinwu Yan, Yaqing Chen, Rongman Hong, Wenbo Zuo, and Keyan Jin.
\newblock Ldgnet: Llms debate-guided network for multimodal sarcasm detection.
\newblock In {\em ICASSP 2025-2025 IEEE International Conference on Acoustics, Speech and Signal Processing (ICASSP)}, pages 1--5. IEEE, 2025.

\bibitem{zhou2023evaluation}
Juliann Zhou.
\newblock An evaluation of state-of-the-art large language models for sarcasm detection.
\newblock {\em arXiv preprint arXiv:2312.03706}, 2023.

\bibitem{zhou2025cascade}
Runlong Zhou and Yi~Zhang.
\newblock Cascade your datasets for cross-mode knowledge retrieval of language models.
\newblock {\em arXiv preprint arXiv:2504.01450}, 2025.

\bibitem{zhu2025ratsd}
Zhengyuan Zhu, Zeyu Zhang, Haiqi Zhang, and Chengkai Li.
\newblock Ratsd: Retrieval augmented truthfulness stance detection from social media posts toward factual claims.
\newblock In {\em Findings of the Association for Computational Linguistics: NAACL 2025}, pages 3366--3381, 2025.

\bibitem{ZhuangWCH25}
Yong Zhuang, Feilong Wang, Dickson K.~W. Chiu, and Kevin K.~W. Ho.
\newblock Leveraging large language models to examine the interaction between investor sentiment and stock performance.
\newblock {\em Eng. Appl. Artif. Intell.}, 150:110602, 2025.

\end{thebibliography}
\end{document}